\documentclass{article} 

\usepackage{graphicx}     
\usepackage{caption}      
\usepackage{subcaption}   
\usepackage{booktabs}     
\usepackage{times}

\usepackage{wrapfig}
\usepackage{enumitem}
\usepackage{etoc}

\usepackage{geometry}[margin=1in]
\usepackage{natbib}
\usepackage{authblk}  

\usepackage{amsmath,amsthm,amsfonts,bm}

\newcommand{\m}[1]{\textmd{\textup{\texttt{#1}}}}

\newcommand{\mm}[1]{\textmd{\textup{\texttt{#1}}}}

\newcommand{\SSM}{\text{SSM}}
\newcommand{\TF}{\text{TF}}
\newcommand{\GDN}{\text{GDN}}

\newcommand{\x}{\mathbf{x}}
\newcommand{\f}{\mathbf{f}}

\def\eqref#1{equation~\ref{#1}}

\def\1{\bm{1}}

\def\eps{{\epsilon}}

\DeclareMathAlphabet{\mathsfit}{\encodingdefault}{\sfdefault}{m}{sl}
\SetMathAlphabet{\mathsfit}{bold}{\encodingdefault}{\sfdefault}{bx}{n}

\def\gM{{\mathcal{M}}}
\def\gN{{\mathcal{N}}}

\def\gQ{{\mathcal{Q}}}

\def\gS{{\mathcal{S}}}

\def\gU{{\mathcal{U}}}
\def\gV{{\mathcal{V}}}

\def\gY{{\mathcal{Y}}}

\newcommand{\E}{\mathbb{E}}

\newcommand{\R}{\mathbb{R}}

\newcommand\card[1]{| #1|}

\renewcommand{\x}{\mathbf{x}}
\newcommand{\z}{\mathbf{z}}

\usepackage{amsthm}

\newtheorem{theorem}{Theorem}
\newtheorem{lemma}{Lemma}

\theoremstyle{definition}
\newtheorem{definition}{Definition}
\newtheorem{assumption}{Assumption}

\theoremstyle{remark}

\newcommand{\Attn}{\mathrm{Attn}}

\newcommand{\AT}{\mathrm{AT}}
\newcommand{\MLP}{\mathrm{MLP}}

\newcommand{\W}{\mathbf W}
\usepackage[colorlinks,citecolor=blue,linkcolor=magenta,bookmarks=true]{hyperref}
\usepackage[nameinlink,capitalise]{cleveref}
\usepackage{url}

\usepackage{tikz}

\title{The Effect of Layer Ordering in Hybrid Architectures}

\author[*1]{John Cooper}
\author[*1]{Mingchen Ma}
\author[*1]{Jiaqi Mo}
\author[*1]{Shengqi Qiu}
\author[2]{\authorcr Nathaniel Bastian}
\author[1]{Ilias Diakonikolas}
\author[1]{Frederic Sala}

\affil[1]{Department of Computer Sciences, University of Wisconsin-Madison}
\affil[2]{Department of Computer Sciences, John Hopkins University}

\affil[ ]{\texttt{\{jfcooper2, mingchen\}@cs.wisc.edu}}
\date{}

\begin{document}

\maketitle
\def\thefootnote{*}\footnotetext{Equal contribution. Authors listed in alphabetical order.}\def\thefootnote{\arabic{footnote}}

\begin{abstract}

Hybrid attention is rapidly becoming the dominant architectural choice for frontier models. 
Despite their successes, such architectures complicate model design by introducing additional choices, including the types and locations of layers.
This combinatorially large design space has been studied empirically, but given the vast expense of training and evaluating models with different configurations, a principled understanding of the effects of layer selection and ordering is needed.
We provide the first such analysis, showing that the dominant hybrid architecture is highly task-specific.
Concretely, for any layer configuration, we prove that there exists a task where we can construct a model with this configuration that provably outperforms any other model of any different configuration under fixed resource constraints.
Empirically, we observe that for standard training, learned hybrid architectures exhibit separations between layer orderings that match our theoretical findings.  
For instance, under the same memory budget, the optimal architecture predicted by our theory for a segmented retrieval task consistently achieves perfect accuracy, whereas the next strongest plateaus below 0.5. 
These results extend to deeper models, where ``containing'' the dominant architecture for a task indicates whether the model attains high performance.
In addition, we investigate the mechanisms learned by hybrids, observing that Transformer layers contribute the largest share of performance on associative recall across architectures. 
This work establishes insights into task-dependent hybrid architecture design relevant to the training and scaling of frontier-scale hybrids.
\footnote{Code is available at \href{https://github.com/SprocketLab/layer-importance}{https://github.com/SprocketLab/layer-importance}.}

\end{abstract}

\section{Introduction}

Hybrid architectures have become the design of choice for frontier models~\citep{nvidia2025nemotron3, qiu2026qwen38next, kimiteam2025kimilinear}. 
This is due to their superior scalability: the quadratic complexity of full self-attention limits the vanilla Transformer architecture on long-context tasks~\citep{Keles2022OnTC, pope2023kvcache}. Hybrid models replace many self-attention layers with efficient subquadratic alternatives that require smaller memory and offer faster inference. Popular choices include state-space layers like linear attention, Mamba, Gated DeltaNet, etc.~\citep{gu2023mamba, yang2025gated}. The use of these layers produces scalable and highly-performant models.

However, as each layer can be selected from among many variations, layer selection and ordering is a critical design challenge~\citep{yang2025gated,gu2023mamba, 3524938.3525416}. Indeed, each layer type brings different constraints when processing sequences: full-attention is memory-bound by retrieval distance while  recurrent alternatives are bound by their state size. The most popular layer ordering among frontier models uses a block design that involves one softmax-attention layer followed by many subquadratic layers. This is supported empirically~\citep{waleffe2024empiricalstudymambabasedlanguage}. While some work has investigated the selection ratio between softmax-attention and linear attention \citep{wang2026systematicanalysishybridlinear, fu2025nemotronflash} and others have shown a theoretical separation with hybrids utilizing less memory \citep{cooper2026expressivityefficiency}, the benefits and drawbacks of various layer orderings are still unclear. Given the combinatorially large space of different layer orderings, a complete empirical sweep across architectures is intractable.

A theoretical characterization of hybrid architectures would cut down on this combinatorial space.  %
However, identifying differences between hybrid layer orderings is highly challenging. Tasks can be solved by a variety of algorithms, each best suited for a different hybrid architecture. For instance, the greatest difficulty stems from processing order: some architectures focus on reordering the sequence in early layers to allow for a few later layers to solve the task. These shortcuts can reduce model size and depth compared to more direct algorithms with sequential processing.

\begin{figure}
    \centering
\includegraphics[width=\linewidth]{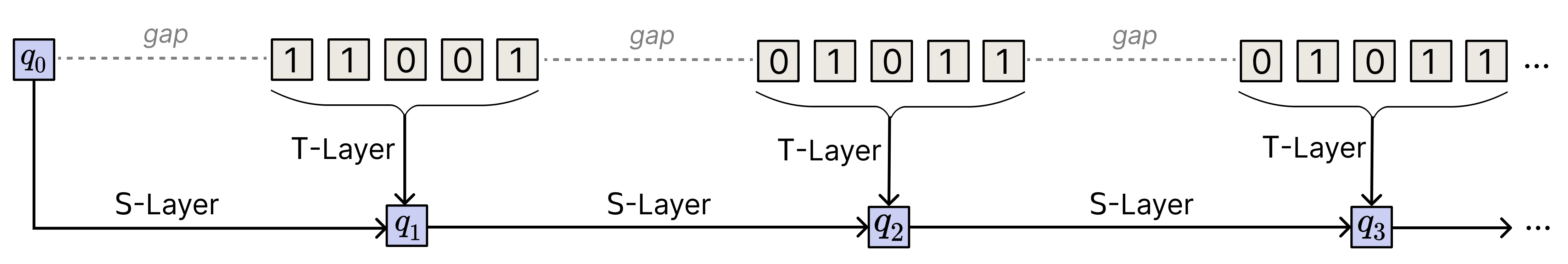}
    \caption{Structure of tasks that exhibit performance differences among different hybrid architectures. Large gaps need SSM layers to communicate each segment's query, while the dense computation per segment is done by the Transformers.}
    \label{fig:figure1}
\end{figure}
In this work, we obtain a new \textbf{theory-driven approach to hybrid architecture design}. Inspired by \cite{cooper2026expressivityefficiency}, we study the expressivity of different hybrid architectures under computational resource constraints. We identify tasks and associated architecture designs where no other design (i.e., layer ordering) is able to solve the task under identical constraints. We begin by abstracting full-attention ($\TF$) and sub-quadratic attention variants as generalized windowed functions and generalized state-space models (SSMs), respectively. This allows us to focus on the critical elements constraining layer types: the window width, state size, and embedding precision. For a given layer ordering of generalized SSMs and generalized causal windowed models, we show \textit{there exists a task and hybrid architecture such that any architecture deviating from the provided one must use more resources at some layer}. 
This indicates that large-scale hybrid architecture design is situation-specific; in other words, structural properties of a desired task determines the optimal hybrid architecture.

In addition, we zoom in on 2- and 3-layer hybrid architectures to characterize and expand the set of such \emph{separating} tasks. These tasks are natural, mirroring behaviors found in real-world natural language scenarios.  
Specifically, we establish tasks where the $\TF\to\SSM$ (\m{TS}) and $\TF\to\SSM\to\TF$ (\m{TST}) architectures are resource optimal. The task for \m{TST} supplies a natural task that requires layer alternation in a hybrid design.
For these tasks, \textbf{we show a separation in learnability between different layer orderings}. 
By modifying the context window for the Transformers and varying the state in Mamba, we constrain the memory of these hybrid under some fixed budget. 
We empirically show that the learned hybrids exhibit identical separations as predicted theoretically. 
For segmented retrieval (designed for the \m{TS} architectures) and the \m{TST} task, we observe a phenomenon crucial for large-scale design: on hybrids of depth 4 and depth 5, \emph{containing} \m{TS} or \m{TST} as a sub-architecture determines performance. This suggests that such ``containment'' is important in deep hybrids.

We empirically study such deeper hybrids as well. We find that a $\mm{T}^*\mm{S}^*$ architecture achieves higher performance than all $\mm{S}^*\mm{T}^*$ architectures for this task.  Mechanistically, through linear probing, we observe that the top-performing model, \m{TTSSSSSS}, shows a slow increase in decodability in its $\SSM$ layers which are not seen in \m{TTTSSSSS}. This supports that hybrid design is nuanced, as slightly different architectures fail to use their $\SSM$ layers to improve performance.

The paper is structured as follows. In Section 2, we begin with proving that there are tasks and resource constraints for any hybrid architecture that any other architecture must violate. For shallow hybrids, we establish a natural task family capturing important operations in natural language. In Section 3, we experiment on these natural tasks, identifying that these separations appear for learned hybrids. Lastly, in Section 4, we experiment on 8-layer hybrids on associative recall, observing that mildly different architectures have drastically different performances.

\section{Theoretical Foundations for Hybrid Layer Ordering}\label{sec theory}

In this section, we introduce our main theorem on model superiority, where we show that for each architecture, there is a task on which it is dominant. We begin with a summary of notation in Appendix~\ref{app:notation} before discussing the main theorem separating architectures of differing layer orderings.

\noindent \textbf{Preliminaries and Notation.}
We consider sequence-to-sequence token prediction problems. Let $\mathcal{V}$ be some vocabulary of tokens, $V = \card{\mathcal{V}}$, and $\vec{\x}=(\x_i)_{i=1}^L$ be an input sequence. A language model $M$ a is sequence-to-sequence map $M: \mathcal{V}^L \to \mathcal{V}^m$ of the form $M(\vec{\x}) = F_N\circ F_{N-1} \circ \dots \circ F_1 (\vec{\x})$, where $\circ$ is function composition and each $F_i$ is a sequence-to-sequence map called a \emph{layer}. 
Through the paper, we will consider the input and output of $F_i$ at each position to be in $\{0,1\}^P$, for some $P \in \mathbf{N}_+$, where we call $P$ the \textit{precision} of the model.
We will consider two types of layers as abstractions of state space models $(\SSM)$ and causal windowed attention $(\TF)$.
The embedded input sequence $\vec{\x}=(\x_1,\dots,\x_L)$ uses $P$ precision, i.e 
$\vec{\x}$ is such that $\x_i \in \{0,1\}^P$. 

\noindent \textbf{Transformer Layer.}
We say a layer $F: (\{0,1\}^P)^L \to (\{0,1\}^P)^L$ is a $\mm{T}$-layer $(\TF)$ with window size $W$ if there exists a map $f$ such that for $i \in [L]$, $F(\vec{\x})_i = f(\x_{i-W+1},\dots,\x_{i}) \in \{0,1\}^P$. In other words, a $\mm{T}$-layer can be an arbitrary $P$-precision causal nonlinear function, whose output depends on the past $W$ positions of the input.
Standard transformer implementations with the windowed attention mechanism falls into this category (See Appendix~\ref{app:notation} for formal definitions).  

\noindent \textbf{State-Space Model Layer.}
We say a layer $F: (\{0,1\}^P)^L \to (\{0,1\}^P)^L$ is an $\mm{S}$-layer $(\SSM)$ with bit complexity $B$ if there exists a finite set $\mathcal{S}$ of size at most $2^B$, an update rule $u: \mathcal{S} \times \{0,1\}^P \to \mathcal{S}$ and an output function $r: \mathcal{S} \to \{0,1\}^P$ such that $S_t = u(S_{t-1}, \vec{\x}_t)$ and $ F(\vec{\x})_t = r(S_t)$ hold.

These definitions capture the core difference between these layers: Transformers perfectly recall their recent context, while SSMs store a compressed version of the context \citep{pmlr-v288-nunez25a}.

We define $\Sigma:=\{\mm{S},\mm{T}\}$. An architecture profile $A$ is a finite word $A=A_1A_2\cdots A_D \in \Sigma^*$, whose depth is $|A|:=D$. For two architecture profiles $A, B$, we write $A\preceq B$ if $A$ is a subsequence of $B$: there exist indices $1\le i_1<\dots<i_{|A|}\le |B|$ such that $\forall j \in [|A|]$, $A_j = B_{i_j}$.
Notice that every nonempty architecture profile $A$ can be written as $A = X_1^{r_1}X_2^{r_2}\dots X_k^{r_k}$, with $\sum_{i=1}^k r_i = |A|$ and $r_i\in \mathbf{N}_{+}$. Here, $X_i \in \Sigma$ and $X_{i} \neq X_{i+1}$ for $i\in [k-1]$. The architecture $X_i^{r_i}$ is comprised of $r_i$ layers of type $X_i$ without a gap. We define $\mathrm{comp}(A): = X_1X_2\cdots X_k$ as the \textit{compression} of $A$.
Let $B, P, W \in \mathbf{N}_+$, and denote $\mathcal M_A(B, P, W)$ for the class of hybrid models with profile $A$, such that every $\mm{S}$-layer has a state space with size at most $2^B$, every $\mm{T}$-layer has a window size at most $W$, and each intermediate tokens between layers belong to a token space with size at most $2^P$.

The empty architecture represents the identity sequence map. Note that since both an $\mm{S}$-layer and a $\mm{T}$-layer contain the identity map, when $A \preceq B$, a model $M \in \mathcal M_A(B, P, W)$ can be simulated by a model $M' \in \mathcal M_B(B, P, W)$. In particular, if $|{A}| = |{B}|$, then $A \preceq B$ forces $A = B$.

\subsection{Expressivity and Efficiency Tradeoff under Budget Constraints}
Our method to evaluate the choice of different layer orderings is to understand, under a given budget $B, P, W$ on the state size, precision, and the window length, whether for a given task there exists a hybrid model $M \in \mathcal{M}_A(B, P, W)$ that can solve the task with good accuracy. 
If there exists a model $M \in \mathcal{M}_{\tau}(B, P, W)$ that solves some given task perfectly, while a model with the same depth but with a different profile must require some computational budgets far beyond $B, P, W$, then $\tau$ is called the \textbf{dominant choice of ordering} for this task.
Our main theoretical result reveals task-dependence in the choice of hybrid layer ordering, formalized in Theorem~\ref{th equal-depth-general-profile}. 
\begin{theorem}
\label{th equal-depth-general-profile}
Let $D \in \mathbf{N}_+$ and $\tau =
    X_1^{r_1}X_2^{r_2}\cdots X_h^{r_h} \in \Sigma^*$ be an arbitrary architecture profile with depth $D$. Let $\alpha:=\mathrm{comp}(\tau)$ and $\epsilon \in (0,1)$ be an arbitrary fixed constant.
    There exist constants $ c_\tau>0,
    C_{\tau,\epsilon}>0,$ a family of task $\{F_{\tau,\epsilon,m}\}_{m\ge 1}$, where $F_{\tau,\eps,m} : (\mathcal{V}_{\eps,m})^{L_{\eps,m}} \to \mathcal{V}_{\eps,m}$ and $\card{\mathcal{V}_{\eps,m}} \le \mathrm{poly}_{\eps}(m), L_{\eps,m} \le \mathrm{poly}_{\eps}(m)$, and a family of distributions $\mu_m$ over a sequence of tokens with length $L_m$ parameterized by $m \in \mathbf{N}_+$, such that for all $m$ sufficiently large, the follows hold.
    \begin{enumerate}
        \item (Completeness)
        There exists a hybrid model $M_{\tau,m} \in \mathcal{M}_\tau(B_{\tau,m},P_{\tau,m},W_{\tau,m})$ such that $\mathrm{err}_{\mu_m}(M_{\tau,m},
        F_{\tau,\varepsilon,m}) = 0$ and $B_{\tau,m}
        \le C_{\tau,\epsilon}\log m, P_{\tau,m}
        \le C_{\tau,\epsilon}\log m, W_{\tau,m}
        \le m+1$.

        \item (Soundness for wrong ordering) Let $A \in \Sigma^{\le D}$ be an architecture profile with $\alpha \not\preceq A$. Then for every hybrid model $M\in\mathcal M_A(B,P,W)$ that satisfies $\mathrm{err}(M,F_{\tau,\epsilon,m}) \le O(1/D)$, must satisfy $W \ge c_\tau m^{1/\epsilon}$ or $\max\{B,P\} \ge c_\tau m$.

\item (Soundness for insufficient profile) Let $A \in \Sigma^{\le D}$ be an architecture profile with $\alpha \preceq A$ but $\tau \not \preceq A$. Then, for every hybrid model $M \in \mathcal{M}_A(B,P,W)$ that satisfies $\mathrm{err}(M,F_{\tau,\epsilon,m}) \le O(1/D)$, must satisfy $W > \lambda_{\tau} (m+1)$ or $B > \lambda_{\tau} B_{\tau,m}$ or $P>\lambda_\tau P_{\tau,m}$, where $\lambda_{\tau}$ is a constant that is strictly larger than 1.
    \end{enumerate}
\end{theorem}
We give an interpretation of Theorem~\ref{th equal-depth-general-profile}. Given any target profile $\tau \in \Sigma^D$, we would like to evaluate models over a family of tasks $\{F_{\tau,\eps,m}\}$, where the size of the input of $F_{\tau,\eps,m}$ grows at an order of $\mathrm{poly}_{\eps}(m)$, for every fixed $\eps \in (0,1)$. 
In particular, with a resource budget $B_{\tau,m}, P_{\tau,m} = O_\eps(\log m)$ and $W_{\tau,m} = m+1$, a hybrid model in $\mathcal{M}_\tau(B_{\tau,m}, P_{\tau,m},W_{\tau,m})$ can perfectly solve $F_{\tau,\eps,m}$.
On the other hand, we design a distribution $\mu_{\eps,m}$ of inputs such that every model $M \in \mathcal{M}_A(B, P, W)$ with the same depth and $\tau \not \preceq A$ must require more resources to achieve a reasonable accuracy.

To quantify the resource separation, we partition the $A \in \Sigma^{\le D}$ into 2 categories: \textbf{(i)} architectures satisfying $\mathrm{comp}(\tau) \not \preceq A$, where $A$ is still not more expressive than $\tau$ with all neighboring layers merged, and \textbf{(ii)} architectures satisfying $\mathrm{comp}(\tau) \preceq A$. We say that $A$ has the \emph{wrong order} when $\mathrm{comp}(\tau) \not \preceq A$, which, by Theorem~\ref{th equal-depth-general-profile}, requires qualitatively more resources to solve the task. In particular, a model $A$ that can solve the task either requires \emph{exponentially} larger state spaces/precisions or needs a polynomially larger window size, but with an \emph{arbitrarily large rate of growth}.
When $\mathrm{comp}(\tau) \preceq A$ but $\tau \not\preceq A$, a qualitative separation is no longer possible.
This is due to our result in  Appendix~\ref{app deeper model theory} showing that consecutive layers of the same type can be merged. However, we show that the resources required by such a profile are still a constant factor larger than that of $\tau$.

We next provide an overview of why Theorem~\ref{th equal-depth-general-profile} is expected, and the technique we use to prove it. We defer the full technical details of Theorem~\ref{th equal-depth-general-profile} to Appendix~\ref{app deeper model theory}.

\noindent \textbf{Warm Up: Examples for Shallow Hybrid Models.}
\label{sec:warm_start}
To start with, we consider a simple associative recall \citep{arora2024zoology} variant, where we show a $\mm{TS}$ profile is the dominant architecture. Consider an input sequence $I$ of the form $(\x_1,\dots,\x_{m-1},q,\vec 0)$ with length $L$, where $q \in [m-1]$ and $\x_i$ is drawn from some finite vocabulary of size $\mathrm{poly}(m)$. The task input is $I$, and the output has $\x_q$ at the last position.
The $\mm{T}$-layer can perform local computations within a window of $m$, so it can output $\x_q$ at position $m$.
Following, when the $\mm{S}$-layer receives $\x_q$ at position $m$, it only needs $O(\log m)$ to memorize it until the end of the sequence. This immediately gives a model $M \in \mathcal{M}_{\mm{TS}}(\log m, \log m, m)$ that solves the task.

In contrast, the $\mm{ST}$ architecture requires at least one constraint to be significantly larger. Assuming $W\leq m$, standard communication complexity \citep{roughgarden2016communication} implies that using a $\mm{T}$-layer as the last layer requires $\Omega(m)$ bits from early in the context. Thus, the $\mm{S}$-layer must store $\Omega(m) \gg \log m$ bits to compress sufficient information to perform the task. This establishes a simple example where $\mm{TS}$ is the dominant architecture. We provide an additional broad family of natural tasks in Section~\ref{sec:natural_tasks} where $\mm{TS}$ is the dominant architecture, solvable using standard implementations of attention with Mamba-formed SSMs.

The most natural, yet flawed, task for a deeper target architecture $\tau$ is a task that iteratively solves the recall problem, separated by long $0$ vectors. For the $\tau=\mm{TST}$ architecture, such a task does separate architectures: this simple \textbf{tensor look-up} task makes $\tau$ the dominant choice and is studied empirically in Section~\ref{sec experiment}. 
Specifically, we are given input $I = (\vec x, q_1, \vec 0, \vec F, q_2)$, where each $\vec x \in [\sqrt{m}]^{m-1}$, $q_1 \in [m-1]$, $\vec F \in \mathcal{V}^m$ is a $\sqrt{m} \times \sqrt{m}$ matrix, $q_2 \in [\sqrt{m}]$, we are asked to output $F_{\x_{q_1},q_2}$. The intuition is clean: a $\mm{T}$-layer extracts $\x_{q_1}$ and sends it through the $\mm{S}$-layer to the second $\mm{T}$-layer to complete the lookup task. 

However, 
for $\tau=\mm{STS}$, \textit{a similar idea no longer works}. For intuition, we modify the tensor lookup task with the input $I = (q_1, \vec 0, \vec F, q_2,\vec 0)$ and ask a model to output $F_{q_1,q_2}$ at the end. Although a small model with an $\mm{STS}$ profile can solve the task, we find that a small model with a $\mm{TS}$ profile can also solve the task. 
The architecture uses a window of a $\mm{T}$-layer that entirely contains $\vec F$, identical to what is used for $\mm{STS}$. This layer can copy $\vec F$ and place it after $q_2$~\citep{pmlr-v235-jelassi24a}. This implies that the $\mm{S}$-layer receives $q_1\to q_2 \to \vec F$, changing the required processing order. This requires an $\mm{S}$-layer with $O(\log m)$ bits to solve the full task, the same as for $\mm{STS}$ but with one fewer layer. A core takeaway is that each layer is not only solving some specific task but also compressing and rearranging information for future layers to function more efficiently. 
A simple task like tensor lookup is too simple to prevent this type of information aggregation from being useful, especially when comparing $\mm{T}$-layers, which are any possible causal-window nonlinear function. 
Thus, to establish Theorem~\ref{th equal-depth-general-profile}, more complicated tasks must be considered.

The task we construct is in Definition~\ref{def wrong order}.
\begin{definition}[Sequential Function Composition]\label{def wrong order}
    Let $\mathcal{X}_{key}$ be a finite space of keys and $\mathcal{C}$ be a finite space of vocabularies. Let $k \in \mathbf{N}_+$. A level-$k$ recursive chain $F_{\tau,\eps,m}$ is defined by a sequence of $k$ functions $\{H_i\}_{i=1}^k$. Specifically, for $i \in [k]$, $H_i(\z_{i-1},\x_i): \mathcal{X}_{key} \times \mathcal{C}^m \to \mathcal{X}_{key}$, where $\z_i := H_i(\z_{i-1},\x_i)$.  
    Given a sequence of input $I:=(\z_0,\Vec{0}_0,\x_1,\Vec{0}_1,\dots,\x_k,\Vec{0}_k)$ of length $L_m$, $F_{\tau,\eps,m}(I):= \z_k$, where each $\Vec{0}_i$ is a sequence of $0$ vectors with different lengths. 
\end{definition}
The task is divided into $k$ stages. In each stage, we take a previous answer $\z_{i-1}$ and fresh independent context $\x_i$ to produce the next answer, $\z_i=H_i(\z_{i-1},\x_i)$.
If we use a layer ordering that subsumes $\mathrm{comp}(\tau)$, then the role of $H_i$ is similar to the lookup table in the tensor lookup task, where a $\mm{T}$-layer computes $\z_i$ and the following $\mm{S}$-layer sends it to the next stage after the long zero vectors. Thus, a model with profile $\tau$ can use a small resource budget to solve the task.
However, when we analyze a profile $A$ with $\alpha \not\preceq A$, motivated by the failure of the tensor lookup task discussed earlier, 
we construct a discrete function $H_i$ that defeats any other information embedding.

For some intuition, consider $\Bar{\alpha}$ starting with $\mm{S}$ as an example. We group $\Bar{\alpha}$ into $(\mm{S}_i,\mm{T}_i)_{i=1}^k$, where $\mm{S}_i(\text{resp. }\mm{T}_i)$ is the $i$-th $\mm{S}(\text{resp. }\mm{T})$ layer.
At stage $t+1$, with the new input block $\x_{i+1}$, the $\vec 0$ prevents $T_{i}$ from directly accessing information from the previous block.
Let $C_i$ denote the past states of the first $i$ $\SSM$ layers before $\x_{i+1}$ and let $\gamma$ denote the the state of $\mm{S}_{i+1}$ right before reading $\x_{i+1}$. Fixing $C_i$ to its current value value $c$, the future output of $\mm{S}_{i+1}$ is a deterministic function completely determined by $\gamma$ and $\x_{i+1}$ and not previous blocks, i.e. $C_{i+1} = G_{c}(\gamma,\x_{i+1})$. We refer to Lemma~\ref{lm state dependence} and Lemma~\ref{lm state representation} for the detailed analysis.

Importantly, since models need not compute $\z_i$ directly to succeed at the task, given the current state record $C_{i}$, we will recursively construct a list of possible $L_i(c) \subseteq \mathcal{X}_{key}$ such that only when $\z_i \in L_i(c)$ can the model succeed with a good probability (Lemma~\ref{lm error analysis}). 
The list $L_i(c)$ contains all possible $\z$ such that for some state $\gamma$, $\z_{i+1}$ has a large probability to fall into the next list $L_{i+1}(C_{i+1})$.
Lemma~\ref{lm H construction} and Lemma~\ref{lm chain definition} use a probabilistic method to show that there exists an $H_i$ such that each of the $L_i(c)$ cannot have a large size, which means $\z_1$ has a small probability of falling into the list; thus, there exists an $H_i$ used for the separating task. Using $H_i$, the competitive model with a wrong ordering will not succeed with a good probability unless it breaks some resource budget.

\noindent \textbf{A Size-2 Cover Method for Deeper Hybrid Model with Insufficient Layers.}
We next give a brief overview of the case where $\mathrm{comp}(\tau) \preceq A$. We defer to Appendix~\ref{app insufficient} for full technical details. As discussed earlier, in this case, a qualitative separation is not possible. Instead, we construct tasks and carefully analyze the bit complexity of solving them. In our construction, we use variants of computing the inner product over a binary field.
The main obstacle is that unlike the case with wrong orderings, there may be too many different competitive profiles to analyze.
Our analysis uses a novel size-2 cover approach to narrow the structures needed to analyze.

\begin{definition}\label{def size-2 cover}
    Let $\tau \in \Sigma^D$. Let $s(\tau) = |\{i \in [D] \mid \tau_i = \mm{S}\}| $ and  $t(\tau) = |\{i \in [D] \mid \tau_i = \mm{T}\}| $. For $i \in [s(\tau)]$, we use $r_i(\tau)$ to denote the number of $\mm{T}$-layers in $\tau$ that is strictly after the $i$-th $\mm{S}$-layer and for $i \in [t(\tau)]$,  we use $l_i(\tau)$ to denote the number of $\mm{S}$-layers in $\tau$ that is strictly after the $i$-th $\mm{T}$-layer. We call the following set of architecture profiles the \textbf{size-2 cover} of $\tau$.
    \begin{align*}
        \mathcal{C}_2(\tau) = \{\mm{S}^{s(\tau)}\} \cup \{\mm{T}^{t(\tau)}\} \cup \{\mm{S}^i\mm{T}^{r_i(\tau)}\mid i \in [s(\tau)]\} \cup \{\mm{T}^i\mm{S}^{l_i(\tau)} \mid i \in [t(\tau)]\}
    \end{align*}
\end{definition}
Roughly speaking, the size-2 cover of $\tau$ contains maximal profiles, whose compressed version has length at most $2$, subsumed by $\tau$. As we show in Lemma~\ref{lm size-2 cover} that if $\tau\not\preceq A$ and $A \in \Sigma^{\le D}$, then there must be some profile in $\mathcal{C}_2(\tau)$ not subsumed by $A$. Thus,\textit{ we only need to design hard tasks for profiles of the form in a size-2 cover}; a mixture of these creates a single task that is hard for $\tau \not\preceq A$.
However, although for each profile in $C_2(\tau)$ there exists a model with profile $\tau$ that can solve it exactly, it is not immediate whether there is a model with profile $\tau$ that can solve the mixture of these tasks exactly. Since $\mm{S}$-layers and $\mm{T}$-layers contain more general SSMs and causal windowed functions, we show in Lemma~\ref{lm size-2 reduction} that there exists a ``meta'' model with profile $\tau$ that can combine these submodels in a black-box way to solve the mixture of tasks.

\subsection{Natural Tasks for 2-layer Hybrid Models}\label{sec:natural_tasks}

Although the tasks we constructed for Theorem~\ref{th equal-depth-general-profile} can be solved by a hybrid model with the target profile using small resource budgets, the tasks themselves are explicitly designed to broaden the family of models for the hardness result. These tasks may not be efficiently solvable using commonly used layer types such as Mamba, DeltaNet, or standard attention. 
Theorem~\ref{th equal-depth-general-profile} establishes
resource separations across architecture profiles.
We next illustrate the role of layer ordering through two
more natural tasks that admit a separation favoring \m{TS}.
We establish lower bounds for \m{ST} and give \m{TS} constructions
in the generalized layer model in
Appendix~\ref{app:constructions}, with full technical details in
Appendix~\ref{app natural task}.
Appendix~\ref{app:ssm_conv} accounts for accumulated convolutional
lookback in these recurrent-state lower bounds, extending their
applicability to SSMs with convolutional
front-ends, such as Mamba.

The core requirement for tasks with a dominant \m{TS} architecture is captured in Assumption~\ref{asp:U_sensitive}. This assumption states that, for each task, there is some position $\tau$ in the context such that changes before some position $\tau$ require $\log |\gQ|$ bits of information from before $\tau$, similar to the \m{TS} task from Section~\ref{sec theory}. Under this assumption, we can show that any model with $\SSM$ layers first and $\TF$ layers second requires a large state or window to perform.

\begin{theorem}[$\SSM\to\TF$ Cut]
\label{thm:hybrid_cut}
Let $F$ be a length-$L$ sequence task with output in
$\mathcal{Y}$. Suppose there exist a cut position $\tau$,
a nonempty family $\mathcal{U}$ of length-$\tau$ prefixes,
and a nonempty family $\mathcal{Q}$ of length-$(L-\tau)$ suffixes,
such that every concatenation $uq$ is a legal input and
$\forall u\neq u'\in\mathcal{U},
    \exists q\in\mathcal{Q} \
    \text{such that} \
    F(uq)\neq F(u'q).$
Let $M$ be a deterministic model of architecture \m{ST}
with recurrent state space $\gS$ and attention window $W$.
If $M$ computes $F$ correctly on all legal inputs and that 
$\tau\leq L-W$, then
$|\gS|\geq|\mathcal{U}|.$
\end{theorem}
This result stems from how information can be processed in a sequence. Without a large window, only the $\SSM$ in \m{ST} communicates with the early part of the sequence, forcing a large state. For a \m{TS} architecture, the Transformer performs computations that are then carried by the $\SSM$, reducing state size and window length. We now define the two natural tasks for the \m{TS} architecture.

\noindent \textbf{Exact Copy.} 
Given a sequence $\mathbf{x}$, the task of \emph{exact copy} is to output $1$ if there exists some repeated subsequence of length $l$, and output $0$ otherwise. 

\begin{definition}[Exact Copy]\label{dfn:exact_copy}
Given a vocabulary $\mathcal{V}$ and an input sequence $\x\in \mathcal{V}^L$, we say that $\x$ contains an \textit{exact copy} of repetition $c$ and string length $l$ if there exists $i\in [L]$ s.t. 
\begin{align*}
\x_{i+j+kl}=\x_{i+j+(k+1)l},\quad j\in \{0,\cdots,l-1\}, k\in \{0,\cdots,c-2\}.
\end{align*}
The task of \textit{Exact Copy} is to output $y_L$, where $y_i:=\mathbf{1}_{\{\x_{1:i} \ \text{has an exact copy}\}}$.
\end{definition}

This task satisfies Assumption \ref{asp:U_sensitive} by setting $\tau$ to be the end of the exact copy. Thus, Theorem \ref{thm:hybrid_cut} applies, ensuring \m{S}${}^*$\m{T}${}^*$ architectures require large memory. This task represents situations where a complex operation occurs early in a sequence, but the result needs to be carried much later in the context. For instance, reasoning-style tasks where a model identifies a useful lemma using $\TF$ layers early in the sequence can be efficiently propagated to the end of the sequence by $\SSM$ layers.

\noindent \textbf{Segmented Retrieval.} 
To define segmented retrieval, we place separator tokens in the sequence to act as a partition. The simplest per-segment operation is a fixed lookup in the past.  Concretely, this means that for some fixed $\lambda \geq 1$, the value at the separator token at position $i$ is $x_{i-\lambda}$. The task of \emph{segmented retrieval} is to output the last of these $x_{i-\lambda}$ in some subset of the vocabulary.

\begin{definition}[Segmented Retrieval]\label{dfn:seg_ret}
Given a vocabulary $\mathcal{V}$, we specify a non-empty subset $\mathcal{N}\subseteq\mathcal{V}$. Given an input sequence $\left(\left(\mathbf{x}_1,b_1\right),\ldots,\left(\mathbf{x}_L,b_L\right)\right)
\in (\mathcal{V}\times\{0,1\})^L$, where $b_i$ indicates whether position $i$ is an internal segment
boundary and is revealed only when position $i$ is
processed. We define $\{t_1,\cdots, t_{k-1}\}:=\{i\in [L-1]:b_i=1\},t_0:=0, t_k:=L$ and refer an element from $\{t_1,\cdots,t_{k-1}\}$ as a \textit{boundary} and define $ l:=\min_{j\in [k]}(t_j-t_{j-1})$ and we set that $l\ge 2$. Select $1\leq \lambda < l$, and define the largest index of the nonempty set $\{j\in [k-1]:\mathbf{x}_{t_j-\lambda}\in \mathcal{N} \}$ as $\lambda^{\star}$. The \textit{Segmented Retrieval} task is to output the token $\mathbf{x}_{t_{\lambda^{\star}}-\lambda}.$ 
\end{definition}

Similar to exact copy, this task satisfies Assumption \ref{asp:U_sensitive}. Again, Theorem \ref{thm:hybrid_cut} applies, ensuring \m{S}${}^*$\m{T}${}^*$ architectures require large memory. However, the source of the cut $\tau$ is different. Here, the cut occurs at the partitions, where the model must recall the past $\lambda$ tokens. The \m{TS} architecture uses its $\TF$ layer to perform the $\lambda$-back recall, and the $\SSM$ carries the relevant tokens through the sequence. This is a simplified multi-document task, where information from each document (segment) may need to be carried to the end of the context; this latter component is best suited for $\SSM$ layers.

\section{Empirical Results of Natural Tasks}\label{sec experiment}

We have established that for certain tasks, there is a theoretical separation in expressivity. However, these theoretical results focus on layerwise resource constraints that may not tie to common constraints in practice. We empirically test small hybrids on these tasks with a total fixed inference memory, showing that hybrids of this specific architecture learn them with the best performance.

\subsection{Two Layers - Segmented Retrieval and Exact Copy}
\label{sec:two_layers_tasks}
\begin{figure}[t]
    \centering
    \begin{subfigure}[b]{0.48\textwidth}
        \centering
        \includegraphics[width=\linewidth]{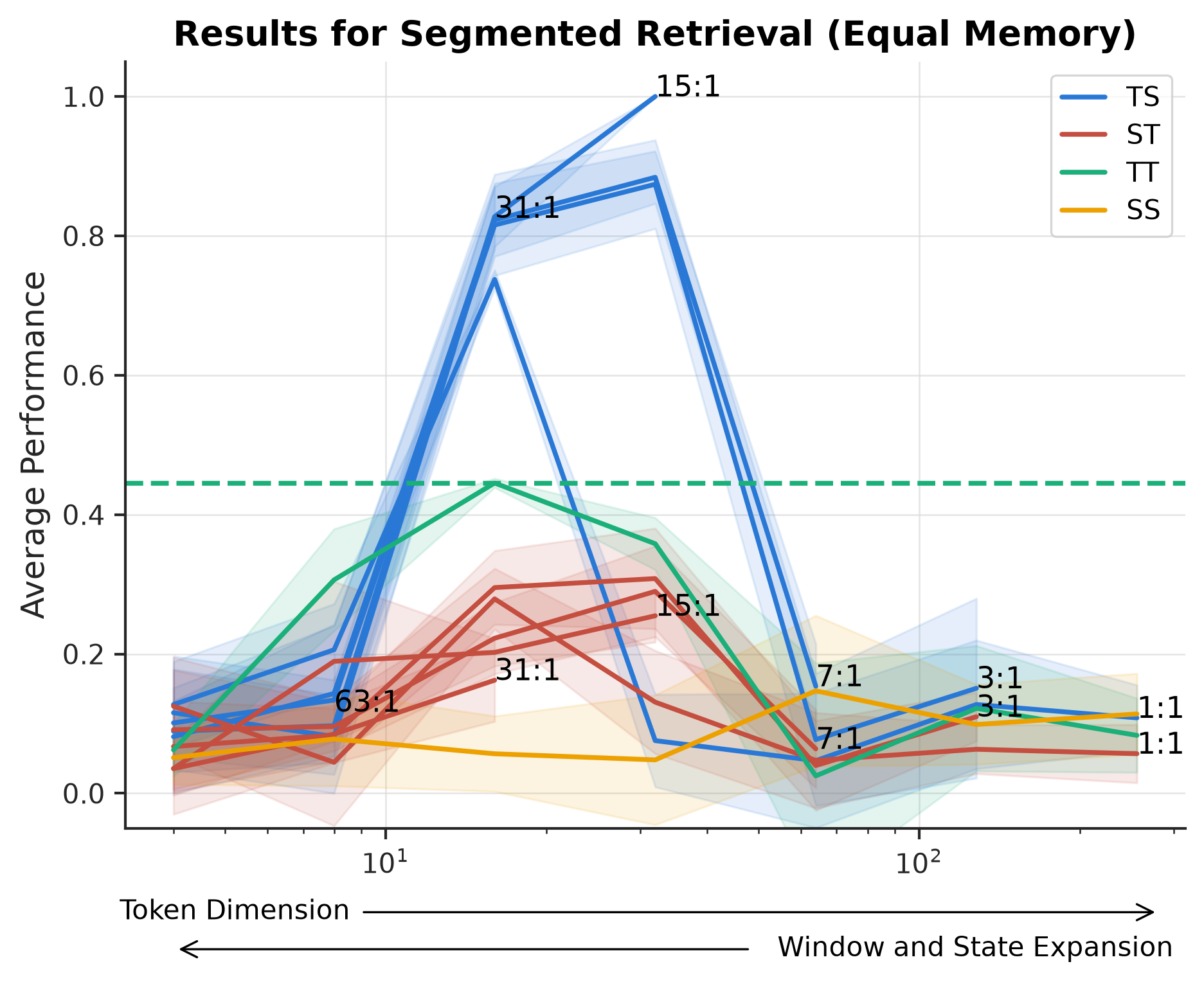}
        \caption{The Segmented Retrieval task.}
        \label{fig:seg_retrieve}
    \end{subfigure}
    \hfill 
    \begin{subfigure}[b]{0.48\textwidth}
        \centering
        \includegraphics[width=\linewidth]{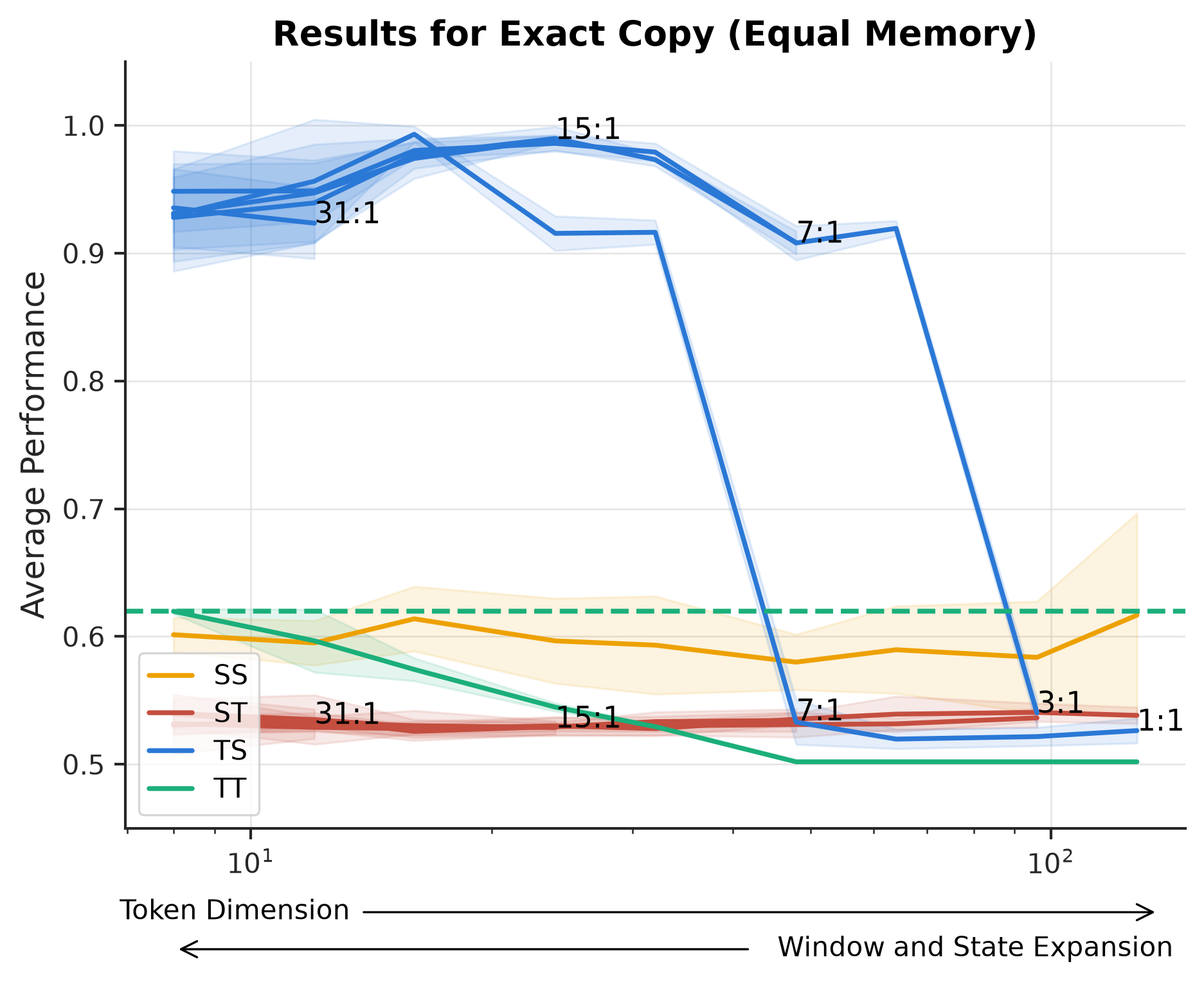}
        \caption{The Exact Copy task.}
        \label{fig:exact_copy_results}
    \end{subfigure}
    \caption{Results for the \m{TS} tasks where an inference memory proxy is held constant. The annotations represent the relationship between window storage and state size for Transformer-Mamba hybrids. The only architecture with high performance on the task is the \m{TS} architecture. Due to fixed memory, models with larger token dimension have smaller window and state size, making the tasks difficult. Conversely, small token dimension limits expressivity.}
    \label{fig:ts_task_results}
\end{figure}

We begin the experiments with a sweep over the possible 2-layer hybrids on tasks where we expect a \m{TS} to be dominant. For two tasks, segmented retrieval and exact copy, we observe that there is a similar separation between architectures as predicted theoretically. These results are conducted under a fixed proxy of inference memory (details in Appendix~\ref{app:memory_calc}).

\noindent \textbf{Setup.} On both tasks, we use a small vocabulary of around 20 tokens and sequence lengths of around 150. We select Transformers with RoPE embeddings for \m{T}-layers and Mamba for \m{S} layers~\citep{gu2023mamba}. These match the model types used in \cite{pmlr-v235-jelassi24a} to compare pure SSMs to pure Transformers. For the \m{TT} and \m{SS} architectures, a fixed memory constraint induces a trade-off between token embedding dimension and window length or state size. 
For the \m{ST} and \m{TS}, there are two degrees of freedom to satisfy the memory budget proxy. We select some window-to-state-size ratios; for each, we execute a sweep over token dimension to blanket the space of possible architectures. The memory calculation is described in Appendix \ref{app:memory_calc}. All experiments are averaged across 5 seeds.

\noindent \textbf{Results.} Results for both tasks are found in Figure~\ref{fig:ts_task_results}. These results show the performance of each hybrid, plotted against the token embedding dimension. For the \m{TS} and \m{ST} hybrids, the ratio of window to state dimension is annotated at the end of each associated curve. Each curve ends once the desired ratio is no longer attainable, i.e. the token dimension is so large that the sum of the window and state expansion factors is less than the sum of the ratio components. Moving to the right increases the token embedding size, which requires the window and state expansion factor to decrease. This allows the model to have richer token representations but reduces the available context. Moving to the left increases the window width and state expansion factor but requires a reduction in token precision. While much, if not all, of the context is available, the token embeddings are so small that the model cannot compute with each token sufficiently. The highest average performance of a model other than \m{TS} is shown by a dashed horizontal line.

Crucially, the only model that achieves high accuracy, and perfect accuracy in one setting, is the \m{TS} architecture. This result also confirms that \textbf{the layer ordering is vital to performance}. The \m{ST} architectures never achieve high performance, as their ability to capture complex operations early in the sequence is hindered by this layer ordering.

\subsection{Containing Subarchitectures Predicts Performance}
\noindent
\begin{figure}[t]
  \centering
  \noindent
  \begin{minipage}{0.48\textwidth}
    \centering
    \begin{tabular}{cc|cc}
        \toprule
        Model & Accuracy & Model & Accuracy \\
        \midrule
        \m{TTTT} & 0.871 & \m{TSSS} & 1.000 \\
        \m{STTT} & 0.706 & \m{STSS} & 1.000 \\
        \m{SSTT} & 0.516 & \m{SSTS} & 1.000 \\
        \m{SSST} & 0.345 & \m{TTSS} & 1.000 \\
        \m{SSSS} & 0.125 & \m{TSTS} & 1.000 \\
                 &       & \m{TSST} & 1.000 \\
                 &       & \m{STTS} & 1.000 \\
                 &       & \m{STST} & 1.000 \\
                 &       & \m{TSTT} & 1.000 \\
                 &       & \m{TTST} & 1.000 \\
                 &       & \m{TTTS} & 1.000 \\
        \bottomrule
    \end{tabular}
    \captionof{table}{Training 4-layer hybrids on segmented retrieval. Left columns are models without \m{TS}; the right columns are models with \m{TS} as a component. There is a clear dichotomy between hybrids with and without this motif.}
    \label{tab:four_layer_seg_ret}
  \end{minipage}
  \hfill 
  \begin{minipage}{0.48\textwidth}
    \centering
    \includegraphics[width=\linewidth]{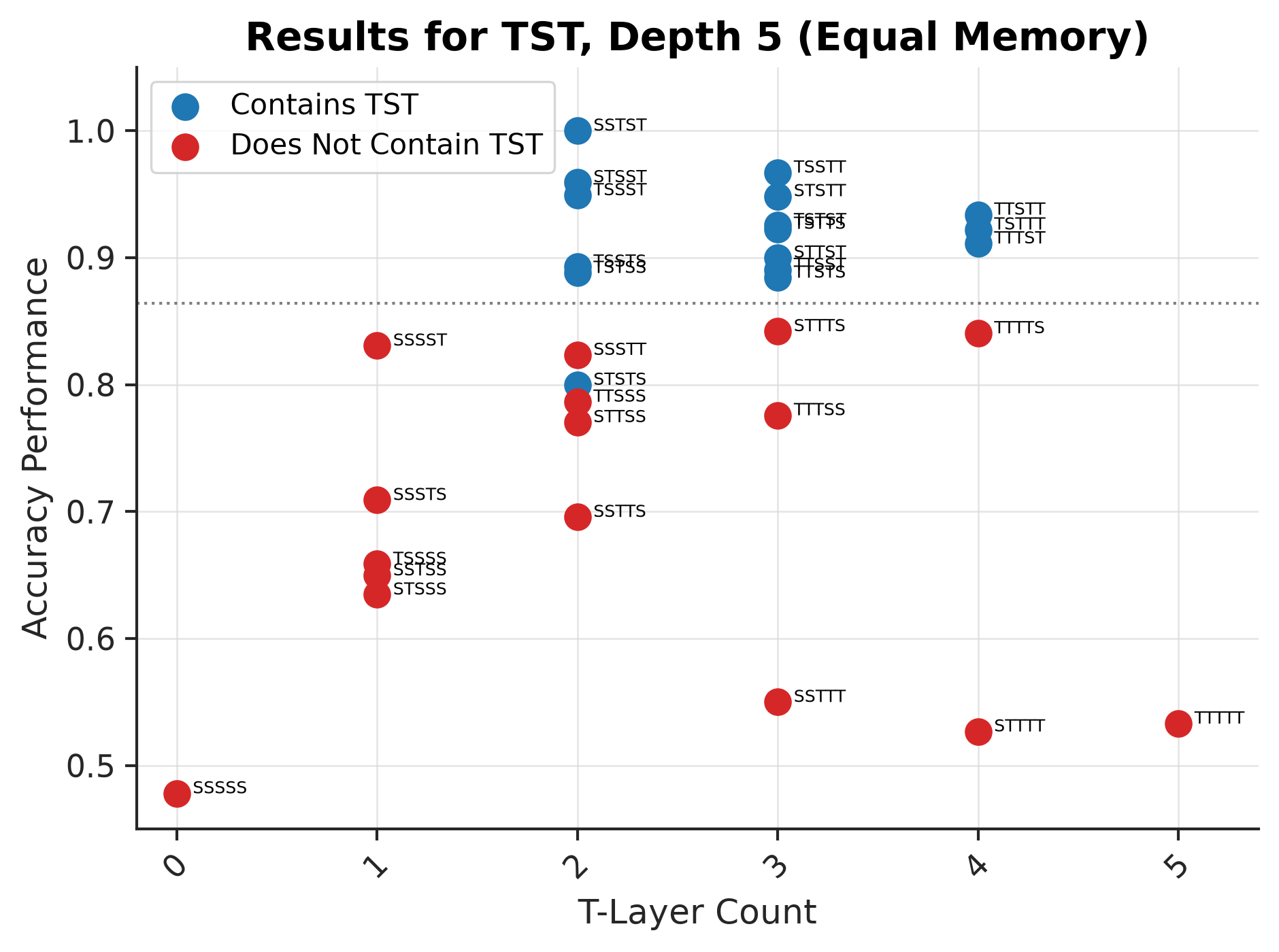} 
    \captionof{figure}{Performance of all possible 5-layer Transformer-Mamba hybrids on the task where \m{TST} is expected to dominate. Models with \m{TST} as a sub-architecture perform above those without.}
    \label{fig:tst_5_layer}
  \end{minipage}
\end{figure}
The two-layer experiments in Section \ref{sec:two_layers_tasks} and the three-layer
experiments in Appendix \ref{app:tensor_lookup}  identify \m{TS} and \m{TST},
respectively, as the best-performing profiles under the tested
settings.
Motivated by our theoretical separations, we next examine whether
containing these profiles as subsequences predicts performance
at greater depth.
We evaluate all four-layer profiles on segmented retrieval
(Table \ref{tab:four_layer_seg_ret}) and all five-layer profiles on tensor lookup
(Figure \ref{fig:tst_5_layer}).

\noindent \textbf{Setup.} For each experiment, we follow the setup for the two- and three-layer architectures. We increase the total allotted memory to account for the models having more layers. For both settings, we test all layer orderings of Transformers and SSM layers. We repeat across 3 seeds for each. 

\noindent \textbf{Results.} The results for training 4-layer hybrids on segmented retrieval are found in Table \ref{tab:four_layer_seg_ret}. This table separates architectures that contain \m{TS} as an architecture and those that do not. Accuracy is taken as the highest across seeds for models without \m{TS} and the lowest across seeds for models with \m{TS}. Even so, \textbf{all architectures with \m{TS} as a sub-architecture achieve perfect performance, while all architectures without \m{TS} do not achieve perfect performance} on any of the 3 seeds.

The results for the \m{TST} task are shown in Figure \ref{fig:tst_5_layer}. The models are positioned based on their number of Transformer layers and are colored based on their containment of \m{TST} as a sub-architecture. Aside from the \m{STSTS} architecture, models with \m{TST} are notably higher in performance than the other architectures.

\section{Observations on Associative Recall}

We have thus far focused on the accuracy of various hybrid architectures. These architectures were trained on tasks where we have clear theoretical results indicating which architecture is predicted to have the highest performance. Here, we probe other aspects of learned hybrids on a variant of associative recall \citep{arora2024zoology}, including a mechanistic study of the learned algorithms by different layer orderings.

\begin{figure}[t]
    \centering
    \begin{minipage}[t]{0.38\textwidth}
        \centering
        \includegraphics[width=\textwidth]{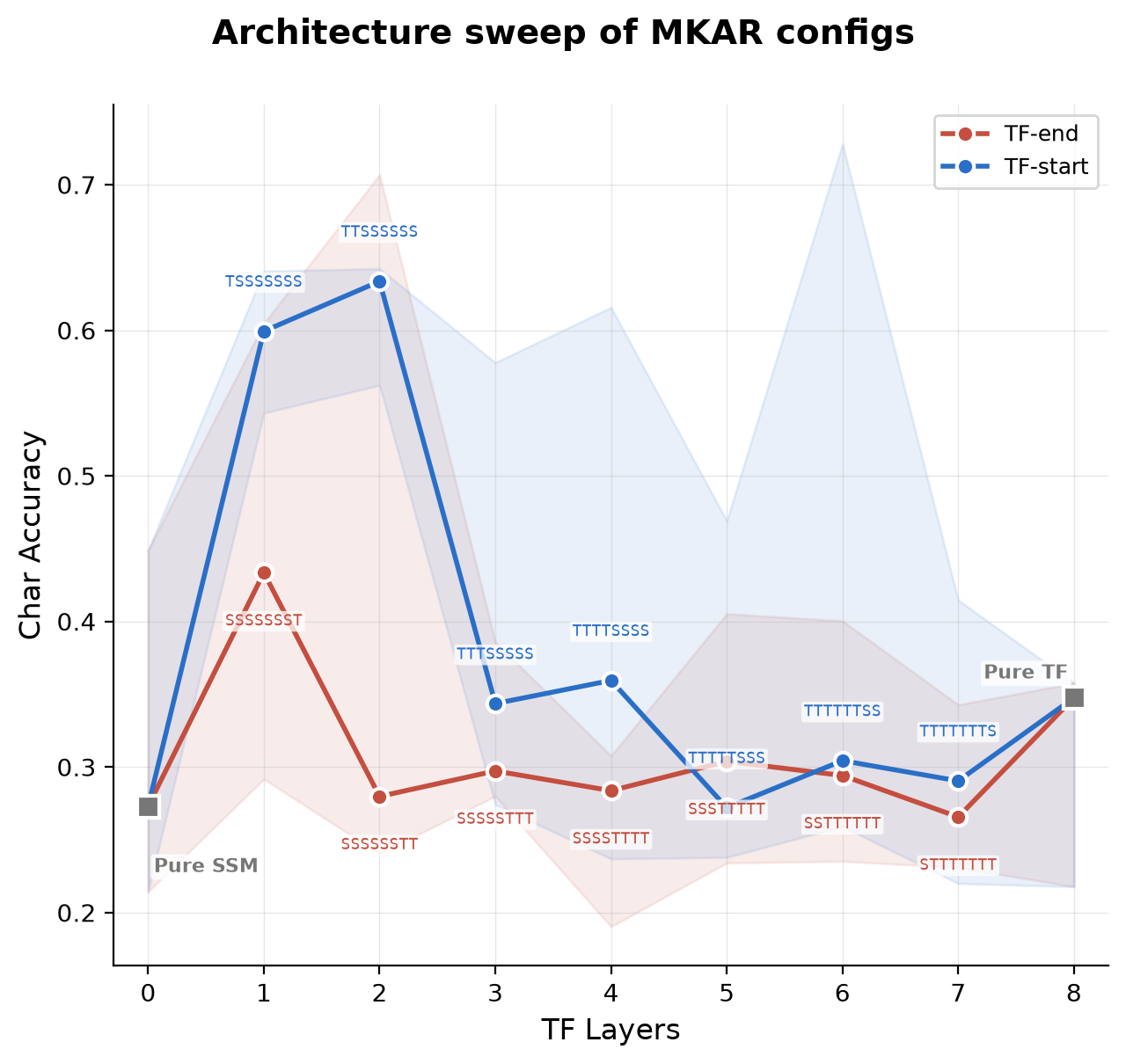}
        \captionof{figure}{Token accuracy of $\mm{T}^*\mm{S}^*$ and $\mm{S}^*\mm{T}^*$ architectures on $(2,1)$-AR. Small differences in architectures affect performance dramatically.}
        \label{fig:sweep}
    \end{minipage}
    \hfill 
    \begin{minipage}[t]{0.58\textwidth}
        \centering
        \includegraphics[width=\linewidth]{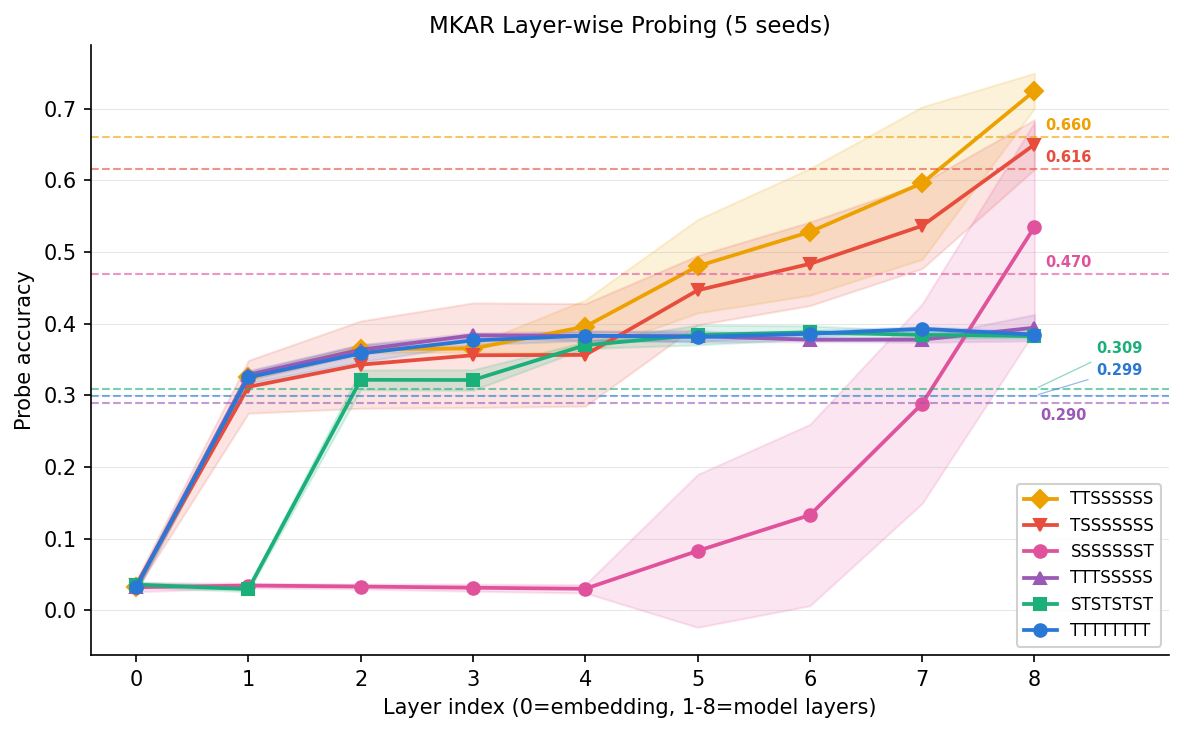}
        \captionof{figure}{Probe accuracy from the output stream at each model layer. All models have a significant gain in probe accuracy at their first TF layer, even in the pure TF where all layers are identical. \m{TTTSSSSS} and \m{STSTSTST} both see little increase at \m{S}-layers, while the other models do.}
        \label{fig:MKAR_probing}
    \end{minipage}
\end{figure}

\noindent \textbf{Setup.} We test $(2,1)$-AR (two tokens per key, one token per value) on hybrid architectures with 8 layers. This task represents a very compressed version of a multidocument task, where every triplet of tokens is a document identifier and associated value. The chosen architecture families are size-2 cover architectures $\mm{T}^*\mm{S}^*$ and $\mm{S}^*\mm{T}^*$. We vary the \m{T}-layer count between 0 and 8, repeating each setting with 5 seeds. The median, minimum, and maximum performance are reported. 
Additionally, for each architecture, we take the last token representations at each layer of the model and linearly probe for the next token. This roughly measures how much of the next token is identified by each layer.
Additional details can be found in Appendix \ref{app:synth_tasks}.

\noindent \textbf{Results.} The results on performance are shown in Figure \ref{fig:sweep}. As expected, hybrid models outperform pure architectures. However, each architecture family behaves differently, including a different peak $\TF$-to-$\SSM$ ratio. Models with Transformers in the first few layers benefit most from having a very small subset of layers as Transformer layers. We also observe that the $\mm{T}^*\mm{S}^*$ architecture outperforms the $\mm{S}^*\mm{T}^*$ architecture. We predict this is due to the multi-document nature of this task. 

The linear probe results in Figure \ref{fig:MKAR_probing} further support this. Consistently, the first $\TF$ layer improves decodability significantly. This $\TF$ layer retrieves the key-value pairs within its context window, but cannot capture the rest. More subtle is the behavior of models with many $\SSM$ layers stacked back-to-back. With these long strings of $\SSM$ layers, the model's decodability slowly increases as depth increases; we hypothesize the $\SSM$ layers are bringing these remaining key-value pairs from outside the context window. We also note that models trained on natural language typically exhibit a similar behavior, with $\TF$ layers performing the majority of recall~\citep{michalak2025some}.

\section{Conclusion}

Hybrid architectures are critical to building frontier models. Their performance and flexibility come at the cost of an expanded design space that cannot be easily navigated without an understanding of the effect of layer order and design on task performance. We provided an analysis taking the first step towards such an understanding. We showed that for each design, there exist tasks that provably favor a model with this type of architecture over all other choices. We do so by explicitly constructing such models, additionally finding that trained hybrids exhibit matching behaviors. 
Our general theory relies on deliberately constructed tasks and expressive abstractions of SSM and attention layers, and the corresponding optimal architectures need not always be realizable by standard practical models. While our natural tasks and experiments begin to bridge this gap, an important direction is to extend the theory to practical layer classes, richer mixtures of tasks, and natural-language settings. Ultimately, we hope to identify task-level properties that predict effective layer orderings before expensive model training.

\section{Acknowledgements}

John Cooper and Frederic Sala were supported by the National Science Foundation (NSF) (CCF2106707), the Defense Advanced Research Projects Agency (DARPA Young
Faculty Award) and the Wisconsin Alumni Research Foundation (WARF). Ilias Diakonikolas was supported by a Guggenheim Fellowship and the ONR (N00014-
25-1-2268). Mingchen Ma was supported by NSF Award
CCF-2144298 (CAREER) and the ONR (N00014-
25-1-2268).

\newpage

\subsection*{AI use statement}

In this work, generative AI models including GPT 5.6 sol and GPT 6 Astra are used interactively as a routine for discussion, helping with proof construction, writing code for some of the tasks as well as checking potential flaws during the preparation of the manuscript.
The analysis of Case 4 of Theorem~\ref{th insufficient layers} was mainly established and proved by GPT 6 Astra based on a hardness construction provided by the authors that contained a non-trivial mathematical flaw. We include the generated proof at \href{https://github.com/SprocketLab/layer-importance}{https://github.com/SprocketLab/layer-importance}, as supplementary material to support the correctness of the theory.
We additionally used generative AI tools to polish some grammar in the text.
We did not use generative AI tools for any other required-disclosure task, including the core writing or literature review.
We have reviewed all AI-assisted work: AI-generated proofs and code were read, executed, and verified for correctness by at least 2 authors, and all suggested grammatical edits were checked to ensure no claims or technical content were altered. 
We take responsibility for the final content of this work, including text, claims, or artifacts produced with the aid of generative AI.

\bibliography{ref}
\bibliographystyle{plain}

\pagebreak
\appendix

\etocsettocstyle{\section*{Table of Contents}}{}
\addcontentsline{toc}{chapter}{Appendix Contents}
\localtableofcontents
\pagebreak

\section{Notation}
\label{app:notation}

We provide an overview of the notation used throughout the paper in Table \ref{tab:notation}.

\begin{table}[t]
    \centering
    \begin{tabular}{c|c}
        Symbol & Meaning \\
        \hline \hline
        $\gV$ & The vocabulary space. \\
        $\gU$ & The intermediate space used between architecture layers. \\
        $\gY$ & The output space. \\
        \midrule
        \m{S} & An $\SSM$ layer. \\
        \m{T} & A $\TF$ layer. \\
        $\Sigma$ & Defined to be $\{\mm{S}, \mm{T}\}$, the possible letters in an architecture. \\
        $A,B$ & Model architectures from $\Sigma^*$. \\
        $\tau$ & The target architecture relating to Theorem~\ref{th equal-depth-general-profile}. \\
        $\mathrm{comp}(\cdot)$ & Compressing all common layers into a single layer. \\
        $\alpha$ & The compression of $\tau$. \\
        $\bar{\alpha}$ & The flipped architecture, $(\mm{S}, \mm{T}) \mapsto (\mm{T}, \mm{S})$. \\
        \midrule
        $\gS$ & The state space for the $\SSM$ layers. \\
        $W$ & The window for the $\TF$ layers. \\
        $B$ & Number of bits available to an $\SSM$ layer. \\
        $P$ & Number of bits available to the hidden space between layers. \\
        $\gM_A(B,P,W)$ & Model family of architecture $A$ under the constraints $B,P,W$. \\
        \midrule
        $d$ & The dimension of the hidden state. \\
        $N$ & The Mamba state expansion factor. \\
        \midrule
        $\x$ & An input sequence from $\gV$. \\
        $\f$ & A binary vector marking the endpoints of partitions for 2-layer tasks. \\
        $L$ & The sequence length. \\
        \midrule
        $\W_k,\W_q,\W_v$ & Attention parameter matrices. \\
        $u, r$ & Update and emission rules for the $\SSM$. \\
        \bottomrule
    \end{tabular}
    \caption{Notation}
    \label{tab:notation}
\end{table}

An attention head $\Attn$ is defined by matrices $\W_k,\W_q,\W_v \in \R^{d \times d}$ such that 
$\Attn(\vec{\x})_j = \sum_{i=1}^n \alpha_{ji} \W_v\x_i$, where
\begin{align*}
    \alpha_{ji} := \frac{\exp\left((\W_q \x_j) \cdot (\W_k \x_i)\right)}{\sum_{i=1}^n \exp\left((\W_q \x_j) \cdot (\W_k \x_i)\right)}.
\end{align*}
An attention layer $\AT$ is defined by $H$ attention heads $\Attn_1,\dots,\Attn_H$ and a projection matrix $\W_o \in \R^{d \times dH}$. Denote by $\mathbf{O}:= (\Attn_1(\vec{\x})^\top,\dots,\Attn_H(\vec{\x})^\top)^\top \in \R^{dH \times L}$ the concatenation of the outputs of the $H$ attention heads, so the attention layer $\AT(\vec{\x})$ outputs $\W_o \mathbf{O} \in \R^{d \times L}$.

A Transformer layer $\TF$ is defined by an attention layer $\AT$ and an MLP layer. In particular, an MLP layer is defined as $f(\x): \R^d \to \R^d$ as $f(\x) = \mathbf{U}_2 \sigma(\mathbf{U}_1 \x)$, where $\mathbf{U}_1,\mathbf{U}_2$ are matrices and $\sigma$ is an activation function applied coordinate-wise. Specifically, $\TF(\vec{\x}) = \MLP(\AT(\vec{\x}))$.

An SSM layer $\SSM$ is defined with a state $S_t$, an update rule $s$, and an emission rule $r$. Let $s_0 \in \mathcal{S}$ be some initial state. Given some sequence $\x_1, \ldots, \x_L$, the state of the model at iteration $i$ is denoted by $s_i = S_i(\x_1, \ldots, \x_i)$ and the output token is denoted by $r_i=F(\vec{\x})_i = R_i(\x_1, \ldots, \x_i)$, where the state and output are defined recursively:
\begin{enumerate}
    \item $S_0(\emptyset) = s_0$,
    \item $S_i(\x_1, \ldots, \x_i) = u(S_{i-1}(\x_1, \ldots, \x_{i-1}), \x_i)$,
    \item $R_i(\x_1, \ldots, \x_i) = r(S_i(\x_1, \ldots, \x_i))$.
\end{enumerate}

\section{Related Work}

\subsection{Hybrid Models}

\noindent \textbf{Empirical Analysis on Hybrid Models.}
    Hybrid designs now appear throughout frontier-scale open models, where attention placement is conventionally fixed and only the mixing ratio is tuned. Jamba \citep{lenz2025jamba} is one of the first models to combine attention, Mamba, and mixture-of-experts layers. It interleaves one attention layer for every seven Mamba layers and reports no substantial difference relative to a 1:3 ratio. Nemotron-H \citep{nvidia2025nemotronhfamilyaccurateefficient} allocates roughly 8\% of layers to attention, spacing them evenly through the stack; Nemotron~3 \citep{nvidia2025nemotron3} retains only a handful among interleaved Mamba-2 and mixture-of-experts layers; and Nemotron-Flash \citep{fu2025nemotronflash} selects its layer sequence via search under a latency budget. Kimi Linear \citep{kimiteam2025kimilinear}, Olmo Hybrid \citep{merrill2026olmohybrid}, Qwen3-Next \citep{qwen2025qwen3next}, and Qwen3.8-Next \citep{qiu2026qwen38next} all place one attention layer in every four. Within each period, attention follows the recurrent layers, except in Jamba where it appears among them.

Systematic studies explore hybrid design more broadly but still leave placement largely unexplored. \cite{waleffe2024empiricalstudymambabasedlanguage} show at the 8B-parameter scale that a hybrid with 7\% attention layers outperforms a Transformer on all twelve standard tasks. \cite{wang2026systematicanalysishybridlinear} sweep six linear-attention variants across five ratios, recommending 3:1 to 6:1 ratios of $\SSM$ to $\TF$ layers, and find that the best standalone linear model is not the best hybrid component. \cite{lee-etal-2025-understanding} compare sequential and parallel hybrids on memory recall. Only \cite{bae2026hybridarchitectureslanguagemodels} vary where attention blocks are placed: at every ratio, placing attention in the early layers underperforms pure models, whereas placing it in the middle performs best. Some works have also observed that pure Transformers learn many heads that can be simulated closely by an SSM, allowing for distillation to more efficient architectures \citep{xia2026distillthenreplaceefficienttaskspecifichybrid, bick2026retrievalaware}.

Since each configuration requires a full pre-training run, deployed models each commit to a single interleaving, and the one controlled study of placement \citep{bae2026hybridarchitectureslanguagemodels} explores only a narrow slice of the design space. Our synthetic sweeps cover a broad set of orderings within an 8-layer stack and show that the best ordering depends on the task.

\noindent \textbf{Theoretical Understanding for Hybrid Models.}
Hybrid architectures have been studied less in a theoretical setting than in empirical settings. The work by \cite{cooper2026expressivityefficiency} established a separation in memory resource constraints between hybrid architectures and pure architectures. We build on this work, studying hybrids of greater depth. Additionally, \cite{merrill2026olmohybrid} proved hybrid architectures can implement tasks that exist in NC${}^1$ while both SSMs and Transformers are stuck in TC${}^0$. This work establishes more tasks and distinguishes between layer orderings, which has yet to be studied using tools from communication complexity
Communication complexity is a classical tool for converting information-transfer
bottlenecks into computational lower bounds
\citep{kushilevitz1997communication,roughgarden2016communication}.
It has recently become useful for understanding the expressivity of sequence models.
\cite{sanford2024transformers} use communication-complexity arguments to establish
representational limitations of attention, while \cite{bhattamishra2024separations} obtain
separations between Transformers and recurrent architectures through reductions from
communication problems. Closer to large language models, \cite{peng2024limitations} use
communication complexity to prove limitations of Transformers on function composition,
and \cite{chen2025theoretical} introduce an autoregressive communication model to obtain
unconditional depth--width lower bounds for multi-layer decoder-only Transformers and
a provable advantage for chain-of-thought reasoning.

\subsection{Mechanistic Tasks}

\noindent \textbf{Associative recall.} The most common task to test a model's retrieval abilities is associative recall \citep{olsson2022context}. It probes a model's capability to maintain an associative memory, keeping track of facts indexed by prior tokens. The original form of associative recall separated pure SSMs from pure TFs as worse at storing large memories. 

This task, however, has grown too simple to the point where many similar architectures perform nearly perfectly. A few variants have been proposed, such as MQAR \citep{arora2024zoology}--eliminating issues from only querying a single key-value pair--and phonebook \citep{pmlr-v235-jelassi24a}--where each key and value are made of multiple tokens. We focus on the last of these as a probe into the differences between hybrid architectures.

\section{Experiment Details}

We provide some additional experimental details for those conducted in the main body. We begin with a description of the layer architectures used. In Section~\ref{sec:app_shallow_details}, we provide the experimental details omitted in the main body. In Section\ref{sec:app_lr_selection}, we detail the learning rate sweeps conducted for each of the experiments. Lastly, in Section~\ref{app:memory_calc}, we describe the calculation used to determine the size of memory used by these models.

\paragraph{Transformers.} The Transformers come from GPTNeoX and are given RoPE positional embeddings, both in the pure model and in all hybrids. All experiments use Transformer layers with a single head each. We observed that for this scale of experiments, increasing the number of heads by partitioning the square key and query matrices into thin rectangles hindered performance severely. 

\paragraph{SSMs.} The chosen SSM for all experiments is Mamba. This architecture allows for the state size to be increased independently of token dimension. 

\subsection{Two- and Three- Layer Hybrid Details}
\label{sec:app_shallow_details}

Here, we detail the different experimental parameters relevant to each of the tasks tested.

\paragraph{Segmented Retrieval.} We use a vocabulary of 40, with half of the vocabulary within the set $\gN$. We use sequences of length 150 and a partition of 10 segments. The segment boundaries are chosen uniformly at random in a sequence. We use a lookup distance of $\lambda=8$, halfway to the expected segment length of 15. This prevents too many sequences from being so short as to not have any valid lookup, nor is $\lambda$ so small that it falls within Mamba's learned convolution. Using a uniform sampling of positions makes the task easier than if all segments were of length $\lambda$ or greater. In fact, we observed that when $\lambda$ is within this convolution length, the pure $\SSM$ architectures perform the task nearly as well as \m{TS}. It is possible to modify this task so the lookup distance is not fixed, though, circumventing the convolution altogether. Each model has 512 dimensions of memory. Due to the output space containing 20 tokens from the vocabulary, random performance is 0.05.

\paragraph{Exact Copy.} For sequence lengths of 100, we have 20 vocabulary tokens. The copied sequence length is 5, which is repeated 3 times in the relevant sequences. To assist with training, we train the model to output 1s at every position after the exact copy is observed rather than only at the last token. Each model has 512 dimensions of memory. Due to the task being binary, random performance is 0.5.

\paragraph{\m{TST} Task.} We use a sequence length of 150 and a vocabulary size of 7. The selection of 7 puts the final matrix in the last 1/3 of the context, and the initial lookup lies within the initial 1/3 of the context. This leaves a comfortable 1/3 of the middle of the context as null tokens, requiring an $\SSM$ layer to prevent large memory usage. We use a special copy of the vocabulary for the query token $q$ to perform the first lookup. This helped the task be learnable by separating the tokens that are within the tensors and those which act as queries.

\subsection{Learning Rate Selection}
\label{sec:app_lr_selection}

In all experiments, we conduct a learning rate sweep to identify the best learning rate for each set of hyperparameters. For the segmented retrieval and \m{TST} tasks, 9 learning rates are sampled multiplicatively evenly between $10^{-4}$ and $10^{-3}$. Each learning rate is run for one epoch and for 2 different sets. The learning rate with the smallest loss is selected as the learning rate for the 5 seeds tested.

For the exact copy experiments, 4 learning rates are multiplicatively spaced between $10^{-3}$ and $10^{-1}$. For each of these, all 5 seeds are run for complete training runs. The best learning rate on average was selected. These experiments are thus more computationally expensive than the other tasks, although this worked in this setting, as we experimented on the other two tasks with deeper hybrids with more settings.

The learning rates for associative recall were selected with 2 epochs of training and a collection of learning rates between $10^{-5}$ and $10^{-2}$. More details are provided in Appendix \ref{app:lr_sensitivity}.

\subsection{Memory Proxy Calculation}
\label{app:memory_calc}

We calculate a proxy for the inference memory used, $M$, by each model in a simplified manner as follows. Let $d$ be the hidden dimension of the model.

For a Transformer with window $W$, we account for the memory needed for the Transformer to store the tokens in its context. Thus, we calculate the memory needed as
$$M_{\TF} = dW.$$

For a Mamba layer (also titled $\SSM$), we use the state expansion factor $N$. The hidden state of a Mamba layer is a matrix of size $N \times d$. Thus, we calculate the memory needed as
$$M_{\SSM} = dN.$$

Lastly, for the few experiments containing Gated DeltaNet, we only need the token dimension. The state for these layers is a square matrix of the same size as the input. Thus, we calculate the memory needed as
$$M_{\GDN} = d^2.$$

For a model with $N_{\TF}$ Transformer layers, $N_{\SSM}$ Mamba layers, and $N_{\GDN}$ GDN layers, the total memory is
$$M = N_{\TF} dW + N_{\SSM} dN + N_{\GDN} d^2.$$

\paragraph{Enforcing identical memory.} When enforcing that each experiment uses the same allotted memory, we only modify $W$ and $N$. The desired token dimension $d$ is maintained. Additionally, the ratio $W/N$ is roughly maintained so that a desired weight budget favoring one of these layer types over the other is reflected in the used parameters in the experiment. 

When $W$ and $N$ are calibrated, they are always rounded down to stay under the required budget. However, we select memory budgets and token dimensions as powers of 2 so all experiments have precisely equal memory.

\label{app:synth_tasks}

\section{Additional Experiments}
We provide some additional experiments covering other design decisions and further observations relevant to deeper hybrids. 

We begin with results directly related to those in the main body. Those include experiments on the tasks with dominant two layer architectures with more memory and more experiments between dimensions that are powers of 2. We then include some experiments with DeltaNet hybrids, noting that the available design space is much more limited as the state is tied only to token dimension. Some other tasks are then investigated, including the \m{TST} tensor lookup task comparing al 3-layer hybrids as well as a task expected to be optimal for \m{ST}. This final task comes from prior work and confirms that the codebase does not only bias towards the \m{TS} architecture. Lastly, we investigate the results of 8-layer hybrids on two variants of associative recall. We also report their learning-rate sensitivities, noting that for these tasks, each of these architectures shows a surprisingly large range of usable learning rates.

\subsection{Segmented Retrieval}

\begin{figure}
    \centering
    \includegraphics[width=0.8\linewidth]{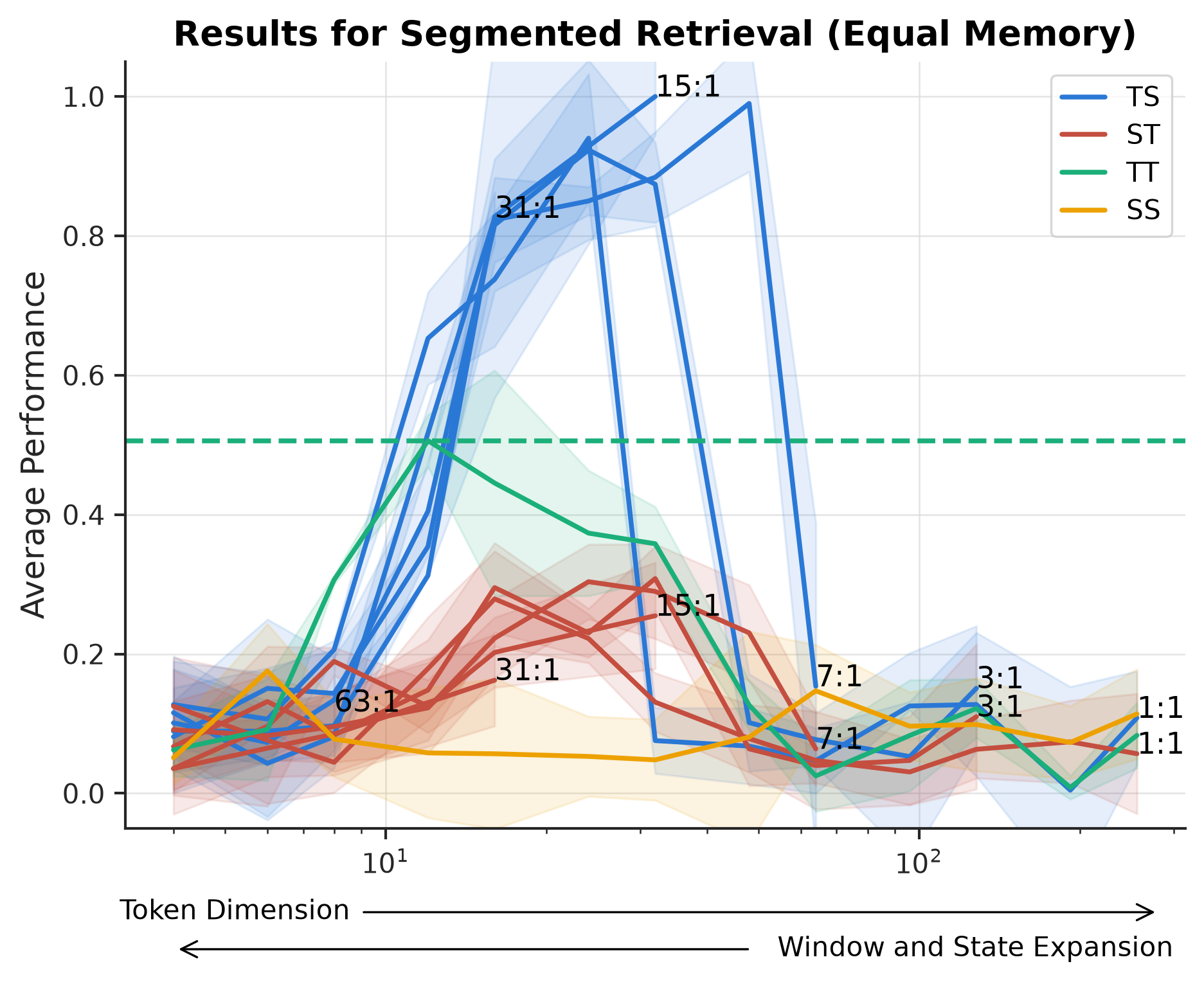}
    \caption{Extended results for segmented retrieval.}
    \label{fig:app_seg_ret_param}
\end{figure}

In the main body, we showed the results for a sweep of token dimensions, windows, and state sizes. These experiments use token dimensions that are all powers of 2. We conducted additional experiments with more token dimensions, with results shown in Figure \ref{fig:app_seg_ret_param}. These are no longer powers of two and thus vary slightly in memory. However, the results still mimic those in the main body, where only the \m{TS} architecture performs the task.

These additional points also allow us to observe an interesting structure between the \m{TT} and \m{TS} architectures. The \m{TT} architecture attains its greatest average performance at a smaller token dimension and larger window than the \m{TS} architectures. These optima show a clear improvement in memory efficiency for \m{TS} architectures: the \m{TS} architectures use their allotted state to increase the computation per token instead of dedicating much of it to a large window to capture only some small components early in the sequence.

\subsection{Different Quantities of Available Memory}

\begin{figure}[t]
    \centering
    \begin{subfigure}[b]{0.48\textwidth}
        \centering
        \includegraphics[width=\linewidth]{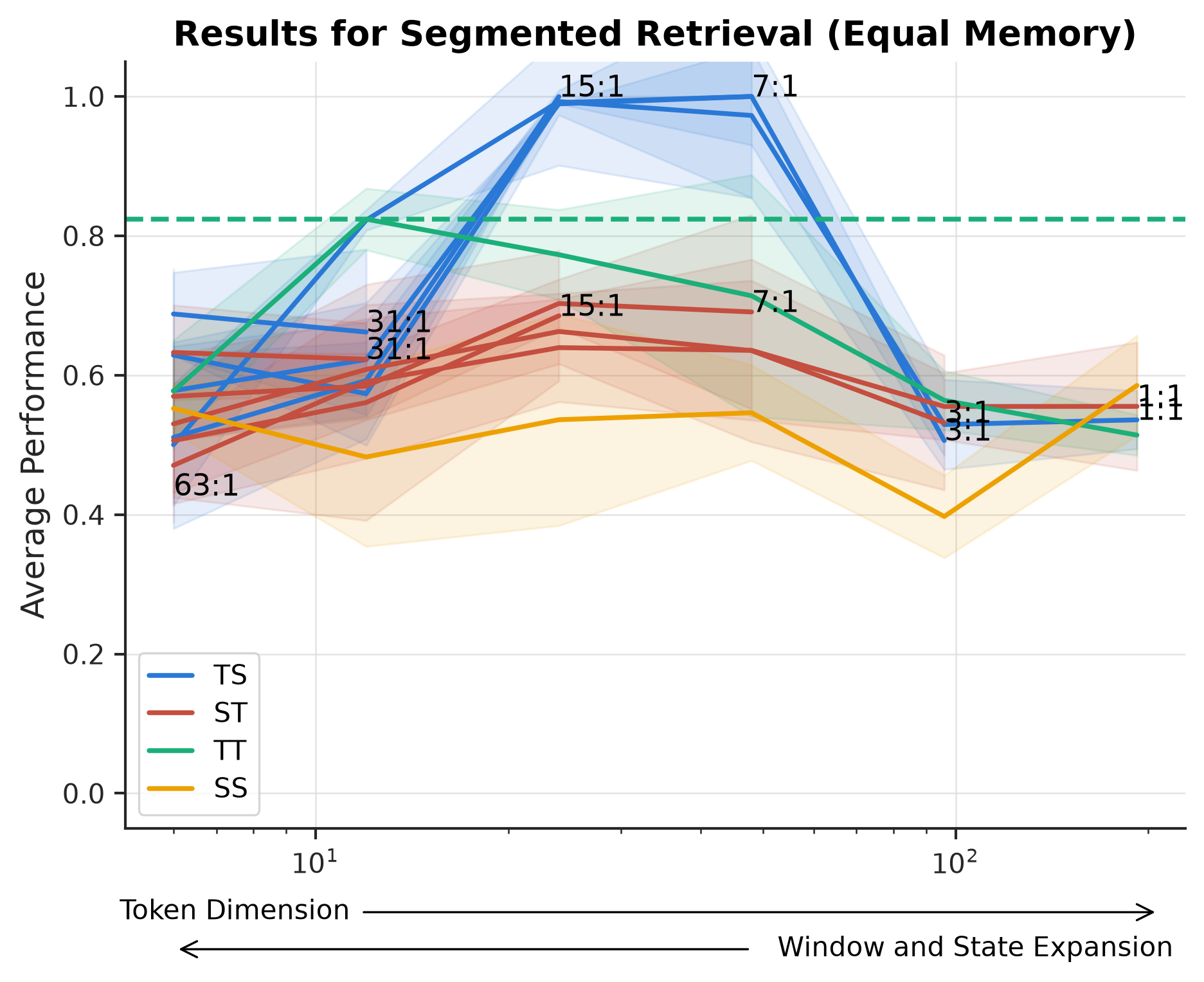}
        \caption{The Segmented Retrieval task.}
    \end{subfigure}
    \hfill
    \begin{subfigure}[b]{0.48\textwidth}
        \centering
        \includegraphics[width=\linewidth]{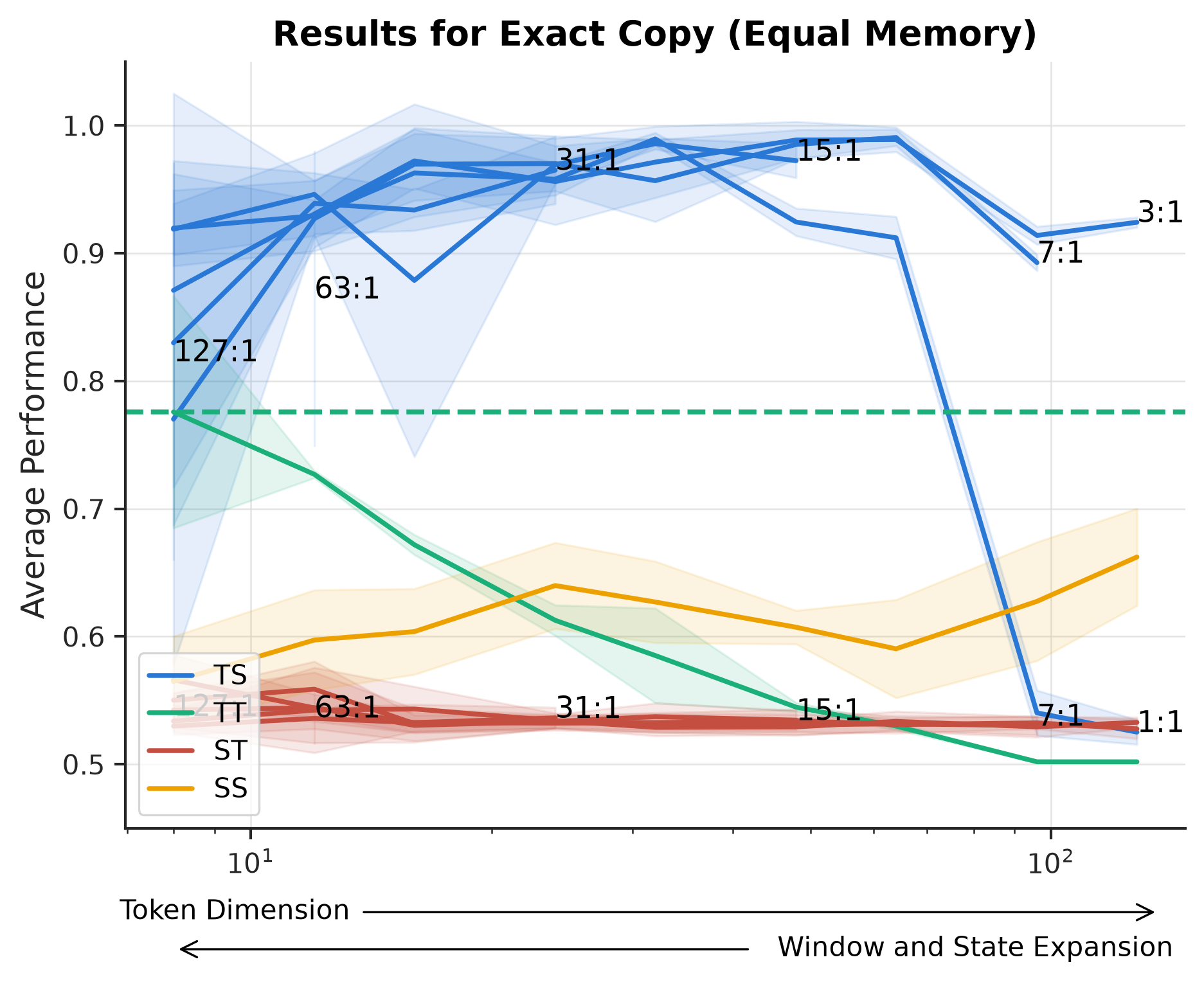}
        \caption{The Exact Copy task.}
    \end{subfigure}
    \caption{Hybrid performance on exact copy for 1024 dimensions of memory.}
    \label{fig:app_mem1024}
\end{figure}

\begin{figure}[t]
    \centering
    \begin{subfigure}[b]{0.48\textwidth}
        \centering
        \includegraphics[width=\linewidth]{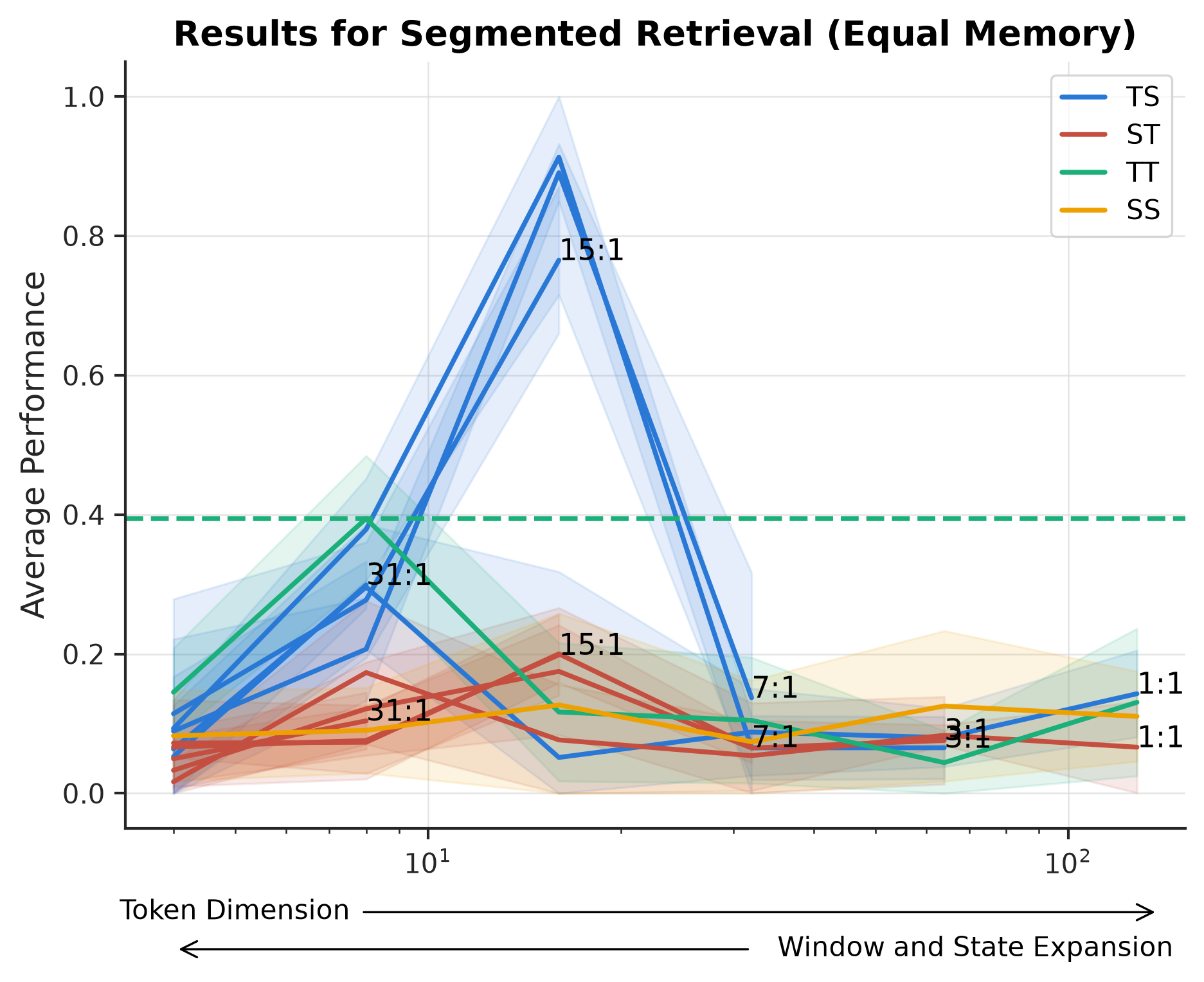}
        \caption{The Segmented Retrieval task.}
    \end{subfigure}
    \hfill
    \begin{subfigure}[b]{0.48\textwidth}
        \centering
        \includegraphics[width=\linewidth]{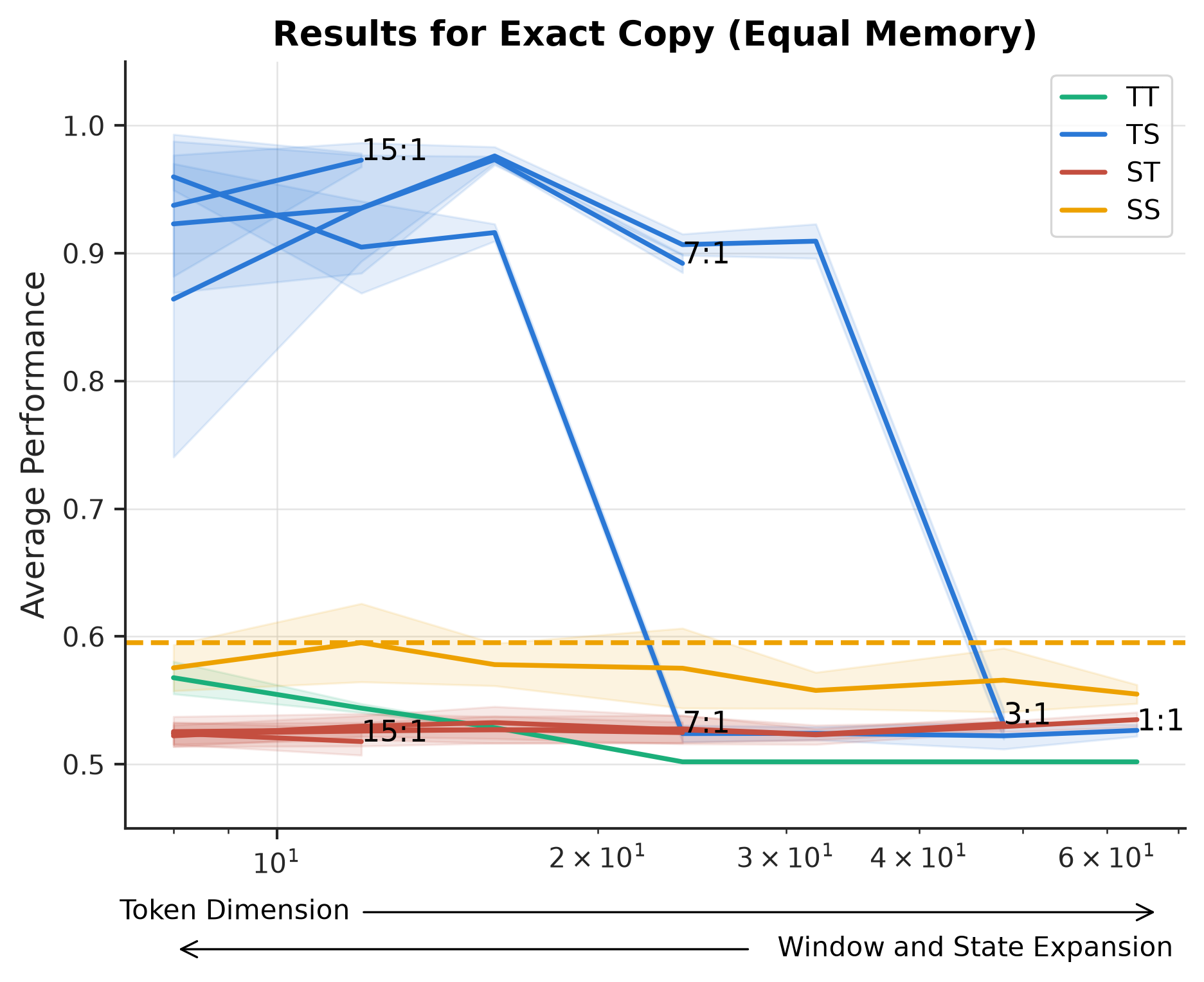}
        \caption{The Exact Copy task.}
    \end{subfigure}
    \caption{Hybrid performance on exact copy for 256 dimensions of memory.}
    \label{fig:app_mem256}
\end{figure}

We first showed a separation between hybrid architectures on segmented retrieval and exact copy using 512 dimensions of memory. We also test architectures with larger and small quantities of available memory.

The results for double the memory can be seen in Figure~\ref{fig:app_mem1024}. We still observe that the \m{TS} architecture is the highest-performing architecture. However, for exact copy, unlike the 512 memory experiments, the \m{TT} architecture gains significantly in performance at small token dimensions. This is likely due to the increased window compared to the main body, where dimension 8 only leaves a window of 128 now, which is large enough to capture most copied subsequences. For the segmented retrieval experiments, we observe that performance on all architectures increases compared to 512 dimensions of memory.

\begin{table}[]
    \centering
    \begin{tabular}{cc|cc}
    \toprule
    Model & Performance & Model & Performance \\
    \midrule
    \m{TTTST} & 0.911  & \m{TTTTT} & 0.533 \\
    \m{TTSTT} & 0.934  & \m{TTTTS} & 0.840 \\
    \m{TTSTS} & 0.884  & \m{TTTSS} & 0.776 \\
    \m{TTSST} & 0.890  & \m{TTSSS} & 0.787 \\
    \m{TSTTT} & 0.922  & \m{STTTT} & 0.527 \\
    \m{TSTTS} & 0.923  & \m{STTTS} & 0.842 \\
    \m{TSTST} & 0.926  & \m{TSSSS} & 0.659 \\
    \m{TSTSS} & 0.888  & \m{STTSS} & 0.770 \\
    \m{TSSTT} & 0.967  & \m{STSSS} & 0.635 \\
    \m{TSSTS} & 0.893  & \m{SSTTT} & 0.550 \\
    \m{TSSST} & 0.949  & \m{SSTTS} & 0.696 \\
    \m{STTST} & 0.900  & \m{SSSTT} & 0.823 \\
    \m{STSTT} & 0.948  & \m{SSTSS} & 0.650 \\
    \m{STSTS} & 0.800  & \m{SSSTS} & 0.709 \\
    \m{STSST} & 0.959  & \m{SSSST} & 0.831 \\
    \m{SSTST} & 1.000  & \m{SSSSS} & 0.478 \\
    \bottomrule
    \end{tabular}
    \caption{Quantitative results for a  deeper hybrids on the \m{TST} task. The left architectures contain \m{TST} as a sub-architecture, while the other three do not.}
    \label{tab:app_tst}
\end{table}

\begin{wrapfigure}[20]{R}{0.5\textwidth}
    \centering
    \includegraphics[width=0.48\textwidth]{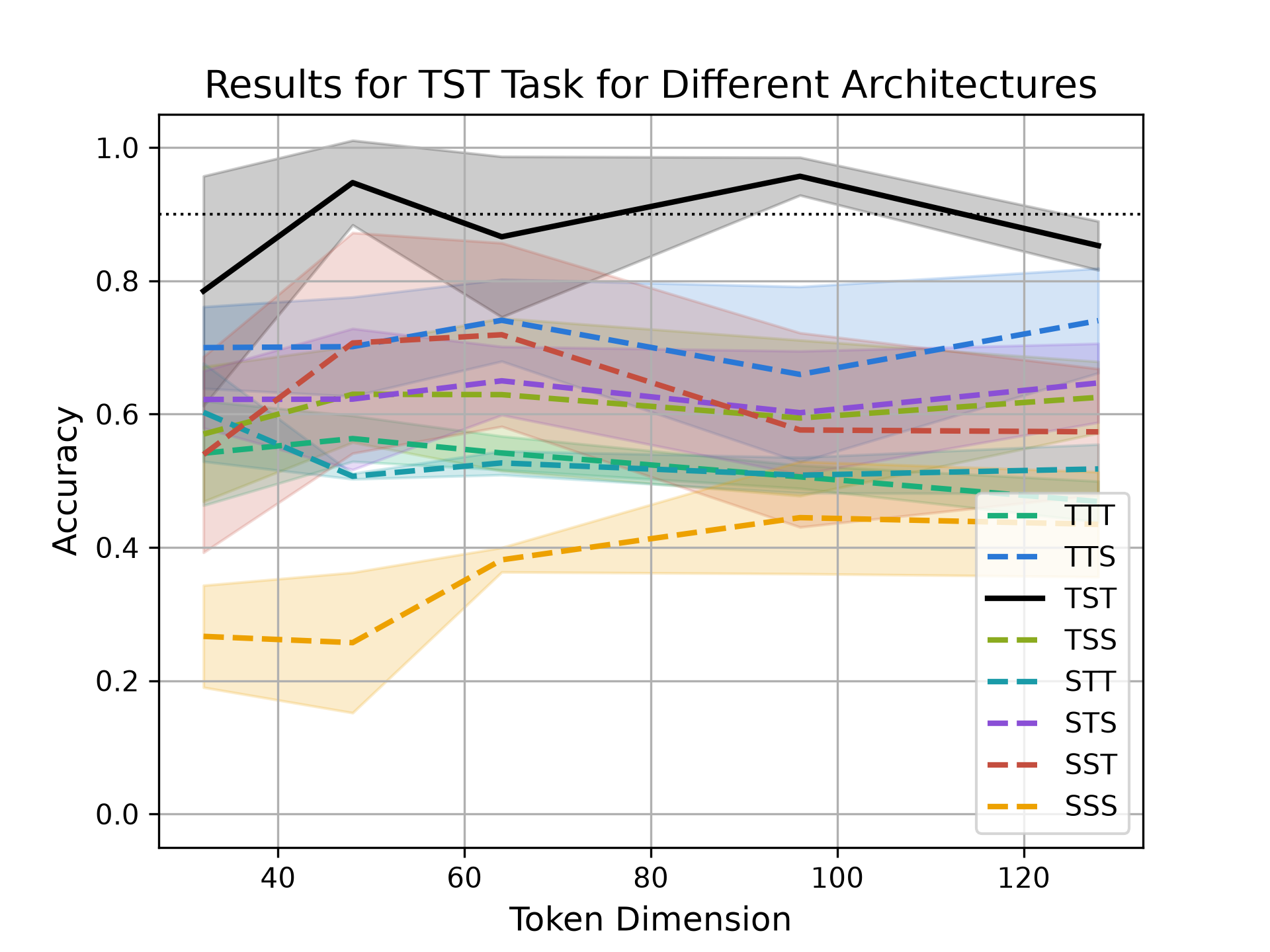}
    \caption{Results for training various architectures with different token dimensions on the \m{TST} task. All models used the same window size and state size. As expected, only the \m{TST} model attains a high accuracy.}
    \label{fig:tst-results}
\end{wrapfigure}

\subsection{Testing DeltaNet Hybrids}

\begin{figure}
    \centering
    \includegraphics[width=0.8\linewidth]{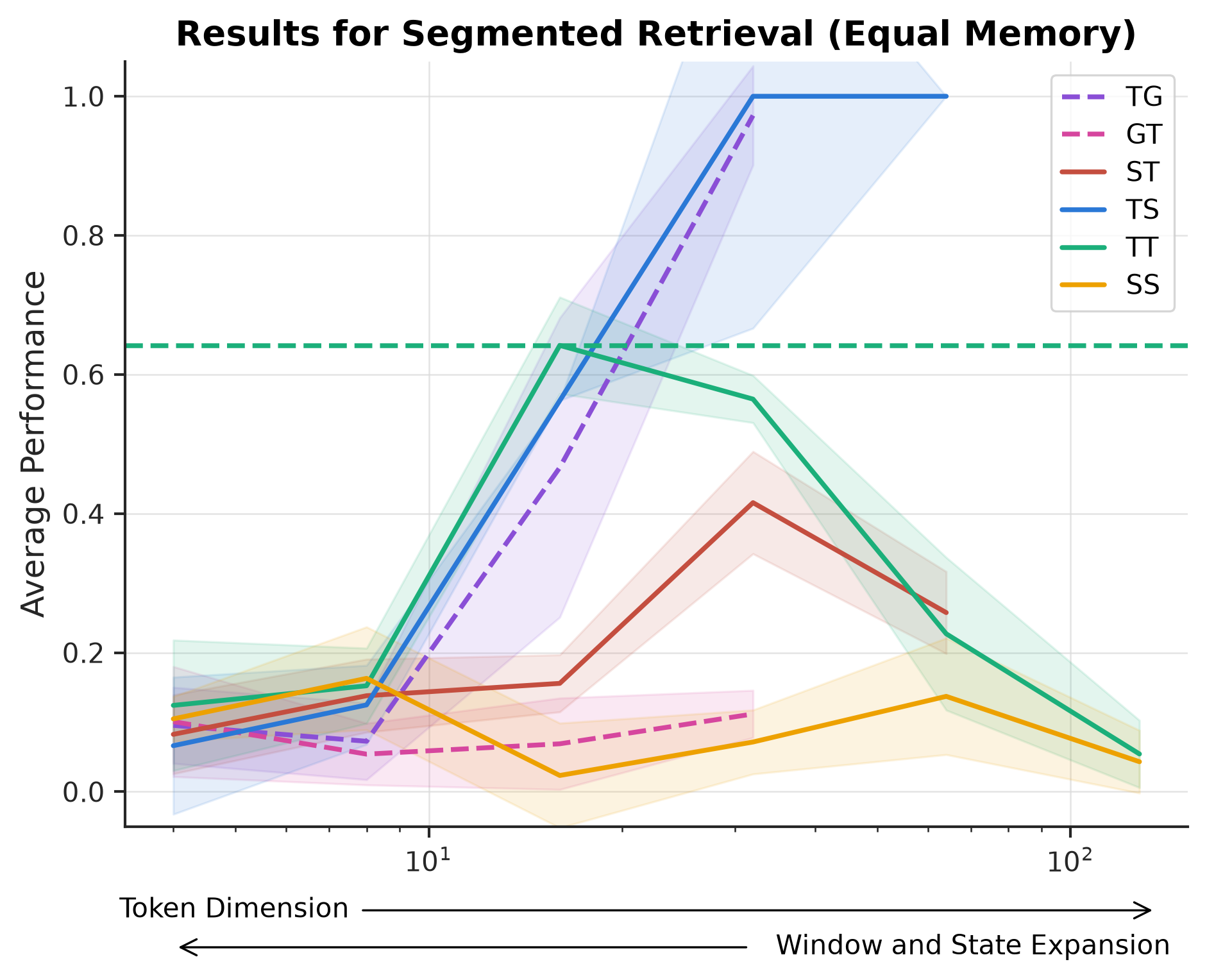}
    \caption{Performance on segmented retrieval, including Gated DeltaNet (\m{G}) hybrids. Only the 15:1 ratio of window to $\SSM$ state for \m{TS} and \m{ST} hybrids is shown.}
    \label{fig:app_seg_ret_param_gdn}
\end{figure}

The majority of this work focuses on the Mamba $\SSM$ architecture. This architecture includes a state-expansion factor that allows the size of the state to be changed independently of other model parameters. This section describes an experiment conducted on hybrids of Transformer and Gated DeltaNet (GDN) layers.

An important note about Gated DeltaNet is that its state matrix, $S_t$, exists in $\R^{d \times d}$. This state size is tied to the token dimension. Thus, unlike the Transformer-Mamba hybrids where the window-to-state-size dimension was free, we can only modify the token dimension and window to achieve the desired memory.

The results are in Figure \ref{fig:app_seg_ret_param_gdn}. For the Transformer-Mamba hybrids, we fix the ratio of window to state-expansion to be 15 to 1, matching the highest performance seen across ratios. We observe the same behavior for Gated DeltaNet as with Mamba, with \m{TS} mirroring \t{TG} closely. However, due to the larger state size of GDN as compared to Mamba, larger token dimensions violate the memory constraints faster; hence, the \m{TG} and \m{GT} curves end earlier. This confirms that available inference memory is a core bottleneck across a couple of different SSM architectures. 

\subsection{Three Layers - Tensor Lookup}
\label{app:tensor_lookup}
As established by the theory about deeper hybrids, there is a natural tensor-lookup task that is predicted to have the \m{TST} architecture be dominant. We investigate this task empirically as well to test the results in a learned setting.

These experiments mirror closely to those for the two-layer architectures. The architectures used are also Transformers with RoPE and Mamba for the SSM. In these experiments, we also hold memory constant. However, we only test on one window-to-state-size ratio for each architecture: 15 to 1. The results are found in Figure~\ref{fig:tst-results}. As expected, the \m{TST} model performs the best, reaching an accuracy above 90\%, unlike all competing architectures. 

We report the exact average performances from Figure~\ref{fig:tst_5_layer} in Table~\ref{tab:app_tst}. This table lists all possible 5-layer architectures. All models with the \m{TST} motif, with the exception of 1, performed better than the remaining models. All models were held to roughly the same memory of 6000. One additional takeaway from this table comes from the architectures of the form $\mm{T}^*\mm{S}^*$ and $\mm{S}^*\mm{T}^*$. For the $\mm{T}^*\mm{S}^*$ models, there is a bias towards more \m{T} layers, while for $\mm{S}^*\mm{T}^*$, there is a bias towards fewer \m{S} layers. Differences such as these support the need for a more principled understanding of various hybrid architecture designs.

\subsection{An \m{ST} Task}

\begin{figure}[t]
    \centering
    \includegraphics[width=0.8\linewidth]{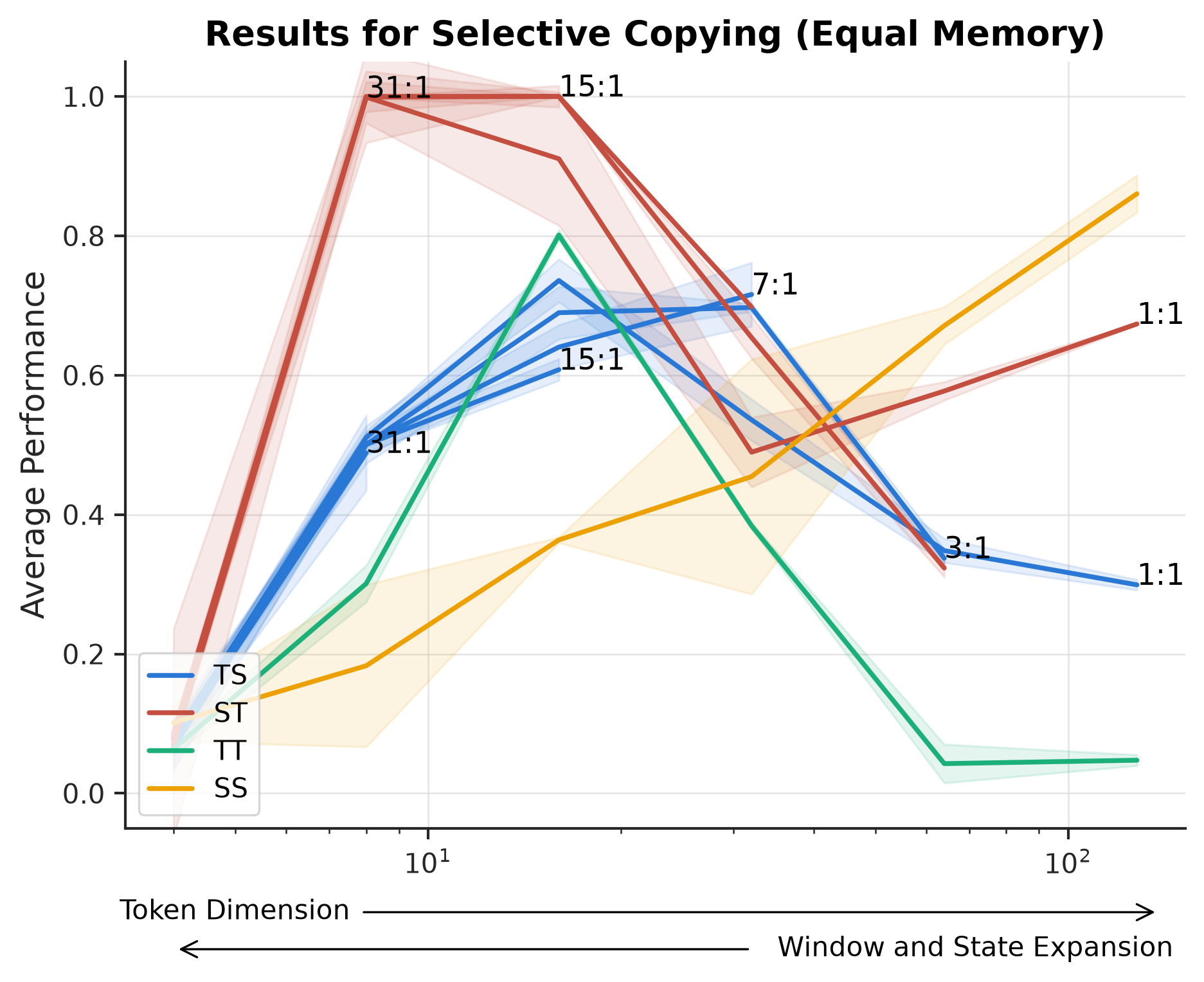}
    \caption{Performance of architectures trained on selective copy. We observe that the \m{ST} architecture is largely dominant, as expected. }
    \label{fig:app_selective_copy_results}
\end{figure}

In \cite{cooper2026expressivityefficiency}, the selective copying task was used to show a separation in hybrid expressivity over pure model expressivity. Empirically, this task was shown to be better suited for \m{ST} rather than \m{TS}. To confirm that the code used does not artificially bias the \m{TS} architecture towards being the dominant architecture, we run the same suite of experiments as used for the 2-layer \m{TS} tasks. 

The results are found in Figure~\ref{fig:app_selective_copy_results}. We observe that the \m{ST} architectures do indeed achieve perfect performance on this task, while the other architectures do not. Thus, the codebase trains architectures in a way that can show various architectures as the highest performing.

Beyond this confirmation, we observe some additional effects in their own right. First, the $\TF$ and $\SSM$ architectures plateau at different locations in the figure. In fact, the $\SSM$ Mamba continues to increase in performance as the dimension increases and the state expansion factor shrinks. 

Because of these different regimes of success for each of the component architectures, the hybrid architectures also have two local maxima in their performance curves. This may indicate that for the hybrids with larger token dimension and smaller window and state-expansion factor, the pure-$\SSM$ algorithm is favored. In the other regime with moderate token dimension, the \m{TS} architecture likely implements an algorithm not too different to the pure $\TF$ one. 

\subsection{Associative Recall Variants}

\begin{figure}[t]
    \centering
    \begin{subfigure}[b]{0.48\textwidth}
        \centering
        \includegraphics[width=\textwidth]{sections/fig/MKAR_sweep.png}
        \caption{Architecture sweep for the $(2,1)$-AR task.}
    \end{subfigure}
    \hfill 
    \begin{subfigure}[b]{0.48\textwidth}
        \centering
        \includegraphics[width=\textwidth]{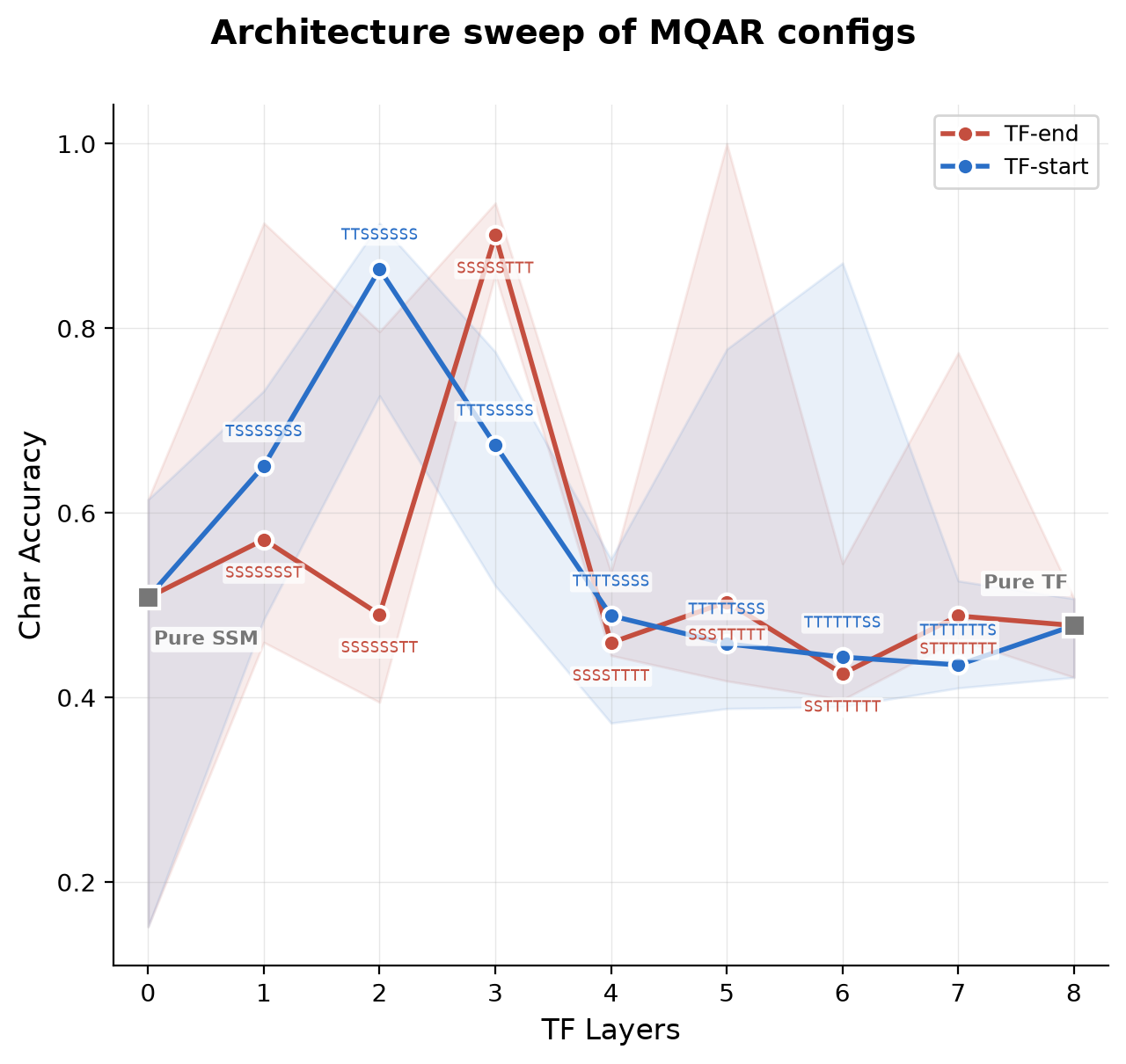}
        \caption{Architecture sweep for the $(1,2)$-AR task.}
    \end{subfigure}
    \caption{Token accuracy of $\mm{T}^*\mm{S}^*$ and $\mm{S}^*\mm{T}^*$ architectures on $(2,1)$ associative recall. Small differences in architectures affect performance dramatically.}
    \label{fig:app_sweep}
\end{figure}

We have focused on experiments pertaining to $(2,1)$-AR as this mirrors a compressed version of multi-document tasks. However, we also want to consider the other variant of associative recall, namely $(1,2)$-AR. This task instead has only $|\gV|$ possible keys instead of $|\gV|^2$, making the key space much easier for an $\SSM$ to store.  

The results for both are shown in Figure~\ref{fig:app_sweep}. The new variant of associative recall, $(1,2)$-AR, shows on average a higher performance, likely due to the decreased size in the key space. However, we also see that the top-performing hybrid layer ordering is different for each task. While both tasks favor low Transformer-to-SSM ratios, $(1,2)$ recall sees stronger performance with more \m{T}-Layers than for $(2,1)$-AR.

\subsection{Learning Rate Sensitivity}
\label{app:lr_sensitivity}

\begin{figure}
    \centering
    \begin{subfigure}[b]{0.48\textwidth}
        \centering
        \includegraphics[width=\textwidth]{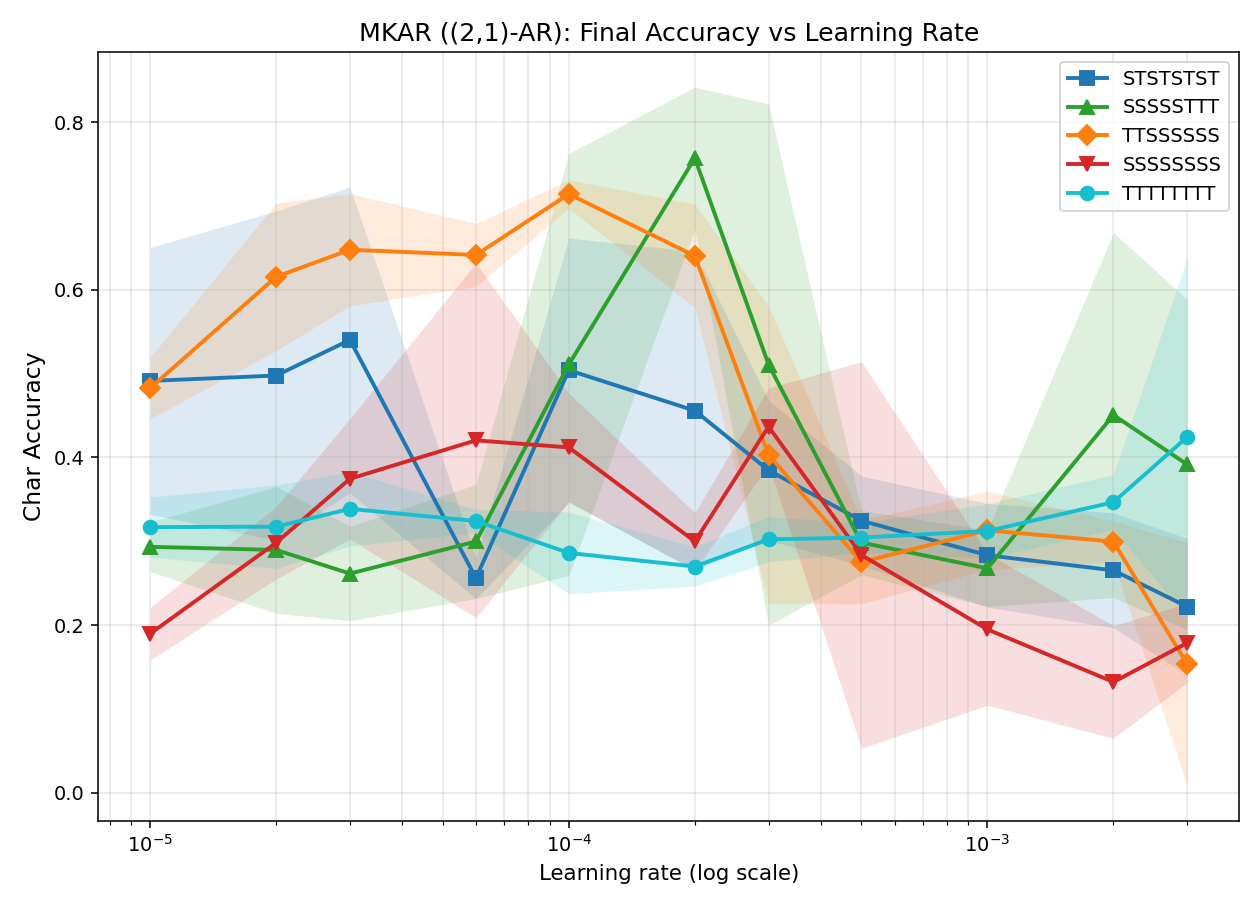}
        \caption{Sweep for $(2,1)$-AR.}
    \end{subfigure}
    \hfill 
    \begin{subfigure}[b]{0.48\textwidth}
        \centering
        \includegraphics[width=\textwidth]{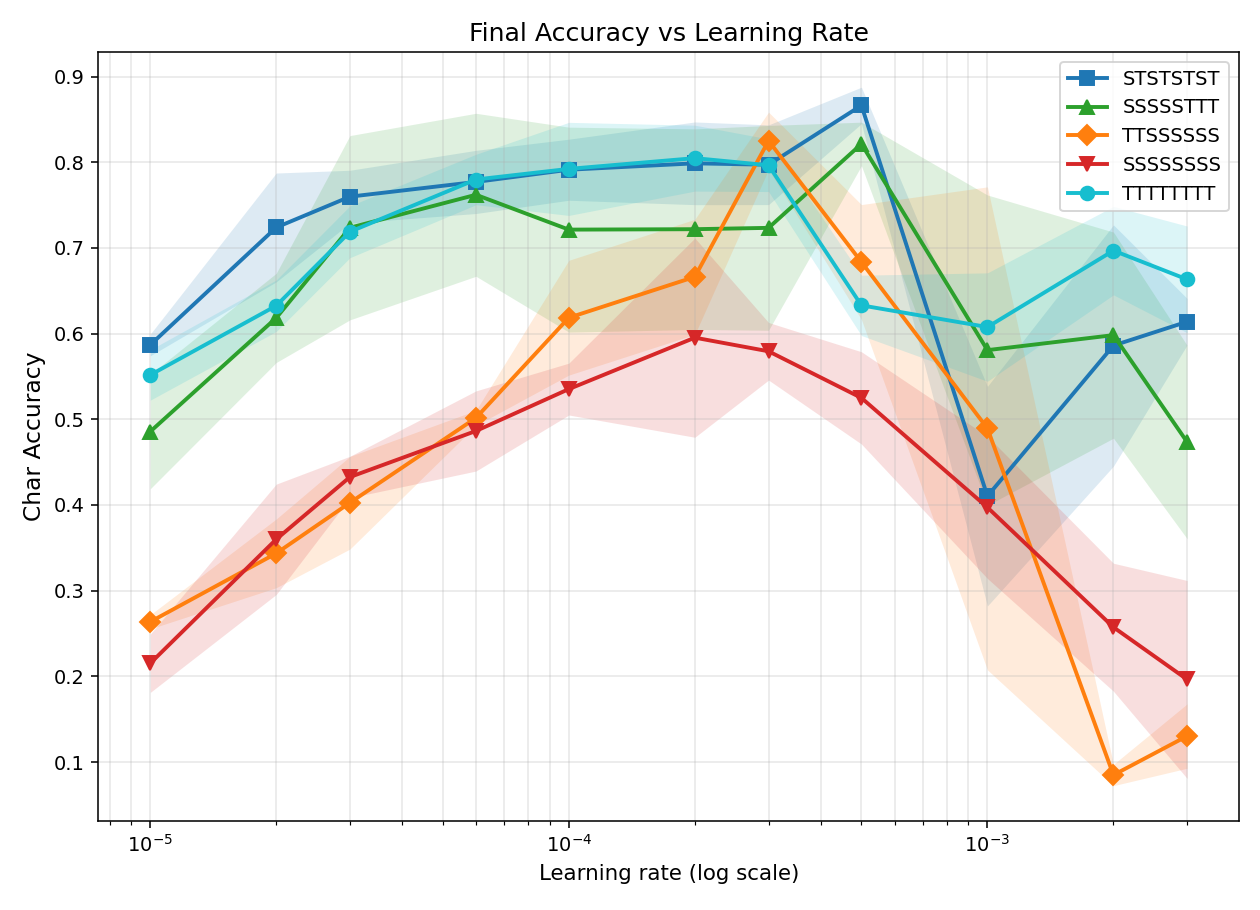}
        \caption{Sweep for $(1,2)$-AR.}
    \end{subfigure}
    \caption{Learning rate sweep for select architectures trained on associative recall.}
    \label{fig:app_lr_sweep}
\end{figure}

Prior work has established that Mamba has a large learning rate sensitivity on the associative recall task \citep{okpekpe2025revisitingassociativerecallmodern}. As a result, we ran learning rate sweeps within our own codebase. The results can be seen in Figure \ref{fig:app_lr_sweep}. Surprisingly, SSMs showed remarkable resilience to learning rate selection. However, other SSM-heavy architectures did show a very narrow band of learning rates that resulted in the highest performance. However, the order of the different architectures' optimal performances remains mostly intact when selecting a learning rate in the optimal band between $10^{-4}$ and $10^{-3}$.

\section{Proving Theorem~\ref{th equal-depth-general-profile}}\label{app deeper model theory}

This section is dedicated to the needed machinery and proofs to prove Theorem~\ref{th equal-depth-general-profile}. We begin with two lemmas from prior work that allow for neighboring layers of the same type to be combined. We then complete proofs for the two parts of Theorem~\ref{th equal-depth-general-profile}: the first focuses on hybrids of the \textit{wrong order} (Theorem~\ref{th wrong order}, while the second focuses on hybrids of the correct order but with \textit{insufficient layers} (Theorem~\ref{th insufficient layers}).

\paragraph{Model Merge.} We borrow two lemmas from prior work. These allow neighboring layers of common type to be combined.

\begin{lemma}[Lemma 3.5 in \cite{cooper2026expressivityefficiency}]\label{lm group ssm}
Consider a $k$-layer model $M$ defined that is a composition of $k$ state space layers $\SSM_j, j \in [k]$. There is a model $\SSM'$ that only consists of a single layer SSM, which behaves the same as $M$. In particular, denote the state space of $\SSM_j$ as $\mathcal{S}_j$, $j \in [k]$, and denote by the state space of $\SSM'$ as $\mathcal {S}'$, then $|\mathcal {S}'| \le \prod_{j=1}^k |\mathcal S_j|$.
\end{lemma}

\begin{lemma}\label{lm group tf}
Consider a $k$-layer model $M$ defined that is a composition of $k$ causal windowed functions $f_j, j \in [k]$. There is a model $M'$ that only consists of a single layer causal windowed function $f$, which behaves the same as $M$. In particular, denote the window size of $f_j$ as $W_j$, $j \in [k]$, and denote by the window size of $f$ as $W$, then $W' \le \sum_{j=1}^kW_i-k+1$.
\end{lemma}

\subsection{Wrong use of ordering}\label{app wrong ordering}

We start with the case where a competitive model $A$ does not contain $\mathrm{comp}(\tau)$ as a subsequence.

\begin{theorem}\label{th wrong order}
    Let $D \in \mathbf{N}_+$ and $\tau =
    X_1^{r_1}X_2^{r_2}\cdots X_h^{r_h} \in \Sigma^*$ be an arbitrary architecture profile with depth $D=\sum_{j=1}^h r_j$. Let $\alpha:=\mathrm{comp}(\tau)$ and $\epsilon \in (0,1)$ be an arbitrary fixed constant.
    There exist constants $ c_\tau>0,
    C_{\tau,\epsilon}>0,$ a family of tasks $\{F_{\tau,\epsilon,m}\}_{m\ge 1}$ and a family of distributions $\mu_m$ over a sequence of tokens of length $L_m$ parameterized by $m \in \mathbf{N}_+$, such that for all $m$ sufficiently large, the following hold.  Let $A \in \Sigma^{\le D}$ be an architecture profile with $\alpha \not\preceq A$. Then for every hybrid model $M\in\mathcal M_A(B,P,W)$ that satisfies $\mathrm{err}(M,F_{\tau,\epsilon,m}) \le 1/3$, must satisfy $W \ge c_\tau m^{1/\epsilon}$ or $\max\{B,P\} \ge c_\tau m$.
\end{theorem}

To begin with, we present the following model simplification lemma.
\begin{lemma}\label{lm model simplification}
     Let $\tau =
    X_1^{r_1}X_2^{r_2}\cdots X_h^{r_h} \in \Sigma^*$ be an arbitrary architecture with depth $D$ and $\alpha=\mathrm{comp}(\tau)$. Denote by $\bar \alpha = \bar X_1,\dots,\bar X_h$ be the complementary of $\alpha$, where $\bar X_i \neq X_i \in \Sigma$.
    Let $A \in \Sigma^{\le D}$ such that $\alpha \not\preceq A$, then $\mathrm{comp}(A) \preceq \bar\alpha$. Furthermore, for $B,P,W \in \mathbf{N}_+$, and $M \in \mathcal{M}_A(B,P,W)$, there exist a hybrid model $M'$ in $\mathcal{M}_{\bar \alpha}(DB,P,D(W-1))$ that can represent $M$.
\end{lemma}

\begin{proof}[Proof of Lemma~\ref{lm model simplification}]
    Since $\alpha = X_1,\dots,X_k$ and $\mathrm{comp}(A)$ is an alternating architecture profile, we consider two cases. In the first case, $\mathrm{comp}(A)$ begins with $X_1$.
    We claim that it must be the case where $|\mathrm{comp}(A)|\le k-1$, otherwise, $\alpha\preceq \mathrm{comp}(A)$. This implies that $\mathrm{comp}(A)$ must be a substring of $\bar \alpha$, starting from the second layer. $\mathrm{comp}(A) \preceq \bar \alpha$.
    In the second case, $\mathrm{comp}(A)$ start with $\bar X_1$. In this case,  $|\mathrm{comp}(A)|\le k$, otherwise $\alpha \preceq  \mathrm{comp}(A)$. So, $\mathrm{comp}(A) \preceq \bar \alpha$.

    Since $\mathrm{comp}(A) \preceq \bar \alpha$, the second part of Lemma~\ref{lm model simplification} follows directly from Lemma~\ref{lm group ssm} and Lemma~\ref{lm group tf}.
    
\end{proof}

Lemma~\ref{lm model simplification} implies that instead of competing with an arbitrary architecture profile, it is sufficient to compete with hybrid models in $\mathcal{M}_{\bar \alpha}(DB,P,D(W-1))$. A central part of our analysis is to characterize the format and dependence of information through different layers and at different positions.

Before proceeding, we first introduce some notations that we will use in the rest of the analysis to keep track of different types of layers. Let $\bar \alpha = \bar X_1 \bar X_2 \dots \bar X_h$ and $M' \in \mathcal{M}_{\bar \alpha}(DB,P,D(W-1))$.
Denote by $k=\{i \in [h] \mid \bar X_i =\mm{S}\}$. We introduce indicator $\delta_L=\mathbf{1}\{\bar X_1 =\mm{T}\}$ and $\delta_R=\mathbf{1}\{\bar X_h =\mm{T}\}$.
That is to say, we can write $\bar \alpha = \mm{T}_0^{\delta_L}\mm{S}_1\mm{T}_1,\dots,\mm{S}_{k-1}\mm{T}_{k-1}\mm{S}_k\mm{T}_k^{\delta_R}$, where $\mm{T}_{0}^{0},\mm{T}_{s}^{0}$ mean the corresponding boundary layers are absent. In particular, we have $h = 2k-1+\delta_L+\delta_R$.

We introduce the family of tasks we are considering as well as the format of input.
\begin{definition}\label{def wrong order restate}
    Let $\mathcal{X}_{key}$ be a finite space of keys, and $\mathcal{C}$ be a finite space of vocabularies. Let $\eps \in (0,1)$ and $m \in \mathbf{N}_+$. Let $\tau$ be an architecture profile with $\bar \alpha = \mm{T}_0^{\delta_L}\mm{S}_1\mm{T}_1,\dots,\mm{S}_{k-1}\mm{T}_{k-1}\mm{S}_k\mm{T}_k^{\delta_R}$.
    A level-$k$ recursive chain $F_{\tau,\eps,m}$ is defined by a sequence of $k$ functions. Specifically, for $i \in [k]$, $H_i(\z_{i-1},\x_i): \mathcal{X}_{key} \times \mathcal{C}^m \to \mathcal{X}_{key}$, where $Z_i := H_i(\z_{i-1},\x_i)$.  
    Given a sequence of input $I:=(\z_0,\Vec{0}_0,\x_1,\Vec{0}_1,\dots,\x_k,\Vec{0}_k)$ of length $L_m$, $F_{\tau,\eps,m}(I):= \z_k$, where each $\Vec{0}_i$ is a sequence of $0$ vectors. 
\end{definition}

Let $\Delta_m:= D \lfloor m^{1/\epsilon}\rfloor$. We define position variables in the input sequence $I$.
We consider two cases. If $\delta_L=0$, we set $u_1 = 2$, otherwise, we set
$u_1 = \Delta_m+1$, $v_1=u_1+m-1$ and for $i \ge 1$, $b_i=v_i+i\Delta_m$, $u_{i+1} = b_i+\Delta_m+1$, $v_{i+1} = u_{i+1} + m-1$. Then, $z_0 = I_1$, $x_i = I_{[u_i,v_i]}$ and the other positions of $S$ are $0$. For the output position, if $\delta_R = 1$, we ask the output to be at $b_k$; otherwise, $\Vec{o}_k$ does not exist and we ask the output to be as $v_k$.

The input distribution $\mu_m$ for $I$ that we are considering samples an input sequence $I$ by sample $\z_0$ and each $\x_i$ uniformly and independently from the domains.

We remark that Definition~\ref{def wrong order restate} immediately implies the existence of a model in $\mathcal{M}_\tau$ that can efficiently solve the task exactly.
To see this, we consider a model $M$ with profile $\alpha=\mathrm{comp}(\tau)$, for every $i \ge 1$, since the window size of a $\mm{T}$ layer is $m+1$, we use $O_\tau(\log m)$ bits to feed $((\z_{i-1},\x_{i,1}),\dots,(\z_{i-1},\x_{i,m}))$ to $\mm{T}$, which is enough for it to compute $\z_i$, precisely. This implies that if $\mathcal{X}_{key}$ has size $\mathrm{poly}_\tau(m)$, then the next $\mm{S}$-layer can use $O_\tau(\log m)$ to feed $((\z_{i},\x_{i+1,1}),\dots,(\z_{i},\x_{i+1,m}))$ to the next $\mm{T}$ and thus it can solve the task exactly. So, in the rest of the analysis we focus on the hardness part.

Given an input sequence $I$, we denote by $Y^{(j)}(I)$ the intermediate input of $\mm{T}_j$ and denote by $h_j(t;I)$ the internal state of $\mm{S}_j$ after processing position $t$.
By Lemma~\ref{lm model simplification}, we without loss of generality assume that $h_j(t;I) \in \{0,1\}^{DB}$.
Based on the notation of Definition~\ref{def wrong order}, for $0 \le j \le i$, we define position $a_{ij} = v_i +j\Delta_m$. In particular, $a_{i0} = v_i$ and $a_{ii}=b_i$. We define the state record at stage $i$ with input $I$ as
\begin{align*}
    C_i(I):=(h_1(a_{i1};I),h_2(a_{i2};I),\dots,h_i(a_{ii};I)).
\end{align*}

Our first lemma shows that if $C_i$ is fixed, then for every $\ell>b_i$, $Y^{(i)}_\ell$ is only a deterministic function of the parts of the input after $b_i$.

\begin{lemma}\label{lm state dependence}
     Let $D \in \mathbf{N}_+$ and $\tau =
    X_1^{r_1}X_2^{r_2}\cdots X_h^{r_h} \in \Sigma^*$ be an arbitrary architecture with $\alpha=\mathrm{comp}(\tau)$ and depth $D$. Denote by $\bar \alpha = \bar X_1,\dots,\bar X_h$ be the complementary of $\alpha$, where $\bar X_i \neq X_i \in \Sigma$. Let $M' \in \mathcal{M}_{\bar \alpha}(DB,P,D(W-1))$ be a hybrid model.
    Let $I,I'$ be two sequences of inputs of the level-$k$ recursive chain $F_{\tau,\eps,m}$ and let $i \in [k]$.
    Suppose that $I_{\ell} = I'_{\ell}$, for every $\ell > v_i$ and $C_i(I) = C_i(I')$, then for every $j \in [i]$ and for every $\ell>a_{ij}$, $Y^{(j)}(I)_{\ell} = Y^{(j)}(I')_{\ell}$, and for every $\ell \ge a_{ij}$ $h_j(\ell;I) = h_j(\ell;I')$.
\end{lemma}

\begin{proof}[Proof of Lemma~\ref{lm state dependence}]
    We prove the statement by induction. For $j = 1$, we consider two cases. In the first case, $\delta_L=0$, which implies that $\bar \alpha$ begins with $\mm{S}_1$. In this case, the states of $h_1$ only depend on the past states and input and thus we have for every $\ell \ge a_{i1}$ $h_1(\ell;I) = h_1(\ell;I')$. In the second case, $\bar \alpha$ begins with $\mm{T}_0\mm{S}_1$. By the choice of $\Delta_m$, for every $\ell>v_i+\Delta_m = a_{i1}$, $\mm{T}_0(I)_\ell = \mm{T}_0(I')_\ell$, which implies that $h_1(\ell;I) = h_1(\ell;I')$. The results for the states automatically translate to the results for $Y^{(1)}$.

    Now suppose the statement holds for $j$. We prove it for the case of $j+1$.
    Since the window size of a transformer layer is at most $\Delta_m$, we know that the input of $\mm{T}_j$ only depends on positions after $a_{ij}$ when evaluating after $a_{i(j+1)}$. Since for every $\ell>a_{ij}$, $Y^{(j)}(I)_{\ell} = Y^{(j)}(I')_{\ell}$, we know that $\mm{T}_j(Y^{(j)}(I))_\ell = \mm{T}_j(Y^{(j)}(I'))_\ell$ for $\ell>a_{i(j+1)}$.
    On the other hand, since for every $\ell>a_{ij}$,  $h_j(\ell;I) = h_j(\ell;I')$, we know that for every $\ell>a_{i(j+1)}$, $h_{j+1}(\ell;I) = h_{j+1}(\ell;I')$ and thus $Y^{(j+1)}(I)_\ell = Y^{(j+1)}(I')_\ell$.
    
\end{proof}

Next, we characterize the evolution of $C_i$ when $i$ increases. i.e how the record of the hidden states of each SSM layer changes according to the position.

\begin{lemma}\label{lm state representation}
     Let $D \in \mathbf{N}_+$ and $\tau =
    X_1^{r_1}X_2^{r_2}\cdots X_h^{r_h} \in \Sigma^*$ be an arbitrary architecture with $\alpha=\mathrm{comp}(\tau)$ and depth $D$. Denote by $\bar \alpha = \bar X_1,\dots,\bar X_h$ be the complementary of $\alpha$, where $\bar X_i \neq X_i \in \Sigma$. Let $M' \in \mathcal{M}_{\bar \alpha}(DB,P,D(W-1))$ be a hybrid model.
    Let $I$ be an input of a level-$k$ recursive chain, realized by $(\z_0,\x_1,\dots,\x_k).$ Denote by $J_i$ the prefix $(\z_0,\x_1,\dots,\x_i)$.
    For every $i \in [k-1]$ and for every $c$ such that $C_i(I) = c$, there exists a map $H_{i,c}: \mathcal{C}^m \to \{0,1\}^{iDB}, E_{i,c}: \mathcal{C}^m \to \mathcal{C}^{b_{i+1}-u_{i+1}+1}, \Phi_{h_{i+1}(u_{i+1}-1;I)}:\mathcal{C}^{b_{i+1}-u_{i+1}+1} \to \{0,1\}^{DB}$ such that 
    $(h_1(a_{(i+1)1};I),h_2(a_{(i+1)2};I),\dots,h_i(a_{(i+1)i};I)) = H_{i,c}(\x_{i+1})$ and $h_{i+1}(a_{(i+1)(i+1)};I) = \Phi_{h_{i+1}(u_{i+1}-1;J_i)}(E_{i,c}(\x_{i+1}))$.
\end{lemma}
In other words, once we have fixed the prefix of the input $J_i$ as well as the value of $C_i(I)$, there exists maps $H_{i,c},E_{i,c}$ that only depends on the compressed information $C_i(I)$ and a map $\Phi_{h_{i+1}(u_{i+1}-1;I)}$ that only depends on the state $h_{i+1}(u_{i+1}-1;I)$ such that $C_{i+1}(I) = (H_{i,c}(\x_{i+1}),\Phi_{h_{i+1}(u_{i+1}-1;I)}(E_{i,c})(\x_{i+1}))$. To simplify the notation, for any fixed $i,c$ and $h_{i+1}(u_{i+1}-1;I) = \gamma$, we denote by $C_{i+1}(I) = \Gamma_{i,c,\gamma}(\x_{i+1})$.

\begin{proof}[Proof of Lemma~\ref{lm state representation}]
    We start by proving the existence of $H_{i,c}$. By Lemma~\ref{lm state dependence}, we know that once we fix $C_i(I) = c$, for every $j \le i$ and for every $\ell \ge a_{ij}$, the hidden state $h_j(\ell,I)$ is only a function of $\x_{i+1}$. This implies that the vector $(h_1(a_{(i+1)1};I),h_2(a_{(i+1)2};I),\dots,h_i(a_{(i+1)i};I))$ only depends on $\x_{i+1}$, given the current state $C_i(J_i)$.

    We next prove the existence of $\Phi_{J_i}$ and $E_{i,c}$. We notice that the state $h_{i+1}(a_{(i+1)(i+1)};I)$ is a function of the state $h_{i+1}(b_i,I)$ and the input context $\mm{T}_i(Y^{(i)})$. 
    In particular, if we denote by $\delta$ the transition function of the state of $\mm{S}_{i+1}$. Then, it holds that for $j \ge 0$, we have $h_{i+1}(u_{i+1}+j;I) = \delta(h_{i+1}(u_{i+1}+j-1;I),\mm{T}_i(Y^{(i)}(I))_{u_{i+1}+j})$.

    By Lemma~\ref{lm state dependence}, we know that for  $\ell>a_{ii}$, $Y^{(i)}(I)_{\ell}$ is a function that only depends on $\x_{i+1}$. This implies that given $C_i = c$, $\mm{T}_i(Y^{(i)}(I))_{u_{i+1}+j})$, for $j = 0,\dots,b_i-v_i$ only depends on $\x_{i+1}$, while the state $h_{i+1}(u_{i+1}-1;I)$ can depend on the entire prefix $J_i$. This implies that there exist maps $E_{i,c}: \mathcal{C}^m \to \mathcal{C}^{b_{i+1}-u_{i+1}+1}, \Phi_{h_{i+1}(u_{i+1}-1;I)}:\mathcal{C}^{b_{i+1}-u_{i+1}+1} \to \{0,1\}^{DB}$ such that $h_{i+1}(a_{(i+1)(i+1)};I) = \Phi_{h_{i+1}(u_{i+1}-1;I)}(E_{i,c}(\x_{i+1}))$
\end{proof}

Based on Lemma~\ref{lm state dependence} and Lemma~\ref{lm state representation}, we now give the construction for the functions $H_i$ that we are using.
Our construction follows the following intuition. By Lemma~\ref{lm state dependence} and Lemma~\ref{lm state representation}, conditioned over $C_i$ and $J_i$, the behavior of the model is a fixed function of $\x_{i+1}$. In particular, the total number of such functions can be bounded.
Since each $\x_i$ is drawn uniformly at random, assuming for the time that the model succeed with a good probability at the next stage only if the true key
$\z_{i+1}$ falls into a small list of candidates of length $l$, we will next show that by carefully choosing the size of $\mathcal{X}_{key}$, there must be some Boolean function $H$ such that no matter which function in the class is chosen, there are also at most $l$ possible $\z_i$ that makes $\z_{i+1}$ fall into the list with a good probability.
This implies that if we sequentially choose all the $H_i$ based on this, then the model finally will only have a small probability to succeed.

To start with, we first show that for a single step such a function $H_i$ exists.
Using a probablistic construction, we show the following lemma.
\begin{lemma}\label{lm H construction}
    Let $\mathcal{K},\mathcal{X}.\mathcal{Q},\mathcal{A}$ be finite sets. Denote by $N=|\mathcal{K}|$ and $n = |\mathcal{X}|$. Let $\mathcal{G}: \mathcal{A} \times \mathcal{X} \to \mathcal{Q}$ be a finite set of maps.
    For $l \in [N]$, let $\mathcal{L}_l = \{\mathcal{L}: \mathcal{Q} \to 2^{\mathcal{K}}\mid |\mathcal{L}(q)| \le l, \forall q \in \mathcal{Q}  \}$.
    For $\theta \in (0,1)$, if 
    \begin{align*}
        \binom{N}{l+1}|\mathcal{G}| |\mathcal{L}_l|[|\mathcal{A}|2^n(l/N)^{\theta n}]^{l+1}<1,
    \end{align*}
    then there exists a map $H: \mathcal{K} \times \mathcal{X} \to \mathcal{K}$ such that
    for every $G \in \mathcal{G}$ and every $\mathcal{L} \in \mathcal{L}_l$ the set
    \begin{align*}
        B(G,\mathcal{L}):=\{\z \in \mathcal{K} \mid \exists a \in \mathcal{A}, \Pr_{\x\sim U(\mathcal{X})}[H(\z,\x) \in \mathcal{L}(G(a,\x))] \ge \theta \}
    \end{align*}
    has cardinality at most $l$
\end{lemma}

\begin{proof}[Proof of Lemma~\ref{lm H construction}]
    Fix $\z \in \mathcal{K}, G \in \mathcal{G}, \mathcal{L} \in \mathcal{L}_l, a \in \mathcal{A}$. We choose $H$ from all possible maps uniformly at random.
    Since for each $\x \in \mathcal{X}$, the list $\mathcal{L}(G(a,\x))$ has size at most $l$, we have 
    \begin{align*}
        \Pr_{H}[H(\z,\x) \in \mathcal{L}(G(a,\x))] \le l/N.
    \end{align*}
    By taking a union bound over $\mathcal{X}$, we have 
    \begin{align*}
        \Pr_{H}[\{H :\Pr_{\x \sim U(\mathcal{X})}[H(\z,\x) \in \mathcal{L}(G(a,\x))]\ge \theta\}] \le \sum_{r \ge \theta n}\binom{n}{r}(l/N)^r \le 2^n(l/N)^{\theta n}.
    \end{align*}
    Now we take a union bound over $\mathcal{A}$, for every fixed $\z \in \mathcal{K}$ we have 
    \begin{align*}
        \Pr_{H}[\z \in B(G,\mathcal{L})] \le |\mathcal{A}|2^n(l/N)^{\theta n}.
    \end{align*}
Now we consider the probability that a bad event happens over the randomization of the choice of $H$, which means there exists $l+1$ possible values in $B(G,\mathcal{L})$.
Fix a list of $l+1$ value $\z^{(1)},\dots,\z^{(l+1)}$, since the events $\z^{(i)} \in B(G,\mathcal{L})$ only depend on the function $H(\z^{(i)},\cdot)$ and are independent, we know that 
\begin{align*}
    \Pr_{H}[\z^{(1)},\dots,\z^{(l+1)} \in B(G,\mathcal{L})] \le (|\mathcal{A}|2^n(l/N)^{\theta n})^{l+1}.
\end{align*}
We now take the union bound of $G,\mathcal{L}$ and all possible tuples  $\z^{(1)},\dots,\z^{(l+1)}$, the assumption of Lemma~\ref{lm H construction} implies that the probability that we choose an $H$ that creates some $B(G,\mathcal{L})$ with size more than $l$ is strictly less than $1$.
This implies the existence of $H$ that satisfies the statement of Lemma~\ref{lm H construction}.
    
\end{proof}

We next fit Lemma~\ref{lm H construction} into our framework using Lemma~\ref{lm state dependence} and Lemma~\ref{lm state representation}.

\begin{lemma}\label{lm chain definition}
    For $\eps \in (0,1)$, let $r_\epsilon = \lfloor(1/\eps)\rfloor+3$, $N_m = m^{2r_\epsilon}$, $l_m = m^{r_\eps}$ and
    let $\mathcal{X}_{key} = [N_m]$. For all sufficiently large $m$, 
    there exist functions $H_1,\dots,H_k: \mathcal{X}_{key} \times \mathcal{C}^m \to \mathcal{X}_{key}$ such that for every $M \in \mathcal{M}_{\bar \alpha}(DB,P,D(W-1))$, an arbitrary hybrid model that satisfies $B,P < c_\tau m$ and $W < c_\tau m^{1/\eps}$, $c_\tau \ll 1$ a constant, the follows hold.
    Fix a level $0\le i<k$ and $c= C_i$, for every $\mathcal{L}:\mathrm{Im}(C_{i+1}) \to 2^{\mathcal{X}_{key}}$such that $|\mathcal{L}(c')| \le l_m$ for all $c'\in \mathrm{Im}(C_{i+1})$, the set 
    $\{z \in \mathcal{X}_{key} \mid \exists \gamma \in \{0,1\}^{DB}, \Pr_{\x_{i+1}\sim \mathcal{C}^m}[H_{i+1}(\z,\x_{i+1}) \in \mathcal{L}(\Gamma_{i,c,\gamma}(\x_{i+1}))] \ge 1/(4k) \}$ has size at most $l_m$.
\end{lemma}

\begin{proof}[Proof of Lemma~\ref{lm chain definition}]
    We apply Lemma~\ref{lm H construction} to prove Lemma~\ref{lm chain definition}.
    Let $\mathcal{K}: = \mathcal{X}_{key}, N = N_m$ and $\mathcal{X} = \mathcal{C}^m, n = |\mathcal{C}|^m, |\mathcal{C}|\ge 2$.
    Let $\mathcal{A}=\{0,1\}^{DB}$ be the state space of $\mm{S}_{i+1}$ and $\mathcal{Q} = \mathrm{Im}(C_{i+1})$ Define the collection of map $\mathcal{G} = \{G_{i,c}(\gamma,\x_{i+1})=\Gamma_{i,c,\gamma}(\x_{i+1}) \mid c \in \mathrm{Im}(C_i)\}$.
    $l = l_m$ and $\theta = 1/(4k)$.
    We show that these choices of parameters satisfy the assumption of Lemma~\ref{lm H construction}.

    We first bound the size of $\mathcal{G}$. By Lemma~\ref{lm state representation}, we know that $\Gamma_{i,c,\gamma} = (H_{i,c}(\x_{i+1}),\Phi_\gamma(E_{i,c}(\x_{i+1})) )$.
    We count all possible $H_{i,c},E_{i,c},\Phi_\gamma$. 
    Since the total number of output bits of $H_{i,c}$ is at most $iDB$, we know that there are at most $2^{iDB n}$ possible $H_{i,c}$.
    Similarly, denote by $s_{i+1}:= b_{i+1}-u_{i+1}+1 = (i+1)\Delta_m + m$, then the total number of output bits of $E_{i,c}$ is $s_{i+1} P$ and thus there are at most $2^{s_{i+1}Pn}$ many $E_{i,c}$.
    Finally, $\Phi$ is determined by the state transition function $\delta:\{0,1\}^{DB} \times \{0,1\}^{P} \to \times \{0,1\}^{DB}$ and has at most $2^{DB2^{DB+P}}$ many possibilities. So the total number of $\Phi$ is at most $2^{s_{i+1}DB2^{DB+P}}$
    This gives
    \begin{align*}
        \log(|\mathcal{G}|) &\le n(iDB+s_{i+1}P) + s_{i+1}DB2^{DB+P}\\
        &\le |\mathcal{C}|^m(km^2+m^{1/\eps+1})+k2^{3c_\tau m}m^{1/\eps+2} \\
        &= o(r_\epsilon\log m (m^{r_\epsilon}+1)|\mathcal{C}|^m/(4k)).
    \end{align*}
    Next, we upper bound the total number of list functions. For any fix $q \in \mathcal{Q}$, the number of subsets of $|\mathcal{K}|$ with size at most $l$ is $\sum_{j=0}^l \binom{|\mathcal{K}|}{j} \le (l+1)(eN/l)^l$. This gives the total number list functions is at most $((l_m+1)(eN_m/l_m)^{l_m})^{|\mathcal{Q}|}$.
    In particular, by taking the logorim, we have 
    \begin{align*}
        \log (((l_m+1)(eN_m/l_m)^{l_m})^{|\mathcal{Q}|}) = |\mathcal{Q}|\log(l_m+1) + l_m|\mathcal{Q}| \log(eN_m/l_m) = o(r_\epsilon\log m (m^{r_\epsilon}+1)|\mathcal{C}|^m/(4k)).
    \end{align*}

    Finally, we verify the assumption of Lemma~\ref{lm H construction}.
    By taking the logarithm, the left hand side is.
    \begin{align*}
        \log(|\mathcal{G}|) + \log(|\mathcal{L}_{l_m}|) + \log(\binom{N_m}{l_m+1}) + (l_m+1)(\log(|\mathcal{A}|)+ n + \theta n \log(l_m/N_m)).
    \end{align*}
    We notice that the negative term is 
    \begin{align*}
     (l_m+1)\theta n \log(l_m/N_m) = -r_\epsilon\log m (m^{r_\epsilon}+1)|\mathcal{C}|^m/(4
     k),  
    \end{align*}
    which dominates the other terms above. Thus, the assumption of Lemma~\ref{lm H construction} is satisfied.
\end{proof}

Now, in the remainder of the proof, we will translate Lemma~\ref{lm chain definition} to the error analysis of a fixed model.
\begin{lemma}\label{lm error analysis}
    For $\eps \in (0,1)$, let $r_\epsilon = \lfloor(1/\eps)\rfloor+3$, $N_m = m^{2r_\epsilon}$, $l_m = m^{r_\eps}$ and
    let $\mathcal{X}_{key} = [N_m]$. Let $H_1,\dots,H_k$ be functions chosen from Lemma~\ref{lm chain definition}. Let $M \in \mathcal{M}_{\bar \alpha}(DB,P,D(W-1))$, an arbitrary hybrid model that satisfies $B,P < c_\tau m$ and $W < c_\tau m^{1/\eps}$, $c_\tau \ll 1$ a constant. For $i \in [k]$, denote by $J_i$ the prefix $(\z_0,\x_1,\dots,\x_i)$ and $\mathrm{err}_i(J_i):=\Pr_{I\sim \mu_m}[M(I) \neq \z_k \mid J_i]$.
    For $i \in [k-1]$, and $c \in \mathrm{Im}(C_i)$, there exists sets $L_i(c) \subseteq \mathrm{X}_{key}$ with size $l_m$ such that for every $J_i$, $\mathrm{err}_i(J_i) \le \delta_{i+1}:= 1/3+i/(6k)$ implies that $\z_i \in L_i(C_i(J_i))$.
\end{lemma}

\begin{proof}[Proof of Lemma~\ref{lm error analysis}]
    We prove the lemma using a backward induction on $i$.
    The base case is $i = k-1$. Notice that if $C_k$ is given, then the output of $M$ is a deterministic function of $C_k$. We define the list function $\mathcal{L}(C_k)$ to be a set that only contains the output of $M$, given $C_k$.
    By applying Lemma~\ref{lm chain definition} to $\mathcal{L}$, we know that  $L_{k-1}(c):=\{z \in \mathcal{X}_{key} \mid \exists \gamma \in \{0,1\}^{DB}, \Pr_{\x_{k}\sim \mathcal{C}^m}[H_{k}(\z,\x_{k}) \in \mathcal{L}(\Gamma_{k-1,c,\gamma}(\x_{k}))] \ge 1/(3k) \}$ has a size at most $l_m$.
    Now fix any $J_{k-1}$ such that $\mathrm{err}_{k-1}(J_{k-1}) \le 1/3+(k-1)/(6k)$. By Lemma~\ref{lm state dependence}, we know that there there exists some $\gamma$ such that $C_k = \Gamma_{k-1,c,\gamma}(\x_k)$, where $C_{k-1}(J_{k-1}) = c$. 
    The assumption that  $\mathrm{err}_{k-1}(J_{k-1}) \le 1/3+(k-1)/(6k)$ implies that $\Pr_{\x_k\sim \mathcal{C}^m}[H_{k}(\z_{k-1},\x_k) \in \mathcal L(\Gamma_{k-1,c,\gamma}(\x_k))] \ge 1/3k$, which implies that $\z_{k-1} \in L_{k-1}(C_{k-1})$.

    Now suppose that we have proven this for $i$, we prove this inductively for $i-1$. For every $c \in \mathrm{Im}(C_{i-1})$. We define $L_{i-1}(c)=\{z \in \mathcal{X}_{key} \mid \exists \gamma \in \{0,1\}^{DB}, \Pr_{\x_{k}\sim \mathcal{C}^m}[H_{i}(\z,\x_{i}) \in L_i(\Gamma_{i-1,c,\gamma}(\x_{i}))] \ge 1/(3k) \}$. By Lemma~\ref{lm chain definition}, we know that $|L_{i-1}(c)| \le l_m$. Now, let $J_{i-1}$ such that $\mathrm{err}_{i-1}(J_{i-1}) \le \delta_i$. 
    Since $\mathrm{err}_{i-1}(J_{i-1}) = \E_{\x_i}[\mathrm{err}_{i}(J_{i-1},\x_i)]$, by Markov's inequality we know that 
    $\Pr_{\x_i}[\mathrm{err}_{i}(J_{i-1},\x_i) > \delta_{i+1}] \le \delta_i/\delta_{i+1}$ and thus
    $\Pr_{\x_i}[\mathrm{err}_{i}(J_{i-1},\x_i) \le \delta_{i+1}] \ge 1- \delta_i/\delta_{i+1}$. By induction when $\mathrm{err}_{i}(J_{i-1},\x_i) \le \delta_{i+1}$, we have $\z_i \in L_i(C_i)$, which means
    \begin{align*}
        \Pr_{\x_i}[H_i(\z_{i-1},\x_i) \in L_i(\Gamma_{i-1,c,\gamma}(\x_i))] \ge 1/(3k).
    \end{align*}
    Thus, $\z_{i-1} \in L_{i-1}(c)$.
\end{proof}

We now give the full proof of Theorem~\ref{th wrong order}.

\begin{proof}[Proof of Theorem~\ref{th wrong order}]
    By Lemma~\ref{lm model simplification}, it suffices to consider a model $M' \in \mathcal{M}_{\bar \alpha}(DB,P,D(W-1))$. Suppose that $M'$ has error at most $1/3$ over $\mu_m$. Then, we have $\E_{J_1}[\mathrm{err}_1(J_1)] \le 1/3.$
    Since $\delta_2 = 1/3+1/(6k)$, by Markov's inequality, we know that $\Pr_{J_1}[e_1(J_1)\le \delta_2] \ge 1/(2k+1)$.
    By Lemma~\ref{lm error analysis}, we know that $z_1 \in L_1(C_1)$. In particular, this implies that $\Pr[\z_1 \in L_1(C_1)] \ge 1/(2k+1)$.
    However, by Lemma~\ref{lm H construction}, we will show that the choice of $H_1$ must also have $\Pr[\z_1 \in L_1(C_1)] < 1/(2k+1)$. To see this, recall that the set of $\z_0 \in \mathcal{X}_{key}$ such that there exists some internal state $\gamma$ such that the probability that $H_1(\z_0,\x_1) \in L_1(\Gamma_{\gamma}(\x_1))$ is more than $1/4k$ has a size at most $l_m$. Denote by this set of keys to be $B$, then we have
    \begin{align*}
        \Pr_{\z_0,\x_1}[\z_1 \in L_1(C_1)] \le \Pr_{\z_0 \in B} + (1/4k) \le l_m/N_m + (1/4k) < 1/(2k+1),
    \end{align*}
    where $m$ is large enough.
    \end{proof}

\subsection{Correct ordering but insufficient layers}\label{app insufficient}

\begin{theorem}\label{th insufficient layers}
    Let $D \in \mathbf{N}_+$ and $\tau =
    X_1^{r_1}X_2^{r_2}\cdots X_h^{r_h} \in \Sigma^D$ be an arbitrary architecture profile with depth $D$. Let $\alpha:=\mathrm{comp}(\tau)$. There exists a constant $\lambda_\tau>1$,
    $C_{\tau}>0,$, a family of task $\{G_{\tau,m}\}_{m\ge 1}$ and a family of distributions $\nu_m$ over a sequence of tokens with length $L_m$ parameterized by $m \in \mathbf{N}_+$, such that for all $m$ sufficiently large, the follows hold.  
    Let $A \in \Sigma^{\le D}$ be an architecture profile with $\alpha \preceq A$ but $\tau \not\preceq A$. Then for every hybrid model $M\in\mathcal M_A(B,P,W)$ that satisfies $\mathrm{err}(M,G_{\tau,m}) \le O(1/D)$, must satisfy $W \ge \lambda_\tau (m+1) $ or $\max\{B,P\} \ge \lambda_\tau C_\tau \log m$.
    On the other hand, there exists a model $M' \in \mathcal{M}_\tau(C_\tau\log m,C_\tau\log m,m+1)$ that can solve $G_{\tau,m}$ exactly.
\end{theorem}

The main proof of Theorem~\ref{th insufficient layers} will focus on the hardness part.
We introduce the following definitions

\begin{definition}\label{def size-2 cover restate}
    Let $\tau \in \Sigma^D$. Let $s(\tau) = |\{i \in [D] \mid \tau_i = \mm{S}\}| $ and  $t(\tau) = |\{i \in [D] \mid \tau_i = \mm{T}\}| $. For $i \in [s(\tau)]$, we use $r_i(\tau)$ to denote the number of $\mm{T}$-layers in $\tau$ that is strictly after the $i$th $S$-layer and for $i \in [t(\tau)]$,  we use $l_i(\tau)$ to denote the number of $\mm{S}$-layers in $\tau$ that is strictly after the $i$th $\mm{T}$-layer. We call the set of architecture profiles 
    \begin{align*}
        \mathcal{C}_2(\tau) = \{\mm{S}^{s(\tau)}\} \cup \{\mm{T}^{t(\tau)}\} \cup \{\mm{S}^i\mm{T}^{r_i(\tau)}\mid i \in [s(\tau)]\} \cup \{\mm{T}^i\mm{S}^{l_i(\tau)} \mid i \in [t(\tau)]\},
    \end{align*}
    the size-2 cover of $\tau$.
\end{definition}

We first give the following property of the size-2 cover.

\begin{lemma}\label{lm size-2 cover}
    Let $D \in \mathbf{N}_+$ and $\tau \in \Sigma^D$. Let $A \in \Sigma^{\le D}$ such that $\tau \not \preceq A$, then there exists some $\sigma \in \mathcal{C}_2(\tau)$ such that $\sigma \not\preceq A$.
\end{lemma}

\begin{proof}[Proof of Lemma~\ref{lm size-2 cover}]
    Suppose that instead for every $\sigma \in \mathcal{C}_2(\tau)$, we have $\sigma \preceq A$. In particular, we have $\mm{S}^{s(\tau)} \preceq A$ and $\mm{T}^{t(\tau)} \preceq A$.  This implies that it must be the case that $|A| = D$ and $s(A) = s(\tau), t(A) = t(\tau)$.
    We notice that once $s(A),t(A)$ and $r_1(A),\dots,r_{s(A)}(A)$ is fixed, $A$ is completely determined.
    Since for $i \in s(\tau)$, $\mm{S}^i\mm{T}^{r_i(\tau)} \preceq A$, we know that $r_i(A) \ge r_i(\tau)$. 
    Notice that the total number of $(i,j)$ such that $i<j$ and $A_i = \mm{T}, A_j = \mm{S}$ is exactly
    \begin{align*}
        \sum_{i=1}^{s(A)} (t(A)-r_i(A)) = \sum_{i=1}^{t(A)} l_i(A).
    \end{align*}
    $A\neq\tau$ implies that at least one of $r_i(A)>r_i(\tau)$, which implies that $\sum_{i=1}^{t(A)} l_i(A) < \sum_{i=1}^{t(A)} l_i(\tau)$ and thus some $l_i(\tau)>l_i(A)$. However, this implies that $\mm{T}^i\mm{S}^{l_i(\tau)} \not\preceq A$.
\end{proof}

Given Lemma~\ref{lm size-2 cover}, we next provide a reduction showing that it is enough to construct a hard task for every architecture profile in the size-2 cover.

\begin{lemma}\label{lm size-2 reduction}
    Let $D \in \mathbf{N}_+$ and $\tau =
    X_1^{r_1}X_2^{r_2}\cdots X_h^{r_h} \in \Sigma^D$ be an arbitrary architecture profile with depth $D$. Let $\mathcal{C}$ be a set of $R$ architecture profiles such that for $\sigma_j \in \mathcal{C}, j \in [R]$, $\sigma_j \preceq \tau$. Furthermore, suppose there exists constant $C>0$ and $\lambda>1$ such that for $j \in [R]$, there exists a family of task $\{H_{j,m}\}$, there exists a model $M' \in \mathcal{M}_{\sigma_j}(C\log m,C\log m,,m+1)$ that can solve $H_{j,m}$ exactly, while every hybrid model $M\in\mathcal M_A(B,P,W)$ that satisfies $\mathrm{err}(M,H_{j,m}) \le 1/3$, must satisfy $W \ge \lambda (m+1) $ or $\max\{B,P\} \ge \lambda C \log m$, if $A \in \Sigma^{\le D}, \sigma_j \not\preceq A$.
    Then, there exists a family of task $\{G_{\tau,m}\}$ such that 
    for every hybrid model $M\in\mathcal M_A(B,P,W)$ that satisfies $\mathrm{err}(M,G_{\tau,m}) \le 1/(3R)$, must satisfy $W \ge \lambda (m+1) $ or $\max\{B,P\} \ge \lambda C \log m$, if $A \in \Sigma^{\le D}$ and $\exists \sigma \in \mathcal{C}, \sigma \not\preceq A$.
    On the other hand, there exists a model $M' \in \mathcal{M}_\tau(C\log m,C\log m,m+1)$ that can solve $G_{\tau,m}$ exactly.
\end{lemma}

\begin{proof}[Proof of Lemma~\ref{lm size-2 reduction}]
We start with the design of the task $G_{\tau,m}$. Let $\mathcal{X}$ be the context space and let $H_{j,m}: \mathcal{X}^{L_m} \to \mathcal{X}$, where $L_m$ is the length of the input sequence. Let $\nu_{j,m}$ be the distribution of the input sequence for task $H_{j,m}$.
We define the distribution $\nu_m$ over $([R]\times\mathcal{X})^{L_m}$ as follows. First, $j \sim [R]$ is drawn uniformly. Given $j$, a sequence $\x \sim \nu_{j,m}$ is drawn and finally the returned sample has $\bar \x \in ([R]\times\mathcal{X})^{L_m}$ has a form $\bar \x_i = (j,\x_i)$, where $i \in [L_m]$. And the task $G_{\tau,m}$ is asked to compute $H_{j,m}(\x)$, given $\bar \x$.

The hardness part follows immediately. Since $M_A \in \mathcal{M}(B,P,W)$ and some $\sigma_j \not \preceq A$, which implies that $M_A$ fails to solve $H_{j,m}$ over $\nu_{j,m}$ with probability $1/3$, this implies that condition $j \sim [R]$ is sampled, there is at least $1/R$ probability that $H_{j,m}$ has to be solved and $M_A$ has a failuare probability $1/(3R)$.

We now design the model $M$ that can be used to solve the task $G_{\tau,m}$. Since the layers we are considering are generalized linear models and non-linear causal window functions, we will just use a selector between layers to select the correct model to use. 
Specifially, let $M_{j,m} = N_{j,m}^{|\sigma_j|}\circ\dots\circ N_{j,m}^{1}$, where $N_{j,m}^i$ is the $i$th layer of $M_{j,m}$, which can solve $H_{j,m}$ exactly.
Importantly, since $\sigma_j \preceq \tau$, we know that there exists a strictly increasing map $P_j:[|\sigma_j|] \to [D]$, such that $\tau_{P_j(i)} = (\sigma_j)_i$, for $i \in [|\sigma_j|]$.
We define the model $M$ as follows for the $i^*$th layer of $M$, it contains identical map and all types of layers $N^i_{j,m}$ if $P_j(i) = i^*$ as blocks and it will select which block to use when read an input $(j,\x_t)$. Specifially, when it reads $(j,\x_t)$, it will choose to use $N^{P_j(i)}_{j,m}$ if $P_j(i) = i^*$ and identical map otherwise. 
By the property of $M_{j,m}$, the model $M$ has 0 error over $\nu_m$.
 Importantly, the structure of $M$ is defined independent on $j$ and only makes uses one of $M_{j,m}$ on a single input sequence and thus is in  $\mathcal{M}_\tau(C\log m,C\log m,m+1)$
\end{proof}

Based on Lemma~\ref{lm size-2 reduction}, it suffices to construct hard tasks for target profiles $\sigma \in \mathcal{C}_2(\tau)$. In the rest of the proof, we establish Lemma~\ref{lm size-2 hardness}.

\begin{lemma}\label{lm size-2 hardness}
    Let $D \in \mathbf{N}_+$ and $\tau =
    X_1^{r_1}X_2^{r_2}\cdots X_h^{r_h} \in \Sigma^D$ be an arbitrary architecture profile with depth $D$.  There exists constant $C>0$ and $\lambda>1$ such that for $j \in [R]$, there exists a family of task $\{H_{\sigma,m}\}$, $\sigma \in \mathcal{C}_2(\tau)$, such that there exists a model $M' \in \mathcal{M}_{\sigma}(b_m,b_m,m+1)$ that can solve $H_{\sigma,m}$ exactly, where $b_m = C \log m$, while every hybrid model $M\in\mathcal M_A(B,P,W)$ that satisfies $\mathrm{err}(M,H_{\sigma,m}) \le 1/3$, for some input distribution $\nu_{\sigma,m}$ must satisfy $W \ge \lambda (m+1) $ or $\max\{B,P\} \ge \lambda b_m$, if $A \in \Sigma^{\le D}, \sigma \not\preceq A$.
\end{lemma}

 Before presenting the proof, we first state some technical lemmas that will be used for the analysis.

\begin{lemma}\label{lm inner product}
    Let $\x,\mathbf{w}$ be two independent vector sampled from $\{0,1\}^a$ for some $a \in \mathbf{Z}_+$. Let $E:\{0,1\}^a \to \{0,1\}^a$ be an embedding function such that $|\mathrm{Im}(E)| \le 2^b$, for some $b<a$, then for every function $\psi: \{0,1\}^a \times \{0,1\}^a \to \{0,1\}$, $\Pr_{\x,\mathbf{w}}[\psi(E(\x),\mathbf{w}) = \oplus_{i=1}^n \x_i\mathbf{w}_i] \le (1/2) +2^{-(a-b)/2-1}$.
\end{lemma}

\begin{proof}[Proof of Lemma~\ref{lm inner product}]
Write $Y=\langle \x,\mathbf{w}\rangle_2 = \oplus_{i=1}^a \x_i\mathbf{w}_i$.
    Let $h \in \mathrm{Im}(E)$ and denote by $S_h:=\{\x \mid E(\x) = h\}$. Let $s_h = |S_h|$. For each $\mathbf{w} \in \{0,1\}^a$, define $\beta_h(\mathbf{w}) := \mathbb{E}_{\x\sim U(S_h)} \left[ (-1)^{\langle\x,\mathbf{w}\rangle_2} \right].$
    Notice that $\beta_h(\mathbf{w}) = \Pr_{\x}[Y = 0 \mid E(\mathbf{x})=h,\mathbf{w}] - \Pr_{\x}[Y = 1 \mid E(\mathbf{x})=h,\mathbf{w}]$. This implies that given $h,\mathbf{w}$, the Bayesian optimal prediction has an error of $(1+ |\beta_h(\mathbf{w})| )/2.$ 

    Taking the average over $\mathbf{w}$, we have
    \begin{align*}
        \Pr_{\x,\mathbf{w}}[\psi(E(\x),\mathbf{w}) & = \oplus_{i=1}^a \x_i\mathbf{w}_i \mid E(\x) = h] \le (1+ \E_{\mathbf{w}}[|\beta_h(\mathbf{w})|])/2 \\
       & \le (1+ \sqrt{\E_{\mathbf{w}}[|\beta_h(\mathbf{w})|^2]})/2 =(1+1/\sqrt{s_h})/2
    \end{align*}
    Now we take the average over $h$ and get
    \begin{align*}
        \Pr_{\x,\mathbf{w}}[\psi(E(\x),\mathbf{w}) & = \oplus_{i=1}^n \x_i\mathbf{w}_i] \le \frac{1}{2} + \frac{1}{2^{a+1}}\sum_{h}\sqrt{s_h} \le \frac{1}{2} + \frac{1}{2^{a+1}}\sqrt{|\mathrm{Im}(E)|\sum_h s_h} \\
        & \le (1/2) +2^{-(a-b)/2-1}.
    \end{align*}
\end{proof}

\begin{lemma}\label{lm random list}
 Let $n,K \in \mathbf{Z}_+$.    Let $\x^{(1)},\dots,\x^{(K)} $ be drawn uniform at random and independently from $\{0,1\}^n$ and let $J$ be drawn uniformly from $[K]$. Let $\Gamma$ be a discrete function over $(\{0,1\}^n)^K$ that takes at most $2^a$ values, for some $a \in \mathbf{Z}_+$. Suppose that for every $\gamma \in \mathrm{Im}(\Gamma), j \in [K]$, there exists a set $L_{\gamma,j} \subseteq \{0,1\}^n$ with size at most $l < 2^n$.
 Then we have
 \begin{align*}
     \Pr_{\x^{(1)},\dots,\x^{(K)},J}[\x^{(J)} \in L_{\Gamma(\x^{(1)},\dots,\x^{(K)}),J}] \le \frac{a/K+1}{n-\log l}.
 \end{align*}
\end{lemma}

\begin{proof}[Proof of \Cref{lm random list}]
    Notice that 
    \begin{align*}
        \sum_{i=1}^K I(\x^{(i)};\Gamma) & = \sum_{i=1}^K H(\x^{(i)}) - \sum_{i=1}^K H(\x^{(i)}\mid \Gamma) \\
        & \le Kn - H(\x^{(1)},\dots,\x^{(K)} \mid \Gamma) \\
        & = I(\x^{(1)},\dots,\x^{(K)}; \Gamma) \\
        & \le H(\Gamma) \le a.
    \end{align*}
    This implies that 
    \begin{align*}
     I(\x^{(J)};\Gamma,J) &= I(\x^{(J)};J)+I(\x^{(J)};\Gamma \mid J) \\ & = \frac{1}{K}\sum_{i=1}^K I(\x^{(i)};\Gamma) \\ & \le a/K.   
    \end{align*}
    Denote the joint random variable $R = (\Gamma,J)$. We show that $p:=\Pr[\x^{(J)} \in L_R] \le (I(\x^{(J)};R)+1)/(n-\log l)$.
    Define random variable $B = \mathbf{1}\{\x^{(J)} \in L_R\}$, we have 
    \begin{align*}
        H(\x^{(J)},B \mid R) = H(\x^{(J)} \mid R) + H(B \mid \x^{(J)} R) = H(\x^{(J)} \mid R).
    \end{align*}
    This gives
    \begin{align*}
        H(\x^{(J)} \mid R) & = H(\x^{(J)},B \mid R) \\ 
        &= H(B \mid R) + H(\x^{(J)}\mid B, R) \\
        & \le 1 + H(\x^{(J)}\mid B, R) = 1 + pH(\x^{(J)}\mid B=1, R) + (1-p)H(\x^{(J)}\mid B=0, R) \\
        & \le 1+ p \log l + (1-p) n = n+1 -p(n-\log l).
    \end{align*}
    Thus, we have $I(\x^{(J)};R) \ge p(n-\log l) - 1$, which gives $\Pr[\x^{(J)} \in L_R] \le (I(\x^{(J)};R)+1)/(n-\log l)$. Since $I(\x^{(J)};R) \le a/K$, we get
     \begin{align*}
     \Pr_{\x^{(1)},\dots,\x^{(K)},J}[\x^{(J)} \in L_{\Gamma(\x^{(1)},\dots,\x^{(K)}),J}] \le \frac{a/K+1}{n-\log l}.
 \end{align*}
\end{proof}

\begin{lemma}[List Decoding for Hadamard Code]\label{lm list decoding}
    Let $f:\{0,1\}^n \to \{0,1\}$ be an arbitrary Boolean function and let $L:=\{\x \in \{0,1\}^n \mid \Pr_{\mathbf{u}\sim U(\{0,1\}^n)}[f(\mathbf{u}) = \langle\x,\mathbf{u}\rangle_2 ]>\eps + 1/2\}$, $\eps \in (0,1)$. Then, $|L| \le 1/(4\eps^2)$.
\end{lemma}

\begin{proof}[Proof of Lemma~\ref{lm size-2 hardness}]
    We will break the proof into 4 cases to consider the cases where $\sigma$ has the form of $\mm{S}^r,\mm{T}^r,\mm{T}^r\mm{S}^s, \mm{S}^r\mm{T}^s$, with $r+s \le D$. Let $\lambda = 1+ (1/4D).$
    \paragraph{Case 1: $\sigma=\mm{S}^r$}
    In the first case, we consider the case where $\sigma = \mm{S}^r$.
    Let $a_m = b_m -1$ and let $n = ra_m$.
    We design the input distribution $\nu_{\mm{S}^r,m}$ as follows. To draw an input $I$ from $\nu_{\mm{S}^r,m}$, we first draw $\x,\mathbf{w}$ uniform at random independently from $\{0,1\}^n$ and return $I = (\x,\vec 0, \mathbf{w})$, where $\vec 0$ is sequence of $0$ of length at least $D\lambda(m+1)$.
    The task $H_{\mm{S}^r,m}(I)$ asks to compute $Y=<\x,\mathbf{w}>_2 = \oplus_{i=1}^n \x_i\mathbf{w}_i$.
    By Lemma~\ref{lm group ssm}, we show that a single layer SSM with $2^{rb_m}$ states can solve this task. In fact, this can be done since $2^{rb_m}$ bits are enough to memorize $\x,\mathbf{w}$ exactly.

    We next show that if $\mm{S}^r \not \preceq A$, then a model from $\mathcal{M}_A(B,P,W)$ has a bad accuracy unless $B,P,W$ are large.
    Since $\mm{S}^r \not\preceq A$, we know that $A$ contains at most $r-1$ SSM layers. On the other hand, since $\vec 0 $ has length at least $D\lambda(m+1)$, we know that if $W<\lambda(m+1)$, then the $\mm{T}$-layers have no direct access to $\x$. This implies that there exist maps $\psi$ and $E$ such that $M(I) = \psi(E(\x),\mathbf{w})$, with $\log (|\mathrm{Im}(E)|) \le (r-1)\lambda b_m$.
    Since $n - (r-1)\lambda b_m \ge \Omega_D(\log m)$, Lemma~\ref{lm inner product}, we know that $M$ has a success probability at most $2/3$, when $m$ is large enough.
\paragraph{Case 2: $\sigma=\mm{T}^r$}

    In the second case, we consider the case where $\sigma = \mm{T}^r$. Let $n = 2Db_m$. We design the input distribution $\nu_{\mm{T}^r,m}$ as follows. To draw an input $I$ from $\nu_{\mm{T}^r,m}$, we first draw $\x,\mathbf{w}$ uniform at random independently from $\{0,1\}^n$ and return $I = (\x,\vec 0, \mathbf{w})$, where $\vec 0$ is sequence of $0$ of length  $\lambda(r-1)m + n<\Delta<rm$ (by the choice of $\lambda$, such $\Delta$ exists).
    The task $H_{\mm{T}^r,m}(I)$ asks to compute $Y=\langle\x,\mathbf{w}\rangle_2 = \oplus_{i=1}^n \x_i\mathbf{w}_i$.
    Since a $\mm{T}$-layer can be any causal nonlinear function with window size $m+1$,
     by Lemma~\ref{lm group tf}, we can design some $M \in \mathcal{M}_{\mm{T}^r}(b_m,b_m,m+1)$ such that it first move $\x$ across from $\vec 0$ and computes $Y$.

      We next show that if $\mm{T}^r \not \preceq A$, then a model from $\mathcal{M}_A(B,P,W)$ has a bad accuracy unless $B,P,W$ are large.
    Since $\mm{T}^r \not\preceq A$, we know that $A$ contains at most $r-1$ $\mm{T}$-layers, by the choice of $\Delta$, we know that no after $\vec 0$, no $\mm{T}$ layer can reach $\x$ directly. Since there are at most $\SSM$-layers, 
    this implies that there exist maps $\psi$ and $E$ such that $M(I) = \psi(E(\x),\mathbf{w})$, with $\log (|\mathrm{Im}(E)|) \le D\lambda b_m$.
    Since $n - D\lambda b_m\ge \Omega_D(\log m)$, Lemma~\ref{lm inner product}, we know that $M$ has a success probability at most $2/3$, when $m$ is large enough.
\paragraph{Case 3: $\sigma=\mm{T}^r\mm{S}^s$}

    In the third case, we consider the case where $\sigma = \mm{T}^r\mm{S}^s$, with $r+s \le D$.
        Let $a_m = b_m -1$, $n = sa_m$, and $K = 32D$
    We design the input distribution $\nu_{\mm{T}^r\mm{S}^s,m}$ as follows. To draw an input $I$ from $\nu_{\mm{T}^r\mm{S}^s,m}$, we first draw $\x^{(1)},\dots,\x^{(K)}$ and $\mathbf{w}$ uniform at random independently from $\{0,1\}^n$ and draw some $J$ from $[K]$ uniform at random. 
    The sampled sequence $I$ combines $\x^{(1)},\dots,\x^{(K)}, J$ and $\mathbf{w}$ with two $0$ vectors with different lengths.
    Specifically, we put $\x^{(1)},\dots,\x^{(K)}$ at the first $Kn$ positions.
    We append $\x^{(1)},\dots,\x^{(K)}$ with a sequence of $0$ vectors with length $\Delta_1$ that satisfies $\Delta_1+1 >(r-1)(\lfloor \lambda(m+1) \rfloor -1)$ and $Kn+\Delta_1 + n - 1 \le rm$.
    
    We explain the choice of such $\Delta_1$ before proceeding. 
    Notice that the position of $j$ is $q= Kn +\Delta_1+1$, and if $W < \lambda(m+1)$ then at the position $q$, a $r-1$-layer $\mm{T}$ can read input at most $(r-1)(\lfloor \lambda(m+1) \rfloor -1)$ from $q$.
    $\Delta_1+1 >(r-1)(\lfloor \lambda(m+1) \rfloor -1)$ implies that starting from position $q$ no information from $\x^{(1)},\dots,\x^{(K)}$ is known by $\mm{T}^{r-1}$.
    On the other hand, a $\mm{T}^{r-1}$ is able to copy $\x^{(1)},\dots,\x^{(K)}$ to position $[1+(r-1)m,Kn+(r-1)m]$. We would like to require a single layer of $\mm{T}$ with $m+1$ window size to see $1+(r-1)m$ at position $q+n-1$, which implies that $Kn +\Delta_1+ n -m \le 1+(r-1) m$. i.e. $Kn +\Delta_1 +n-1 \le rm$. Notice that $rm-(r-1)\lambda m = \Omega(m)$. So, such a choice of $\Delta_1$ exists.
    
    Then, after adding $J$ to position $Kn+\Delta_1+1$, we append another $\vec 0$ with size $D(\lfloor \lambda(m+1) \rfloor -1)$ and append $\mathbf{w}$ after this $\vec 0$.

    The existence of a model $M \in \mathcal{M}_{\mm{T}^r\mm{S}^s} (b_m,b_m,m+1)$ that can solve the task exactly follows from the construction of $I$. The first $r-1$ layer $\mm{T}$ can move $\x^{(1)},\dots,\x^{(K)}$ to 
    $[1+(r-1)m,Kn+(r-1)m]$ so that another $\mm{T}$ layer has an exact access to both $\x^{(1)},\dots,\x^{(K)}$ and $J$, so can write $x^{(J)}$ after position $q$. Then, an $s$-layer $\SSM$ with space size $2^{b_m}$ can memorize $x^{(J)}$ and compute $\langle\x^{(J)},\mathbf{w}\rangle_2$.

    It remains to show the lower bound part. Let $M \in \mathcal{M}_A(B,P,W)$ be a model such that $\mm{T}^r\mm{S}^s \not \preceq A$. 
    To simplify the terminology, for an $\mm{S}$-layer that appears after $r$ $\mm{T}$-layers, we call it a later $\mm{S}$ layer, otherwise, we call it an early $\mm{S}$-layer. Denote by $\mm{S}^{(e)}$ the set of early $\mm{S}$-layers. Let $E$ be the set of states of $\mm{S}$ layers in $\mm{S}^{(e)}$ evaluated before position $q$ i.e. before $j$ is read. Similarly, denote by $\mm{S}^{(l)}$ the set of late $\mm{S}$-layers and let $Z$ be the set of states of $\mm{S}$ layers in $\mm{S}^{(l)}$ evaluated before $\mathbf{w}$ is read. 
    
    By the design of $\Delta_1$, we know that $E$ is a deterministic function of $\x^{(1)},\dots,\x^{(K)}$, while conditioned on $E$, starting from position $q$, the states of layers in $\mm{S}^{(e)}$ is a deterministic function of $(j,\mathbf{w})$.
    In particular, we know that $E$ can take at most $2^{D\lambda b_m}$ values. Similarly, we know that conditioning on $Z$, the states of layers in $\mm{S}^{(l)}$ after reading $\mathbf{w}$ is a deterministic function of $\mathbf{w}$. In particular, $Z$ can take at most $2^{(s-1) \lambda b_m}$ values, otherwise, $\mm{T}^r\mm{S}^s \preceq A$.

    Together, this implies that the final output $M = \psi(E,Z,j,\mathbf{w})$ for some deterministic function $\psi$.
    Now for $\gamma \in \mathrm{Im}(E), j \in [K]$ and $\z \in \mathrm{Im}(Z)$, we denote by 
    \begin{align*}
        L_{\gamma,j,\z} = \{\mathbf{u} \in \{0,1\}^n \mid \Pr_{\mathbf{w}}[\psi(\gamma,\z,j,\mathbf{w}) =\langle\mathbf{u},\mathbf{w}\rangle_2] \ge (7/12)\}.
    \end{align*}
By Lemma~\ref{lm list decoding}, we know that $|L_{\gamma,j,\z}| \le 36$. This implies that if we denote by $L_{\gamma,j} = \cup_{\z \in \mathrm{Im}(Z)}$, then we have $|L_{\gamma,j}| \le 36 2^{(s-1) \lambda b_m}$.
We now apply Lemma~\ref{lm random list} to $L_{\gamma,j}$ and $\x^{(1)},\dots,\x^{(K)}$ and get $\Pr_{\x^{(1)},\dots,\x^{(k)},J}[\x^{(J)} \in L_{\Gamma(\x^{(1)},\dots,\x^{(k)}),J}] < 1/5$.
We conclude of the proof of the case by analyzing the probability of failure of $M$.
Let $p$ be the random variable of $\Pr_{\mathbf{w}}[M(I) = \langle \x^{(J)},\mathbf{w} \rangle \mid \x^{(1)},\dots,\x^{(K)}, J]$.
Assume by contradiction that $\E[p] \ge 2/3$, then we have 
\begin{align*}
    2/3 \le \E[p] &= \E[p \mid p<7/12] \Pr_{\x^{(1)},\dots,\x^{(K)}, J}[p<7/12] + \E[p \mid p\ge 7/12] \Pr_{\x^{(1)},\dots,\x^{(K)}, J}[p \ge 7/12] \\
    & \le 7\Pr_{\x^{(1)},\dots,\x^{(K)}, J}[p<7/12]/12 + \Pr_{\x^{(1)},\dots,\x^{(K)}, J}[p \ge 7/12] \\
    & \le 7/12 + \Pr_{\x^{(1)},\dots,\x^{(K)}, J}[p \ge 7/12].
\end{align*}
    This implies that $\Pr_{\x^{(1)},\dots,\x^{(K)}, J}[p \ge 7/12] \ge 1/5$. But $\Pr_{\x^{(1)},\dots,\x^{(K)}, J}[p \ge 7/12] \le \Pr_{\x^{(1)},\dots,\x^{(k)},J}[\x^{(J)} \in L_{\Gamma(\x^{(1)},\dots,\x^{(k)}),J}] < 1/5$, which gives a contradiction.
    Thus, $M$ must have a failure probability $1/3$.

\paragraph{Case 4: $\sigma=\m{S}^r\m{T}^s$}

The proof for the case where $\sigma=\mm S^r\mm T^s$ is more technical than the other three
cases. We refer to Lemma~\ref{lem:srts-component} for a detailed proof.

\end{proof}

\paragraph{Analysis of Case 4: $\sigma=\mm S^r\mm T^s$.}
The case $\sigma=\mm S^r\mm T^s$ is more technical than the other three cases.
We isolate the required construction in the following lemma.
For the common constant in Lemma~\ref{lm size-2 hardness}, we take $\lambda>1$
to be the minimum of the constants obtained in the four cases. Due to the length and the techniques involved in the proof, in the rest of the section, we will give a brief and compressed version of the proof. We refer the reader to the supplementary materials for a complete rigorous proof.

\begin{lemma}\label{lem:srts-component}
Let $D\in\mathbb N_+$. There exist constants $C>0$ and $\lambda>1$ such that for every
$r,s\in\mathbb N_+$ with $r+s\le D$, there exists a family of tasks $\{F_m\}_{m\ge1}$
with the following properties. There exists a model
$M'\in\mathcal M_{\mm S^r\mm T^s}(b_m,b_m,m+1)$, where
$b_m:=\lceil C\log_2m\rceil$, that solves $F_m$ exactly. On the other hand, if
$A\in\Sigma^{\le D}$ satisfies $\mm S^r\mm T^s\not\preceq A$, then every
$M\in\mathcal M_A(B,P,W)$ with error at most $1/3$ on the input distribution $\nu_m$
must satisfy $W\ge\lambda(m+1)$ or $\max\{B,P\}\ge\lambda b_m$.
\end{lemma}

We first introduce the notation used in the proof.
Fix $D\in\mathbb N_+$ and let $d:=\max\{D,5\}$.
We choose the finite constants used in the induction below first, and then choose
$\kappa_D>0$ sufficiently large depending only on $D$.
Set $C_D:=\kappa_D+10d$ and $\lambda_D:=1+d/\kappa_D$.
We require $\kappa_D>4dD$, so $1<\lambda_D<1+1/(4D)$.

For sufficiently large $m$, define
$a_m:=\lfloor\kappa_D\log_2m\rfloor$, $K:=2^{a_m}$,
and $b_m:=\lceil C_D\log_2m\rceil$.
We further put $B_*:=P_*:=\lfloor\lambda_Db_m\rfloor$,
$S_*:=2^{B_*}$, $Q_*:=2^{P_*}$,
$W_*:=\lfloor\lambda_D(m+1)\rfloor$, and $w:=W_*-1$.
For fixed $r,s\ge1$ with $r+s\le D$, let $q:=m-r$.
The chain task uses a source vector $\mathbf z_0\in[K]^r$ and independent query
blocks $\mathbf x_1,\ldots,\mathbf x_s\in[K]^q$.

The proof of Lemma~\ref{lem:random-chain} fixes constants
$N_1,\ldots,N_d>0$, depending only on $D$, and sets
$n_j:=\lfloor N_jm/\kappa_D\rfloor$.
They are chosen so that $n_j=\Theta_D(m)$, every replay cut below lies strictly
inside its corresponding query block, and consecutive replay cuts are separated by
at least $w$. In particular, we require $N_{j+1}>N_j+d$ for $j\in[d-1]$.

Define
$L_0:=\lfloor K^r/(m^{10d}S_*^{r-1})\rfloor$.
By taking $\kappa_D$ sufficiently large, for all sufficiently large $m$,
\begin{align*}
\frac{K}{m^{12d^2}}
\le L_0
\le \frac K2,
\qquad
\frac{m^2}{L_0}
=o(1).
\end{align*}

We call a consecutive block of one or more $\mm S$-layers a recurrent run.
By Lemma~\ref{lm group ssm}, a recurrent run may be viewed as one
$\mm S$-layer whose state space is the product of the state spaces of the layers
in the run. We next formalize the residual computation obtained after cutting a
competing model inside the guard.

\begin{definition}[Physical $j$-stage replay]\label{def:physical-replay}
Fix lookup tables $\mathcal H=(H_1,\ldots,H_s)$, a stage number $1\le j\le d$,
an incoming first-query length $n\in[n_j,q]$, a finite source set $\mathcal K$,
a source variable $\boldsymbol\zeta\in\mathcal K$, and a fixed old context $c$.
Let $\mathbf Y_1\in[K]^n$ and
$\mathbf Y_2,\ldots,\mathbf Y_j\in[K]^q$ be the raw query blocks.
Write
$\overline{\mathbf Y}:=(\mathbf Y_1,\ldots,\mathbf Y_j)
=(Y^1,\ldots,Y^{L_j(n)})$, where $L_j(n):=n+(j-1)q$.

The replay has exactly these $L_j(n)$ query positions.
It first forms a physical token sequence
$\mathbf X^{(0)}=(X^{(0)}_1,\ldots,X^{(0)}_{L_j(n)})\in[Q_*]^{L_j(n)}$
causally from the raw queries, where
$X^{(0)}_t=E_t(Y^{1:t};\mathcal H,c)$.
Thus $E_t$ may depend on the lookup tables, the old context, and every raw query
symbol that has arrived by position $t$, but not on the particular source value
$\boldsymbol\zeta$. No additional token positions are introduced.

The continuation consists of $j$ recurrent runs
$R_1,\ldots,R_j$, separated by $j-1$ $\mm T$-layers
$T_1,\ldots,T_{j-1}$. Every $\mm T$-layer has window at most $W_*$,
every intermediate token belongs to $[Q_*]$, and, if $\mathcal S_i$ denotes
the state space of $R_i$, we require
$\prod_{i=1}^j|\mathcal S_i|\le S_*^d$.

The first recurrent run may start from an arbitrary post-table state
$s^{(1)}_0=\phi(\boldsymbol\zeta)\in\mathcal S_1$.
For $1\le i<j$, let $t_i(n):=n+iq-n_{j-i}$.
Immediately before $R_{i+1}$ processes position $t_i(n)+1$, its state may be set to
$s^{(i+1)}_{t_i(n)}=\Xi_i$, where
$\Xi_i:=\psi_i(\boldsymbol\zeta,U_0,Y^{1:t_i(n)};\mathcal H,c)$.
Here $U_0$ is one common auxiliary seed independent of
$(\boldsymbol\zeta,\overline{\mathbf Y})$ and shared by all cuts.

After this initialization, $R_{i+1}$ follows its declared $\mm S$-layer transition
and emission rules. In particular, $\Xi_i$ may depend on the source and on every
raw query symbol that has arrived by its cut, but not on any symbol after the cut.
We denote the output of the final recurrent run at the last physical position by
$\widehat{\mathbf z}$.

The definition provides no additional advice tape: only the original $O_D(m)$
physical positions, one bounded token at each position, and the original raw-input
arrival schedule are available.
\end{definition}

The main technical ingredient for Case~4 is the following random-chain lemma.

\begin{lemma}[Random-chain lemma]\label{lem:random-chain}
There exist a constant $c_D>0$ and functions
$H_1:[K]^r\times[K]^q\to[K]$ and
$H_i:[K]\times[K]^q\to[K]$ for $2\le i\le s$ such that every
$H_i(\cdot,\mathbf x)$, $i\ge2$, is a permutation of $[K]$, and the following
hold simultaneously for every $F\subseteq[K]^r$ with $L_0\le|F|\le K$.

Let $\mathbf z_0$ be uniform on $F$, and let
$\mathbf x_1,\ldots,\mathbf x_s$ be independent and uniform on $[K]^q$.
Define $\mathbf z_1:=H_1(\mathbf z_0,\mathbf x_1)$ and
$\mathbf z_i:=H_i(\mathbf z_{i-1},\mathbf x_i)$ for $2\le i\le s$.
Then:
\begin{enumerate}
    \item[(i)] for all but an $m^{-100d}$ fraction of
    $\mathbf x_1\in[K]^q$, every fiber of
    $\mathbf z\mapsto H_1(\mathbf z,\mathbf x_1)$ on $F$
    has cardinality at most $m^2$;

    \item[(ii)] every physical $s$-stage replay in
    Definition~\ref{def:physical-replay} with
    $\mathcal K=F$, $\boldsymbol\zeta=\mathbf z_0$, and full incoming
    first query $n=q$ satisfies
    $\Pr[\widehat{\mathbf z}=\mathbf z_s]\le m^{-c_D}+o(1)$.
\end{enumerate}
The same table realization works uniformly over all eligible source sets $F$,
fixed old contexts, and post-table physical replay maps.
\end{lemma}

\begin{proof}[Proof of Lemma~\ref{lem:srts-component}]
Let $C_D,\lambda_D,\kappa_D$ and
$\mathcal H=(H_1,\ldots,H_s)$ be as in Lemma~\ref{lem:random-chain}.
We take $C=C_D$ and $\lambda=\lambda_D$.
Recall that $q=m-r$ and $K=2^{a_m}$.
We construct $F_m$ as the equal mixture of two tagged tasks.

\paragraph{The chain task.}
Draw $\mathbf z_0\sim\operatorname{Unif}([K]^r)$ and
$\mathbf x_1,\ldots,\mathbf x_s
\stackrel{\mathrm{iid}}{\sim}\operatorname{Unif}([K]^q)$.
Define $\mathbf z_1:=H_1(\mathbf z_0,\mathbf x_1)$ and
$\mathbf z_i:=H_i(\mathbf z_{i-1},\mathbf x_i)$ for $2\le i\le s$.
The input is
$I=(\mathbf z_0,\mathbf 0^{\,4D(m+1)},\mathbf 0^{\,r},
\mathbf x_1,\ldots,\mathbf x_s)$.
We call the first zero block the guard and the second zero block the scratch block.
The desired output is $\mathbf z_s$.

\paragraph{The reach task.}
Let $k_0:=\lfloor m/2\rfloor$.
Draw $\mathbf u_1,\ldots,\mathbf u_{k_0}$ independently and uniformly from $[K]$,
and independently draw $J\sim\operatorname{Unif}([k_0])$.
Place $\mathbf u_1,\ldots,\mathbf u_{k_0}$ at the beginning of the sequence and
$J$ at position $sm+1$, and ask the model to output $\mathbf u_J$.
We use fixed left padding so that the two branches have the same length and include
the branch tag in every token. Let $\nu_m$ denote the resulting equal mixture.

\paragraph{Completeness.}
We first show that $\mm S^r\mm T^s$ solves both branches exactly.
On the chain branch, write
$\mathbf z_0=(z_{0,1},\ldots,z_{0,r})$ and number the $\mm S$-layers from
bottom to top. The $i$th $\mm S$-layer copies its input until it reads
$z_{0,r-i+1}$, stores this coordinate across the guard, releases it at scratch
position $r-i+1$, and then resumes copying.
At each release position, every higher $\mm S$-layer has already released its own
coordinate and acts as a copying layer. Hence, after the $r$ $\mm S$-layers,
the scratch block contains $\mathbf z_0$ in its original order, while the query
blocks remain unchanged.

Since $r+q=m$, the first $\mm T$-layer sees
$(\mathbf z_0,\mathbf x_1)$ inside one window of size $m+1$ and computes
$\mathbf z_1$. Each later $\mm T$-layer similarly sees
$(\mathbf z_{i-1},\mathbf x_i)$ and computes $\mathbf z_i$.
Thus the final output is $\mathbf z_s$.

On the reach branch, all $\mm S$-layers act as identities.
The first $s-1$ $\mm T$-layers shift the database to the right by $m$ positions
per layer, and the final $\mm T$-layer sees the requested entry together with
$J$ and outputs $\mathbf u_J$.
The counters, position/type tags, release modes, and branch tag require only
$O_D(\log m)$ bits, which are absorbed into $b_m$.

\paragraph{Guard reduction.}
Consider any prefix of a competing architecture containing $h$ $\mm S$-layers.
Place a cut immediately after the source vector $\mathbf z_0$.
When crossing an $\mm S$-layer, record its state at the current cut.
When crossing an $\mm T$-layer, move the cut to the right by at most $w$ positions.
Since the architecture has depth at most $D$ and $Dw<4D(m+1)$, every such cut
remains inside the guard.

Consequently, all recorded $\mm S$-states depend only on $\mathbf z_0$.
Their tuple defines a map $\Gamma:[K]^r\to\mathcal G$ with
$|\mathcal G|\le S_*^h$. For $\gamma\in\mathcal G$, let
$F_\gamma:=\Gamma^{-1}(\gamma)$.
After fixing $\Gamma(\mathbf z_0)=\gamma$, replaying the prefix from the recorded
states shows that every prefix-output token on the original scratch and query
positions is a common causal function of the raw query symbols that have arrived
by that position, independent of the particular $\mathbf z_0\in F_\gamma$.

Fix $A\in\Sigma^{\le D}$ with
$\mm S^r\mm T^s\not\preceq A$ and
$M\in\mathcal M_A(B_*,P_*,W_*)$.
We distinguish three cases.

\paragraph{Case 1: $A$ contains fewer than $s$ $\mm T$-layers.}
We use the reach task.
Starting from the final query position and tracing the dependence cone backward,
every $\mm T$-layer moves the cut to the left by at most $w$, while every
$\mm S$-layer contributes only its state across the cut.
Since there are at most $s-1$ $\mm T$-layers,
\begin{align*}
sm-(s-1)w
&=
\bigl(s-(s-1)\lambda_D+o(1)\bigr)m
>
\frac m2 .
\end{align*}
Hence the backward dependence cone does not reach the database positions.

The final prediction is therefore a function of $J$ and a message
$\Gamma(\mathbf u_1,\ldots,\mathbf u_{k_0})$ containing at most $DB_*$ bits.
By independence and entropy subadditivity,
$I(\mathbf u_J;\Gamma,J)\le DB_*/k_0$.
The standard error-indicator/Fano estimate gives
$p\le(DB_*/k_0+1)/\log_2K=o(1)$.
Thus the reach-task success is $o(1)$.

\paragraph{Case 2: $A$ contains at least $s$ $\mm T$-layers but fewer than
$r$ $\mm S$-layers.}
Apply the guard reduction to the entire model.
There are at most $S_*^{r-1}$ source fibers.
By the definition of $L_0$, the total source mass in fibers of size less than
$L_0$ is at most $m^{-10d}$.
Partition every remaining fiber into chunks $F$ satisfying
$L_0\le|F|<2L_0\le K$.

Fix one such chunk.
Conditional on $\mathbf z_0\in F$ and on the query blocks, the output of the
competing model is independent of the particular choice of $\mathbf z_0\in F$.
By Lemma~\ref{lem:random-chain}(i), outside an $m^{-100d}$ fraction of
first-query blocks, every fiber of
$\mathbf z_0\mapsto H_1(\mathbf z_0,\mathbf x_1)$ on $F$ has size at most $m^2$.
Since every later $H_i(\cdot,\mathbf x_i)$ is a permutation, the same bound holds
for $\mathbf z_0\mapsto\mathbf z_s$. Therefore,
\begin{align*}
\Pr[\textnormal{success}\mid F]
&\le
m^{-100d}+\frac{m^2}{|F|}
\le
m^{-100d}+\frac{m^2}{L_0}
=
o(1).
\end{align*}
Adding the discarded source mass gives chain-task success $o(1)$.

\paragraph{Case 3: $A$ contains at least $r$ $\mm S$-layers and at least
$s$ $\mm T$-layers.}
Cut $A$ immediately before its $r$th $\mm S$-layer.
The strict prefix contains exactly $r-1$ $\mm S$-layers.
By the guard reduction, after discarding the same negligible small-fiber mass and
fixing a retained source chunk $F$, the prefix outputs on the original scratch and
query positions are common causal functions of the arrived raw query symbols and
are independent of the particular source $\mathbf z_0\in F$.

Write the residual profile as
$\mm S^{b_0}\mm T\mm S^{b_1}\mm T\cdots\mm T\mm S^{b_j}$,
where $b_0\ge1$ and $b_i\ge0$.
Since $A$ avoids $\mm S^r\mm T^s$, we have $j\le s-1$.
Moreover, since $A$ contains at least $s$ $\mm T$-layers and has depth at most
$D$, the number
$B_{\mathrm{res}}:=\sum_i b_i$ satisfies
$B_{\mathrm{res}}\le D-r-s+1$.

Merge every nonempty recurrent run into one product-state $\mm S$-layer.
Insert identity $\mm T$-layers until there are exactly $s-1$ $\mm T$-layers,
and insert a copying $\mm S$-layer into every empty recurrent slot.
A copying $\mm S$-layer uses at most $Q_*=S_*$ states.
Since at most $s-1$ such layers are inserted,
\begin{align*}
S_*^{B_{\mathrm{res}}+s-1}
\le
S_*^{D-r}
\le
S_*^d .
\end{align*}
The padded displayed profile need not itself have depth at most $D$.
The original depth bound has already been used in the guard reduction, while
Definition~\ref{def:physical-replay} only requires the recurrent-state product
bound together with the original token and window bounds.

It remains to verify that the padded residual is a physical $s$-stage replay.
Put the actual state of the first residual recurrent run immediately before the
first query position into $\phi(\mathbf z_0)$.
Its input tokens on the original query positions are the source-independent
causal tokens provided by the guard reduction.
If $s=1$, this already gives the one-stage replay.

Suppose $s\ge2$.
Let $t_1:=2q-n_{s-1}$ and $t:=t_1-w$.
The parameter choices give $0<t<q$.
Thus, for every payload-relative position $p\ge t_1+1$, the local window
immediately before the second recurrent run starts at $p-w\ge t+1>0$.
Hence this window uses only outputs produced on query positions and cannot reach
the guard or scratch block.
All earlier source-dependent effects are therefore contained in the actual state
of the second recurrent run at $t_1$, which is a legal granted state in
Definition~\ref{def:physical-replay}.

Every later prescribed cut is separated from the previous one by at least $w$.
The same argument therefore iterates.
Each relevant local window lies entirely after the preceding cut, and every
earlier source-dependent effect is contained in the actual state of the next
recurrent run. Each such state depends only on $\mathbf z_0$, the common latent
old-source information, and the raw query prefix already available at its cut.
Thus the padded residual is a physical $s$-stage replay with full incoming first
query $n=q$.

Lemma~\ref{lem:random-chain}(ii) gives chain-task success $o(1)$ on every
retained source chunk.

\paragraph{Conclusion.}
Every profile avoiding $\mm S^r\mm T^s$ falls into one of the three cases above.
In Case~1 its reach-task success is $o(1)$, while in Cases~2--3 its chain-task
success is $o(1)$. We make no claim on the other branch and upper bound its
success by one. Hence the total success is at most $1/2+o(1)$, and the error is
at least $1/2-o(1)>1/3$ for sufficiently large $m$.
For an input-independently randomized model, fix its random seed, apply the
deterministic bound, and average over the seed.
\end{proof}

\begin{proof}[Proof of Lemma~\ref{lem:random-chain}]
We use the probabilistic method.
Sample $H_1$ by choosing all of its entries independently and uniformly from
$[K]$. Independently, for every $i\ge2$ and every $\mathbf x\in[K]^q$,
sample $H_i(\cdot,\mathbf x)$ uniformly from the permutations of $[K]$.
We show that with positive probability the sampled tables satisfy all required
properties simultaneously.

\paragraph{First-table occupancy and balance.}
Fix an eligible source set $F$ with $L_0\le|F|\le K$ and a first-query column
$\mathbf x\in[K]^q$. Independence of the entries of $H_1$ gives
\begin{align*}
\Pr\!\left[
\max_{y\in[K]}
|\{\mathbf z\in F:H_1(\mathbf z,\mathbf x)=y\}|
\ge m^2
\right]
&\le
K\binom{|F|}{m^2}K^{-m^2} \\
&\le
K\left(\frac{e}{m^2}\right)^{m^2}.
\end{align*}
Concentration over the $K^q$ independent columns and a union bound over all
eligible source sets show that, simultaneously for every $F$, at most an
$m^{-100d}$ fraction of first-query columns violate the occupancy bound.
This proves part~(i).

We also use the following simultaneous balance event.
Suppose a cut at position $t<q$ leaves a positive linear suffix of the first query
unseen. Let
$\Gamma=\gamma(\mathbf z_0,\mathbf x_1^{1:t})$ have range at most $S_*^d$,
and let $e:[K]^q\to[J]$ be a source-independent label with
$J\le m^{\chi m}$ for some fixed $\chi<\kappa_D$.
Under the original independent uniform law of
$\mathbf z_0\in F$ and $\mathbf x_1\in[K]^q$, the average conditional law of
$H_1(\mathbf z_0,\mathbf x_1)$ given $(\Gamma,e(\mathbf x_1))$ has total
variation distance at most $m^{-\kappa_D/3}$ from $\operatorname{Unif}([K])$.

For fixed $\Gamma$ and $e$, the expected total variation distance over $H_1$ is
at most
$\frac12\sqrt{KS_*^dJ/(|F|K^q)}$.
A bounded-differences estimate is strong enough to union bound over all eligible
source sets, early-state maps, and source-independent labels used below.
The same argument applies when the first table is a random permutation table.
For an incoming first-query length $n\in[n_j,q]$, the complete label at level $j$
will have rate strictly below $N_j$, whereas the shortest incoming query has
entropy $n_j\log_2K=(N_jm+o(m))\log_2m$.
All balance estimates are taken under the original input law, before conditioning
on a successful group or selecting a representative.

\paragraph{The mutual induction.}
Fix a sufficiently large constant $\rho_D$ depending only on $D$.
For every $1\le j\le d$, we prove simultaneously a program-hardness statement
$G_j$ and a prefix-capture statement $\mathcal C_j$.

The statement $G_j$ is as follows.
The initial live key $\mathbf k$ is uniform on $[K]$ and independent of the raw
queries. The computation is a physical $j$-stage replay whose first query has
length $n_j$, followed by $j-1$ fresh $q$-symbol query blocks, and whose target
is the corresponding $j$-fold random-permutation composition.
Its success probability is at most
$\epsilon_j:=m^{-c_j\kappa_D}$ for some constant $c_j>0$.
The same bound holds for every incoming first-query length
$n\in[n_j,q]$, after shifting all later absolute cuts by $n-n_j$.

The companion statement $\mathcal C_j$ makes the capture property precise.
Consider any legal continuation arising in the induction for which the source is
uniform on a set of size at least $K/m^{\rho_D}$ and the current live-key map has
occupancy at most $m^{\rho_D}$. Fix any source-independent old context, but leave
the bounded source-dependent ancestor-state row unconditioned under the source
law. At the prescribed cut, let $\boldsymbol\xi$ be the arrived prefix of the
first unresolved raw query and let $\mathbf V$ denote the corresponding
source-independent old physical word. Then there exists a source-independent
list map $\mathcal P$ such that
$|\mathcal P(\mathbf v)|\le m^{A_jm+o(m)}$ for every realized $\mathbf v$, and
\[
\Pr\!\left[
\mathrm{Succ}\ \wedge\
\boldsymbol\xi\notin\mathcal P(\mathbf V)
\right]
\le
\epsilon_j .
\]
Here $\mathrm{Succ}$ is the branch-masked success event of the continuation.
The source, context, and ancestor-history families that arise in the induction
have cardinality at most $K^{O_D(m)}$, so all events above can be required
simultaneously.

The constants are chosen in the noncircular order
$(G_1,N_1),(\mathcal C_1,A_1),(G_2,N_2),(\mathcal C_2,A_2),\ldots$.
At level $j$, the finite training-tree size and error thresholds are first chosen
from lower-stage rates. We then choose $N_j$ larger than the total rate of the
complete source-independent label needed to prove $G_j$, and afterward choose
$A_j$ larger than the new capture-certificate rate.
Since $D$ is fixed, $\kappa_D$ is chosen last so that every cut, entropy,
occupancy, and moment inequality has positive slack.

\paragraph{Base case.}
The base case $G_1$ is the single-run random-permutation bound.
Fix the initial-state map, transition and emission rules, and output map of one
recurrent run. For a fixed query column, every source-independent physical
encoding produces one of at most $Q_*^n$ physical words.

Let $\theta:=\frac12m^{-\kappa_D/16}$ and
$h:=\lceil\theta K\rceil$.
For a fixed physical word, correctness on $h$ distinct keys forces $h$ prescribed
distinct constraints on a uniform random permutation. Hence
\begin{align*}
\Pr[\textnormal{agreement on at least }h\textnormal{ keys}]
&\le
\binom Kh\frac1{(K)_h}
=
\frac1{h!}.
\end{align*}
Thus one query column is bad for a fixed machine template with probability at
most $Q_*^n/h!$.

The query columns are independent.
For a sufficiently large fixed polynomial $J_{\mathrm{bad}}$ in $m$,
\begin{align*}
\Pr[\textnormal{at least }J_{\mathrm{bad}}\textnormal{ bad columns}]
&\le
\binom{K^n}{J_{\mathrm{bad}}}
\left(\frac{Q_*^n}{h!}\right)^{J_{\mathrm{bad}}}.
\end{align*}
Choosing the degree of $J_{\mathrm{bad}}$ sufficiently large makes the union
bound over all machine templates valid, while
$J_{\mathrm{bad}}/K^n=o(1)$.
This proves $G_1$, uniformly over the permitted incoming lengths and fixed-prefix
slices. The same independent-column argument after separating heavy and diffuse
prefix atoms proves $\mathcal C_1$.

\paragraph{The capture step.}
Assume that all lower statements have been proved.
We use the following certificate estimate, which is the only property of the
analytical training tree needed below.

After applying the lower capture statements from the leaves upward, normalize
every later unresolved raw prefix by its capture-list rank, inverse rank, and the
finite branch information needed to identify the corresponding fixed-prefix
slice. Suppose the remaining unresolved prefix distribution has every atom of
mass at most $1/N$. If the current continuation has success mass $p$, then for
some $R_j=O_D(1)$,
\begin{align*}
p^{R_j}
&\le
\theta_j+\eta_{\mathrm{bad},j}
+\eta_{\mathrm{coll},j}
+\frac{M_j}{N},
\qquad
M_j\le m^{\widetilde A_jm+o(m)}.
\end{align*}
Here $\theta_j$, $\eta_{\mathrm{bad},j}$, and
$\eta_{\mathrm{coll},j}$ are $m^{-\Omega_D(\kappa_D)}$.
The quantity $M_j$ counts the normalized certificates consisting of the lower
capture-list ranks, inverse-context ranks, branch indices, and the finite
training fingerprint.

To see the estimate, sample $R_j$ independent analytical training branches.
Outside the bad and collision events, every joint success determines one of the
$M_j$ normalized certificates. The representative attached to a certificate is
tested only on its own arrived raw prefixes and on legal causal frontier states.
In particular, the validity test is performed before representatives are grouped.
After a certificate is fixed, a diffuse unresolved prefix contributes mass at
most $1/N$. Summing over the $M_j$ certificates gives the displayed bound.
This is also why replacing a representative inside a normalized group never
tests it against a foreign raw-prefix assignment.

We now prove $\mathcal C_j$.
At each source-independent old context, place every arrived-prefix value whose
posterior mass exceeds $m^{-A_jm}$ into the capture list.
There are fewer than $m^{A_jm}$ such values.
Restrict the remaining source measure to the other values without renormalizing.
Every remaining atom then has mass at most $1/N$ for $N=m^{A_jm}$.
Choose $A_j>\widetilde A_j$. The certificate term satisfies
\begin{align*}
\frac{M_j}{N}
&\le
m^{-(A_j-\widetilde A_j)m+o(m)}
=
o(1).
\end{align*}
The certificate estimate shows that the successful mass remaining outside the
capture list is at most $\epsilon_j$ after decreasing $c_j$ if necessary.
This proves $\mathcal C_j$.

\paragraph{Closing the program bound.}
We next prove $G_j$.
Fix an incoming first-query length $n\in[n_j,q]$.
Write the second query as
$\mathbf x_2=(\boldsymbol\xi,\mathbf y_2)$, where
$\boldsymbol\xi\in[K]^{q-n_{j-1}}$ and
$\mathbf y_2\in[K]^{n_{j-1}}$.
Let $c:=n+q-n_{j-1}$ be the first replay cut and put $t:=c-w$.
The parameter choices give $0<t<n$, so a positive linear suffix of
$\mathbf x_1$ remains unseen at the balance cut.

Let
$\mathbf v(\mathbf x_1,\boldsymbol\xi)
:=\mathbf X^{(0)}_{t+1:c}(\mathbf x_1,\boldsymbol\xi)\in[Q_*]^w$.
This old physical word may depend on the complete first query and on the arrived
second-query prefix $\boldsymbol\xi$, but not on the source or on the remaining
raw future.

Apply $\mathcal C_{j-1}$ at this cut, leaving the bounded ancestor-state row
unconditioned under the source law.
For each $\mathbf v\in[Q_*]^w$, let $\mathcal P(\mathbf v)$ be its captured-prefix
list. Put $A_{<j}:=\max_{i<j}A_i$ and
$A^*:=m^{A_{<j}m+o(m)}$.
For each $\boldsymbol\xi$, define
$B_{\boldsymbol\xi}:=\{\mathbf v:\boldsymbol\xi\in\mathcal P(\mathbf v)\}$.
Double counting gives
$\sum_{\boldsymbol\xi}|B_{\boldsymbol\xi}|\le A^*Q_*^w$.
Hence, for every cutoff $L_{\mathrm{ctx}}\ge1$,
$\Pr[|B_{\boldsymbol\xi}|>L_{\mathrm{ctx}}]
\le A^*Q_*^w/(K^{q-n_{j-1}}L_{\mathrm{ctx}})$.

All label fields other than the inverse-context rank have rates already fixed by
lower stages. We choose $L_{\mathrm{ctx}}$ and then $N_j$ so that the exceptional
prefix mass is $m^{-\Omega_D(\kappa_D)}$ and the rank of
$\mathbf v(\mathbf x_1,\boldsymbol\xi)$ inside $B_{\boldsymbol\xi}$, together
with the normalized training fingerprint and all previously fixed alignment
ranks, gives a complete source-independent label
$e=e_{\boldsymbol\xi}(\mathbf x_1)$ satisfying
$|\operatorname{Im}(e)|\le m^{\chi_jm+o(m)}$ for some $\chi_j<N_j$.
A sentinel is used on excluded inputs.
The branch and training indices are fixed on the current slice and suppressed
from the notation.

For each label, choose one full first-query representative from its group.
The representative is analytical lookup data indexed by the label; its raw
coordinates are not another field of the label.
Let
$\Gamma=\gamma(\mathbf z_0,\mathbf x_1^{1:t})$ be the state of the preceding
recurrent run at the balance cut.
For a fixed label value $a$, early row $g$, and live key $\mathbf k\in[K]$,
let $q_{a,g}(\mathbf k)$ be the maximum fresh branch-masked continuation success
of the selected representative over all legal causal choices of the current
frontier state and all later frontier policies.

The normalization procedure retains a successful training sample only when its
own representative has fresh continuation value greater than a fixed
$\delta_0>0$. Samples failing this test are included in
$\eta_{\mathrm{bad},j}$. Thus every retained successful sample with label $a$,
row $g$, and live key $\mathbf k$ satisfies
$q_{a,g}(\mathbf k)>\delta_0$.

For each fixed $(a,g)$ and every $\mathbf k$, choose one deterministic maximizing
legal causal policy. Each chosen frontier state depends only on the raw history
available at its own cut. Therefore these keywise choices paste into one legal
lower-stage physical continuation on the same positions; no choice depends on an
unseen raw suffix.

For a fixed analytical training tree, let
$\operatorname{Cross}^{\mathrm{opt}}
:=\mathbb E[q_{e,\Gamma}(\mathbf K_{\mathrm{live}})]$ under the original source
and query law, and let $H$ be the retained training-success mass.
The pointwise own-source test, the simultaneous balance event, and $G_{j-1}$ give
\begin{align*}
\delta_0 H
&\le
\operatorname{Cross}^{\mathrm{opt}}
\le
\epsilon_{j-1}+\delta_{\mathrm{TV}} .
\end{align*}
Indeed, the first inequality follows pointwise.
For the second, apply the balance estimate once to the complete label $e$ and
early row $\Gamma$. After replacing the conditional live-key law by uniform,
the pasted optimizing policy is one admissible lower-stage physical replay, so
$G_{j-1}$ applies.

Averaging over the analytical training tree and combining this estimate with the
certificate bound gives
\begin{align*}
p^{R_j}
&\le
\frac{\epsilon_{j-1}+\delta_{\mathrm{TV}}}{\delta_0}
+\theta_j+\eta_{\mathrm{bad},j}+\eta_{\mathrm{coll},j}.
\end{align*}
All expectations over the old source and first query retain their original
weights. Excluded inputs are represented by the sentinel or by validity
indicators, and the balance estimate is applied before averaging over
fixed-prefix slices.

The recursive parameter choice leaves a positive exponent after taking the
$R_j$th root, and hence proves $G_j$.
Notice that the argument uses the optimizing causal frontier policy only as an
analytical upper bound; we do not assert that the competing model's actual
future frontier states have the same fresh continuation success.

\paragraph{Returning to the iid first stage.}
Fix an eligible source chunk $F$ with $L_0\le|F|\le K$.
By part~(i), outside an $m^{-100d}$ fraction of first-query columns, the map
$\mathbf z_0\mapsto H_1(\mathbf z_0,\mathbf x_1)$ on $F$ has occupancy at most
$m^2$. Every later permutation preserves this occupancy, and the bad-column mass
is charged under the original first-query law.

Repeat the $G_s$ certificate construction with full incoming first-query length
$q$. Fix the arrived second-query prefix $\boldsymbol\xi$, the inverse-branch
indices, and the analytical training tree.
On this slice, the complete source-independent label has the form
$e_{\boldsymbol\xi}(\mathbf x_1)$ and depends only on the first query.

Write $\mathbf x_2=(\boldsymbol\xi,\mathbf y_2)$ and let
$\mathbf Y:=(\mathbf y_2,\mathbf x_3,\ldots,\mathbf x_s)$ be the remaining
fresh raw future.
The only change from the permutation-first-stage argument is the first balance
step, where we use the simultaneous iid first-table balance event proved above.

Conditional on the first live key $\mathbf z_1$, the complete label, and the
early row, all old variables have a joint conditional law independent of
$\mathbf Y$. This conditional law can be represented using the common auxiliary
seed allowed in Definition~\ref{def:physical-replay}.
For fixed label and early row, the representative's physical token sequence and
replay rules are fixed. Any remaining old-variable dependence can therefore
affect only the permitted frontier states, and each such state depends only on
the raw prefix available at its own cut.
Thus the continuation remains in the physical replay class.

Applying the optimized crossed-certificate argument on every fixed-prefix slice
and then averaging under the original prefix distribution gives
$\Pr[\widehat{\mathbf z}=\mathbf z_s]\le m^{-c_D}+o(1)$.
For $s=1$, the iid-column single-run factorial argument applies directly.

All occupancy, balance, capture, and program events above are simultaneous before
the source chunk and the post-table continuation are selected.
Their intersection therefore has positive probability.
Fixing one lookup-table realization in this intersection proves the lemma.
\end{proof}

Now we are able to give the proof of Theorem~\ref{th insufficient layers}.
Theorem~\ref{th insufficient layers} follows immediately from Lemma~\ref{lm size-2 reduction}, Lemma~\ref{lm size-2 cover} and Lemma~\ref{lm size-2 hardness}.

\begin{proof}[Proof of Theorem~\ref{th insufficient layers}]
    Let $\mathcal{C}_2(\tau)$ be size-2 cover of the target profile $\tau$. By Lemma~\ref{lm size-2 hardness}, for every profile $\sigma \in \mathcal{C}_2(\tau)$, there exists a family of task $\{H_{\sigma,m}\}$, such that there exists a model $M' \in \mathcal{M}_{\sigma}(b_m,b_m,m+1)$ that can solve $H_{\sigma,m}$ exactly, where $b_m = C \log m$, while every hybrid model $M\in\mathcal M_A(B,P,W)$ that satisfies $\mathrm{err}(M,H_{\sigma,m}) \le 1/3$, for some input distribution $\nu_{\sigma,m}$ must satisfy $W \ge \lambda (m+1) $ or $\max\{B,P\} \ge \lambda b_m$, if $A \in \Sigma^{\le D}, \sigma \not\preceq A$.
    By Lemma~\ref{lm size-2 reduction}, by combining the mixture of these tasks, we obtain a single family of tasks
     $\{G_{\tau,m}\}$ such that 
    for every hybrid model $M\in\mathcal M_A(B,P,W)$ that satisfies $\mathrm{err}(M,G_{\tau,m}) \le 1/(4D)$, must satisfy $W \ge \lambda (m+1) $ or $\max\{B,P\} \ge \lambda C \log m$, if $A \in \Sigma^{\le D}$ and $\exists \sigma \in \mathcal{C}_2(\tau), \sigma \not\preceq A$.
    On the other hand, there exists a model $M' \in \mathcal{M}_\tau(C\log m,C\log m,,m+1)$ that can solve $G_{\tau,m}$ exactly. 
    Finally, by Lemma~\ref{lm size-2 cover}, if $\tau \not\preceq A$, then there must be some $\sigma \in \mathcal{C}_2(\tau)$ such that $\sigma \not \preceq A$, which completes the proof of Theorem~\ref{th insufficient layers}.
\end{proof}

\subsection{Proof of Theorem~\ref{th equal-depth-general-profile}}

Given Theorem~\ref{th wrong order} and Theorem~\ref{th insufficient layers}, the proof of Theorem~\ref{th equal-depth-general-profile} is almost immediate.

\begin{proof}[Proof of Theorem~\ref{th equal-depth-general-profile}]
Let $\alpha := \mathrm{comp}(\tau)$. 
Let $\{F^{\mathrm{ord}}_{\tau,\epsilon,m}\}_{m\geq 1}$
be the task family constructed in Theorem~\ref{th wrong order} for the wrong-order case, with
input distribution $\mu^{\mathrm{ord}}_m$, and let $\{F^{\mathrm{ins}}_{\tau,m}\}_{m\geq 1}$
be the task family constructed in Theorem~\ref{th insufficient layers} for the insufficient-profile
case, with input distribution $\mu^{\mathrm{ins}}_m$. That common constant factor can be selected so that Theorem~\ref{th wrong order} and Theorem~\ref{th insufficient layers} still hold.

We combine the two task families into a single task family
$\{F_{\tau,\epsilon,m}\}_{m\geq 1}$.
Sample $J$ uniformly from $\{0,1\}$.  Conditioned on $J=0$, sample an
input from $\mu^{\mathrm{ord}}_m$ and ask the model to compute
$F^{\mathrm{ord}}_{\tau,\epsilon,m}$; conditioned on $J=1$, sample an
input from $\mu^{\mathrm{ins}}_m$ and ask the model to compute
$F^{\mathrm{ins}}_{\tau,m}$.  We include $J$ as an additional tag in
every input token and pad the two types of inputs to the same sequence
length.  Denote the resulting input distribution by $\mu_m$. 

By the completeness constructions of Theorems~\ref{th wrong order} and~\ref{th insufficient layers}, both component
tasks can be solved exactly by models with profile $\tau$.  Using the
tag $J$ to select between the corresponding layer maps, as in the
selector construction in Lemma~10, we obtain a single model $M_{\tau,m}
    \in
    \mathcal M_{\tau}
    (B_{\tau,m},P_{\tau,m},W_{\tau,m})$.
that solves $F_{\tau,\epsilon,m}$ exactly.  After increasing the
constant $C_{\tau,\epsilon}$ if necessary, $ B_{\tau,m},P_{\tau,m}
    \leq C_{\tau,\epsilon}\log m,
    W_{\tau,m}\leq m+1.$
Moreover, taking the union of the two vocabularies and padding the
inputs preserves $|\mathcal V_{\epsilon,m}|
    \leq \mathrm{poly}_{\epsilon}(m), 
    L_{\epsilon,m}\leq \mathrm{poly}_{\epsilon}(m).$

It remains to prove soundness.  Fix an architecture profile
$A\in\Sigma^{\leq D}$ such that $\tau\not\preceq A$, and let $M\in\mathcal M_A(B,P,W)$
satisfy $ \mathrm{err}_{\mu_m}(M,F_{\tau,\epsilon,m})
    \leq \frac{1}{8D}.$
Let $e_{\mathrm{ord}}$ and $e_{\mathrm{ins}}$ denote the conditional
errors of $M$ conditioned on $J=0$ and $J=1$, respectively.  Since $J$
is uniform,
\begin{align*}
    \mathrm{err}_{\mu_m}(M,F_{\tau,\epsilon,m})
    =
    \frac{1}{2}
    \left(e_{\mathrm{ord}}+e_{\mathrm{ins}}\right).
\end{align*}
    
Therefore, $ e_{\mathrm{ord}},e_{\mathrm{ins}}
    \leq
    2\mathrm{err}_{\mu_m}(M,F_{\tau,\epsilon,m})
    \leq \frac{1}{4D}.$

We now distinguish two cases.

\paragraph{Case 1: $\alpha\not\preceq A$.}
Since $ e_{\mathrm{ord}}\leq \frac{1}{4D}\leq \frac13,$
Theorem~\ref{th wrong order} applies to the wrong-order component and gives
\begin{align*}
    W\geq c_{\tau}m^{1/\epsilon}
    \qquad\text{or}\qquad
    \max\{B,P\}\geq c_{\tau}m.
\end{align*}

\paragraph{Case 2: $\alpha\preceq A$.}
Since $\tau\not\preceq A$ by assumption, we have $\alpha\preceq A
    \qquad\text{and}\qquad
    \tau\not\preceq A.$
Thus $A$ is in the insufficient-profile case.  Since $e_{\mathrm{ins}}\leq \frac{1}{4D},$
Theorem~\ref{th insufficient layers} applies to the insufficient-profile component and gives
\begin{align*}
    W\geq \lambda_{\tau}(m+1)
    \qquad\text{or}\qquad
    \max\{B,P\}\geq
    \lambda_{\tau} C_{\tau}\log m.
\end{align*}
The two cases are exhaustive, since either
$\alpha\not\preceq A$ or $\alpha\preceq A$.  This proves both
soundness statements and completes the proof.
\end{proof}

\section{Omitted Details in Section~\ref{sec:natural_tasks}}\label{app natural task}

We provide proofs to various theorems in the main body. We first prove the core hardness result for tasks with a dominant \m{TS} architecture. We then look specifically at the two tasks, segmented retrieval and exact copy, showing that these tasks are indeed in this family of tasks. Lastly, we provide constructions for these two tasks that use less window and state-size than what is provably required for \m{ST} architectures.


\subsection{Hardness Results}

We begin with the core assumption for all tasks with a \m{TS} dominant architecture before quickly following with the statement and proof of the $\SSM\to\TF$ theorem. After establishing this result, we show the specific bounds for exact copy and segmented retrieval, where the majority of the proofs come from defining the hard data distribution and cut position $\tau$.

\begin{assumption}\label{asp:U_sensitive}
     Let $F$ be a sequence map from $\gV^L$ to $\gY$. We assume $F$ satisfies the following property:
     there exists an index $\tau$, a set of prefixes $\gU \subseteq \gV^\tau \times \{0,1\}^\tau$, and a set of suffixes $\gQ \subseteq \gV^{L-\tau} \times \{0,1\}^{L-\tau}$ such that the map $\Gamma: \gU \to \gY^{|\gQ|}$ defined as
     $$\Gamma(u) := (F(u,q))_{q \in \gQ} \in \gY^{|\gQ|},$$
     is an injection. We call $\tau$ the \textit{cut}.
\end{assumption}


\begin{theorem}[$\SSM\to\TF$ Cut, Restated]
Consider any function $F$ that satisfies 
Assumption \ref{asp:U_sensitive}. Let $M$ be a model of architecture \m{ST} modeling $F$ with state $\gS$ and window $W$. Suppose that the cut position is bounded as $\tau \leq L - W$.
If $M$ computes $f:\mathcal{V}^L\rightarrow \mathcal{Y}$ correctly on all the inputs, 
then $|\mathcal{S}|\ge |\gU|$.
\end{theorem}

\begin{proof}
Let $S_t$ denote the state at position $t$. Let $u,u' \in \gU$ be prefixes such that $S_\tau(u) = S_\tau(u')$. Since the SSM is deterministic, at time $t \geq \tau$, the SSM states remain identical, i.e.
\begin{align*}
    S_t(u,q)=S_t(u',q),\forall t>\tau,
\end{align*}
and the SSM layer outputs are the same. 
\\
Since the $\TF$ layer can only access the tokens from position $L-W$ to $L$, and $\tau \leq L-W$, we have that the model outputs are also the same, i.e. 
\begin{align*}
    M(u,q)=M(u',q).
\end{align*}
Now assume $M$ computes $f$ correctly on all inputs and the map $\Gamma$ of a set of prefixes $\gU$ is injective (Assumption \ref{asp:U_sensitive}). If two prefixes $u\neq u'$ outputs the same state at the cut position $\tau$, then by the theorem we have 
\begin{align*}
    M(u,q)=f(u,q)=f(u',q)=M(u',q),\quad \forall q\in \mathcal{Q},
\end{align*}
which immediately shows that $\Gamma(u)=\Gamma(u').$  
Since the map $\Gamma$ is injective, we have $u=u'$, which contradicts the assumption that $u\neq u'$. Therefore, the map 
\begin{align*}
    u\mapsto S_t(u), \quad \forall u\in \mathcal{U},
\end{align*}
must be injective, which implies that $|\mathcal{S}|\ge |\mathcal{U}|$.
\end{proof}


\begin{theorem}[Exact Copy Lower Bound]
\label{thm:exact_copy_lb}
Set $|\mathcal{V}|\ge5,L\ge cl,$ and define $h:=\min(l,(L-W)_{+})$. 
If a deterministic hybrid model \m{ST} performs the Exact Copy task correctly on all inputs, then $\log |\mathcal{S}| \ge \Omega(h \log |\gV|)$. 
\end{theorem}

\begin{proof}
We use the notation of juxtaposition to mean sequence concatenation. For example, for any two finite sequences
\begin{align*}
    u=(u_1,\ldots,u_m)\in\mathcal{V}^m
\qquad\text{and}\qquad
v=(v_1,\ldots,v_n)\in\mathcal{V}^n,
\end{align*}
we write 
\begin{align*}
   uv:=(u_1,\ldots,u_m,v_1,\ldots,v_n).
\end{align*}
The result follows immediately if $h=0$. Thus, we will assume $h \ge 1$.

We first pick two distinct elements $g_1,g_2 \in \mathcal{V}$, and set $\mathcal{A}:=\mathcal{V}\setminus\{g_1,g_2\}$. Since $|\mathcal{V}|\ge 5$, we have $|\mathcal{A}|\ge 3$, thus we may pick another element $\alpha\in \mathcal{A}$. For each $a_{1:h}=(a_1,\cdots,a_h)\in \mathcal{A}^h$, we define the prefix map $u:\mathcal{A}^{h}\rightarrow \mathcal{A}^l$ by 
\begin{align*}
    u(a_{1:h})=a_{1:h}\alpha^{l-h}=(a_1,\cdots,a_h, \underbrace{\alpha,\cdots,\alpha}_{l-h})\in \mathcal{A}^l.
\end{align*}
We also define
\begin{align*}
\mathbf{g}_{1:L-cl} :=g_1^lg_2^lg_1^lg_2^l\cdots, 
\end{align*}
where the alternating concatenation is truncated after its first $L-cl$ entries.
We refer to $\mathbf{g}_{1:L-cl}$ as the \emph{failure segment}.
By construction, every symbol in $\mathbf{g}_{1:L-cl}$ differs from
the symbol $l$ positions later whenever both are defined.

For each $a_{1:h}\in \mathcal{A}^h$, we define the suffix map $q:\mathcal{A}^h\rightarrow\mathcal{\mathcal{V}}^{L-h}$ by
\begin{align*}
    q(a_{1:h})&:=\alpha^{l-h}u(a_{1:h})^{c-1}\mathbf{g}_{1:L-cl} \\
    &=\alpha^{l-h}(a_1\cdots a_h \alpha^{l-h})^{c-1}g_1^{l}g_2^lg_1^lg_2^l\cdots.
\end{align*}
We define $f_{\text{EC}}:\mathcal{V}^L\rightarrow\{0,1\}$ to be the function 
\begin{align*}
    f_{\text{EC}}(\mathbf{x}_{1:L}):=\mathbf{1}_{\{\mathbf{x}_{1:L} \text{ has an exact copy of repetition $c$ and string length $l$}\}}.
\end{align*}
For some fixed $a_{1:h}\in \mathcal{A}^h$, then it is clear that the sequence
\begin{align*}a_{1:h}q(a_{1:h})&=a_1\cdots a_h\alpha^{l-h}(a_1\cdots a_h \alpha^{l-h})^{c-1}g_1^{l}g_2^lg_1^lg_2^l\cdots \\
    &=(a_1\cdots a_h\alpha^{l-h})^cg_1^{l}g_2^lg_1^lg_2^l\cdots
\end{align*}
contains an exact copy, i.e. $f_{\text{EC}}(a_{1:h}q(a_{1:h}))=1$.

For any $a'_{1:h}\in \mathcal{A}^h$ such that $a'_{1:h}\neq a_{1:h}$, then 
\begin{align*}
    a'_{1:h}q(a_{1:h})=a'_1\cdots a'_h\alpha^{l-h}\left(a_1\cdots a_h \alpha^{l-h}\right)^{c-1}g_1^lg_2^lg_1^lg_2^l\cdots.
\end{align*}
For $\mathbf{x}\in\mathcal{V}^L$, define
\begin{align*}
    E(\mathbf{x})
    :=
    \left\{
        m\in[1,L-l]:
        \mathbf{x}_m=\mathbf{x}_{m+l}
    \right\}.
\end{align*}
By the definition of Exact Copy,
\begin{align}
    f_{\mathrm{EC}}(\mathbf{x})=1
    \quad\Longleftrightarrow\quad
    E(\mathbf{x})
    \text{ contains $(c-1)l$ consecutive integers}.
    \label{eq:exact_copy_E_characterization}
\end{align}
We now show that
$    |E(a'_{1:h}q(a_{1:h}))|<(c-1)l,$
which implies this sequence contains no exact copy.
\\Define $\mathbf{x}':=a'_{1:h}q(a_{1:h})$, we now determine $E(\mathbf{x}')$. 
For $1\le m\le h$, we have
$    \mathbf{x}'_m=a'_m,
    \mathbf{x}'_{m+l}=a_m,$
and hence
$ m\in E(\mathbf{x}')
    \Longleftrightarrow
    a'_m=a_m.$
For $h<m\le l$, both $\mathbf{x}'_m$ and $\mathbf{x}'_{m+l}$ are equal to
$\alpha$. Moreover, for
$l<m\le (c-1)l,$
the positions $m$ and $m+l$ correspond to the same position in two
consecutive copies of
$ a_1\cdots a_h\alpha^{l-h},$
and therefore $\mathbf{x}'_m=\mathbf{x}'_{m+l}$.

On the other hand, for every remaining $m\in[1,L-l]$ with
$m>(c-1)l$, we have
$    \mathbf{x}'_m\neq \mathbf{x}'_{m+l}.$
Indeed, if $m$ lies in the last copy of
$a_1\cdots a_h\alpha^{l-h}$, then $\mathbf{x}'_m\in\mathcal{A}$ while
$\mathbf{x}'_{m+l}\in\{g_1,g_2\}$; if $m$ lies in the failure segment
$\mathbf{g}_{1:L-cl}$, then $\mathbf{x}'_m\neq\mathbf{x}'_{m+l}$ by the construction
of $\mathbf{g}_{1:L-cl}$.

Therefore,
\begin{align*}
    E(\mathbf{x}')
    =
    \left\{
        m\in[1,h]:a'_m=a_m
    \right\}
    \cup
    [h+1,(c-1)l].
\end{align*}
In particular,
$E(\mathbf{x}')
    \subseteq
    [1,(c-1)l].$

Since $a'_{1:h}\neq a_{1:h}$, there exists some $j\in[1,h]$ such that
$a'_j\neq a_j$. Hence
$    j\notin E(\mathbf{x}').$
Thus $E(\mathbf{x}')$ does not contain all $(c-1)l$ integers in
$[1,(c-1)l]$. Since $E(\mathbf{x}')\subseteq[1,(c-1)l]$, it cannot contain
$(c-1)l$ consecutive integers, i.e. $|E(a'_{1:h}q(a_{1:h}))|<(c-1)l$. By
\eqref{eq:exact_copy_E_characterization},
$f_{\mathrm{EC}}\left(a'_{1:h}q(a_{1:h})\right)=0.$

Therefore, we have that 
\begin{align*}
    f_{\text{EC}}\left(a'_{1:h}q\left(a_{1:h}\right)\right)
=
\begin{cases}
1,&a'_{1:h}=a_{1:h},\\
0,&a'_{1:h}\neq a_{1:h}.
\end{cases}
\end{align*}
Define $f:=f_{\text{EC}},\mathcal{U}:=\mathcal{A}^h,\mathcal{Q}:=\{q( a_{1:h}):a_{1:h}\in \mathcal{A}^h \},\tau:=h$ following the setting of Theorem \ref{thm:hybrid_cut}. Notice that the map $a_{1:h}\mapsto q(a_{1:h})$ is injective, since
$q(a_{1:h})$ contains $u(a_{1:h})$; and $\Gamma$ is also injective on $\gU$ because $\Gamma$ is a one-hot vector at the position in the sequence for each input:
\begin{align*}
    a'_{1:h} \mapsto \Gamma(a'_{1:h})=\left(f\left(a'_{1:h}q\left(a_{1:h}\right)\right)\right)_{a_{1:h}\in \mathcal{A}^h}=(0,\cdots,0,1,0,\cdots0),
\end{align*}
where the slot with value $1$ corresponds to the position $a_{1:h}=a'_{1:h}$

By Theorem \ref{thm:hybrid_cut}, if $M$ computes $f$ correctly on all inputs, then 
\begin{align*}
    |\mathcal{S}| \ge |\mathcal{U}| = |\mathcal{A}|^h,
\end{align*}
Therefore,
\begin{equation*}
    \log |\gS| \ge \Omega(h \log |\gV|).
\end{equation*}
\end{proof}

For the corollary, if $L\ge cl$ and $W\le L-l$, then $L-W\ge l$, Thus $h=\min(l,(L-W)_+)=l$ and
\begin{align*}
    \log |\gS| \ge \Omega(l \log |\gV|).
\end{align*}


\begin{theorem}[Segmented Retrieval Lower Bound]
\label{thm:seg_ret_lb}
Set $|\mathcal{N}| \ge 2$ and $L\ge 3\lambda+1$, and $h:=\min(\lambda,(L-W)_+)$. Then any deterministic hybrid model $M$ of architecture \m{ST} that performs the Segmented Retrieval task correctly on all inputs must satisfy $\log |\mathcal{S}|\ge \Omega(h \log|\mathcal{N}|)$.
\end{theorem}

\begin{proof}
For $h=0$ the bound is vacuous, so we assume $h\geq1$. \\
Define the prefix map $u:\mathcal{N}^h\rightarrow (\mathcal{N}\times \{0,1\})^h$ by 
\begin{align*}
    u(n_{1:h})=\left(\left(n_1,0\right),\left(n_2,0\right),\cdots,\left(n_h,0\right)\right), \quad \forall n_{1:h}\in \mathcal{N}^h.
\end{align*}
Since $h\le L-W$, $h$ can be a candidate for the cut position $\tau$. Therefore, adversarially, we set $\tau=h$. We fix a position $j \in [h]$ that serves as the query position for the segmented retrieval; that is to say, for each $v_{1:L-h}\in \mathcal{V}^{L-h}$, we set the suffix map $q_j:\mathcal{V}^{L-h}\rightarrow (\mathcal{V} \times \{0,1\})^{L-h}$ as 
\begin{align*}
    q_j(v_{1:L-h}):=\left( \left(v_1,0\right), \left(v_2,0\right),\cdots \left(v_{j+\lambda-h},1\right),\cdots \left(v_{L-h},0\right) \right).
\end{align*}
Therefore, the Online Segmented Retrieval task for $\mathbf{x}_{1:L}:=u(n_{1:h})q_j(v_{1:L-h})$ is to output $n_j$.
\\
Pick $n'_{1:h},n_{1:h}\in \mathcal{N}^h$ such that $n'_{1:h}\neq n_{1:h}$, and let $j'\in [h]$ be the index such that $n'_{j'}\neq n_{j'}$. Fix $v_{1:L-h}\in \mathcal{V}^{L-h}$. We construct two input sequences $\mathbf{x'}_{1:L}:=u(n'_{1:h})q_{j'}(v_{1:L-h})$ and $\mathbf{x}_{1:L}:=u(n_{1:h})q_{j'}(v_{1:L-h})$.  
Assign an arbitrary deterministic hybrid model $M$ of architecture \m{ST} to the task of Segmented Retrieval. Then
\begin{align*}
    M(\mathbf{x'}_{1:L})=n'_{j'}, \quad M(\mathbf{x}_{1:L})=n_{j'}.
\end{align*}
Therefore, we must have $S_h\left(u\left(n'_{1:h}\right)\right)\neq S_h\left(u\left(n_{1:h}\right)\right)$. Otherwise, by Theorem \ref{thm:hybrid_cut}, we have $M(\mathbf{x'}_{1:L})=n'_{j'}= M(\mathbf{x}_{1:L})=n_{j'},$ which contradicts $n'_{j'}\neq n_{j'}$. Hence, we conclude the map
\begin{align*}
    n_{1:h} \mapsto S_h\left(u\left(n_{1:h}\right)\right), \quad \forall n_{1:h}\in \mathcal{N}^h,
\end{align*}
is injective, and thus implies 
\begin{align*}
    |S|\ge |\mathcal{N}|^h.
\end{align*}
\end{proof}
For the corollary, if $L\ge3\lambda+1$ and $W\le \lambda+1$, then $L-W\ge 2\lambda$, hence $h:=\min(\lambda,(L-W)_+)=\lambda$. Thus,
\begin{align*}
    \log |S|\ge \Omega(\lambda \log|\mathcal{N}|).
\end{align*}

\subsection{Constructions}
\label{app:constructions}

\subsection*{Construction for Exact Copy}
\begin{figure}[t]
    \centering
    \includegraphics[width=0.8\linewidth]{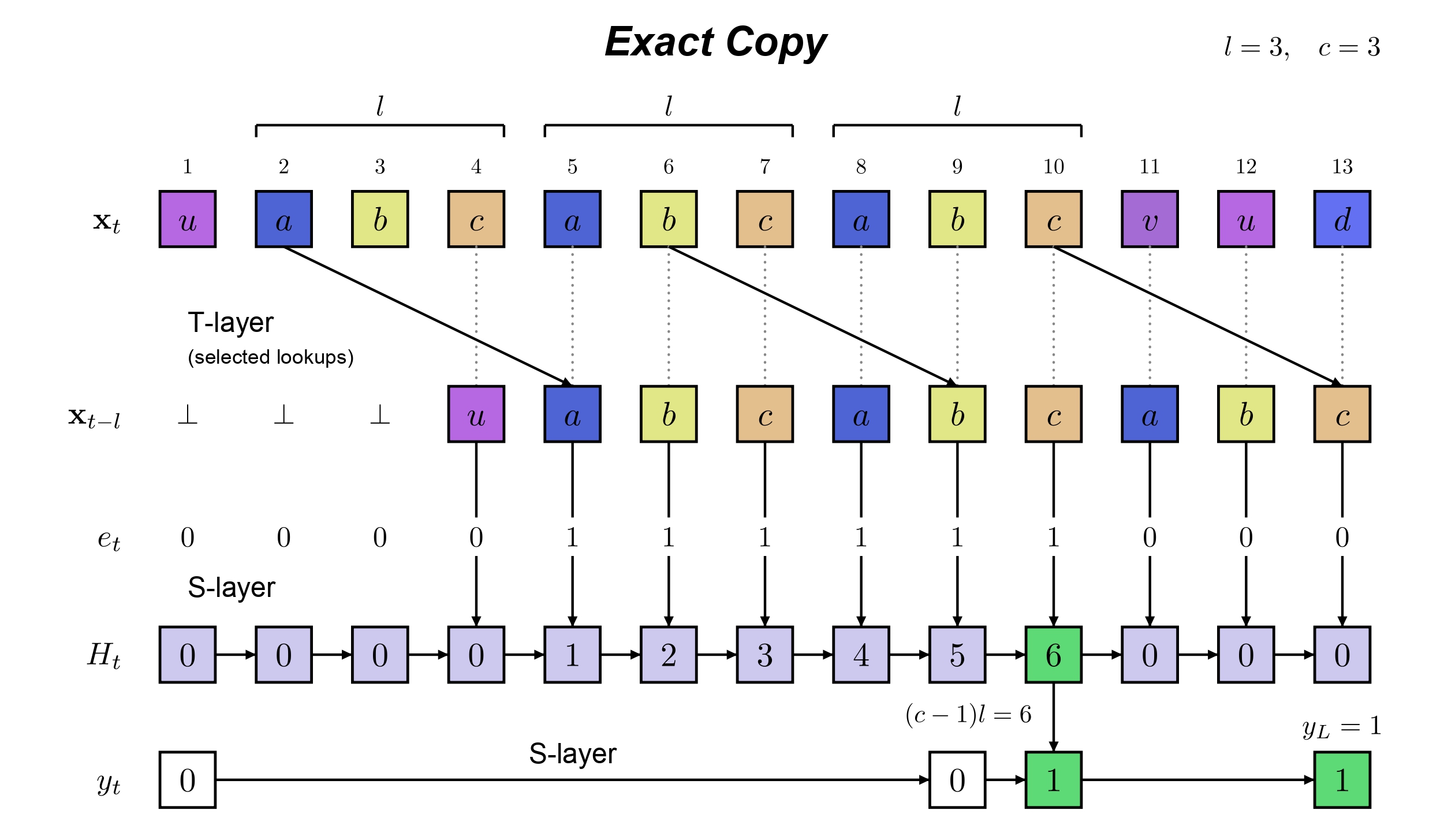}
    \caption{Structure of Exact Copy.
    At each position $t>l$, the $\TF$ layer retrieves $\mathbf{x}_{t-l}$.
    The first $\SSM$ layer counts consecutive matches
    $\mathbf{x}_t=\mathbf{x}_{t-l}$ in $H_t$, while the second
    sets $y_t=1$ once $H_t=(c-1)l$ and retains this value thereafter.
    Here, $abc$ repeats three times consecutively, giving
    $H_{10}=6$ and $y_L=1$ despite subsequent mismatches.}
    \label{fig:exact-copy}
\end{figure}
We first outline the high-level idea behind the construction. At each position $t$, we want to examine whether this position can be an ending position of the desired exact copy. This task comes down to finding the number of equalities up to each position $t$. Let the number of consecutive equalities up to each position $t$ be denoted by $H_t$. The SSM layer is set to calculate $H_t=\mathbf{1}_{\{t\ge l+1,\mathbf{x}_t=\mathbf{x}_{t-l}\}}(H_{t-1}+1)$. In other words, if the current position satisfies $\mathbf{x}_t=\mathbf{x}_{t-l}$, then we add 1 to the counter so $H_t=H_{t-1}+1$; otherwise, we reset the counter. The Transformer layer first extracts $\mathbf{x}_{t-l}$ at each position $t$. Lastly, we need another SSM layer to check if $H_{t}=(c-1)l$, that is, we need SSM to output $y_t=\mathbf{1}_{\{H_{t}=(c-1)l\}}\lor \mathbf{1}_{\{y_{t-1}=1\}}$. 
Since for any two variables $x,y\in \{0,1\}$ we have that $x\lor y=(1-y)x+y$, it follows that
\begin{align*}
    y_t=(1-y_{t-1})\mathbf{1}_{\{H_{t}=(c-1)l\}}+y_{t-1}=(1-\mathbf{1}_{\{H_{t}=(c-1)l\}})y_{t-1}+\mathbf{1}_{\{H_{t}=(c-1)l\}}.
\end{align*}
\\
\\
We now present the detailed hybrid model construction.

\begin{theorem}[Exact Copy Construction]
There exists a model $M$ of architecture \m{TSS} with window $W\leq O(l)$ and state $\log |\gS| \leq O(\log |\gV| + \log (cl))$ that performs Exact Copy (Definition \ref{dfn:exact_copy}).
\end{theorem}

\begin{proof}
We provide the various components of the \m{TS} architecture in distinct sections.

\textbf{Token Embeddings.} We define $\psi':\mathcal{M}\rightarrow \{\pm1 \}^{\log|\mathcal{M}|}$ to be the binary encoding of the vocabulary in $\mathcal{M}$.
\\
\\ \textbf{Positional Encoding.} We define $\phi':[L]\rightarrow \{ \pm1\}^{\log|L|}$ as the binary encoding of numbers in $[L]$. After adding the positional encoding, the input context has the following form:
\begin{align*}
\text{TF-input}(\vec{\mathbf{x}}) := \begin{pmatrix}
\psi'(\mathbf{x}_1) & \cdots & \psi'(\mathbf{x}_{L-1}) & \psi'(\mathbf{x}_L) \\
\mathbf{0} & \cdots & \mathbf{0} & \mathbf{0} \\
\phi'(1) & \cdots & \phi'(L-1) & \phi'(L) \\
\mathbf{0} & \cdots & \mathbf{0} & \mathbf{0} \\
\mathbf{1} & \cdots & \mathbf{1} & \mathbf{1} \\
\end{pmatrix}
\end{align*}
\\
\\
\textbf{Transformer Layer.} As outlined in the high-level idea above, we need to extract the token $\mathbf{x}_{t-l}$ at each position $t$. The required sliding window size is $l+1$. We add a positional bias $\mathbf{B}_{t-l}$ into the attention:
\begin{align*}
    Attn_j(TF-input(\vec{\mathbf{x}})):=\sum_{i=\max(1,j-l)}^{j}\frac{\exp((\mathbf{B}_{j-l})_{i})}{\sum_{i=\max(1,j-l)}^{j}\exp((\mathbf{B}_{j-l})_i)} \psi'(\mathbf{x}_{i})=\psi'(\mathbf{x}_{j-l}),
\end{align*}
where $(\mathbf{B}_{j-l})_i=0,i=j-l,(\mathbf{B}_{j-l})=-\infty,\forall i\neq j-l$. That is, we set $\mathbf{W}_q=\mathbf{W}_k=0$ and $\mathbf{W}_v(TF-input(\vec{\mathbf{x}}))=(\psi'(\mathbf{x}_1),\cdots,\psi'(\mathbf{x}_{L}))^{\top}$.
Therefore, the output of the transformer layer is as follows:
\begin{align*}
\begin{pmatrix}
\psi'(\mathbf{x}_1) & \cdots & \psi'(\mathbf{x}_{L-1}) & \psi'(\mathbf{x}_L) \\
\mathbf{0} & \cdots & \psi'(\mathbf{x}_{L-1-l}) & \psi'(\mathbf{x}_{L-l}) \\
\phi'(1) & \cdots & \phi'(L-1) & \phi'(L) \\
\mathbf{0} & \cdots & \mathbf{0} & \mathbf{0} \\
\mathbf{1} & \cdots & \mathbf{1} & \mathbf{1} \\
\end{pmatrix}
\end{align*}
Here we set that $\forall j\le l,  Attn_j(TF-input(\vec{\mathbf{x}}))=\mathbf{0}$.
\\
\\
\textbf{SSM Layers.} As outlined in the high-level idea of the SSM layer construction, we need two SSM layers to solve the task. Moreover, the first SSM layer is a two-gated SSM layer. The update rule becomes:
\begin{align*}
    H_t=(I-\Delta_{A}(\mathbf{x}_t)W_A)H_{t-1}+\Delta_{B}(\mathbf{x_t})W_B\mathbf{x}_t,
\end{align*}
where $\Delta_B(\mathbf{x}_t)$ is a scalar depending on $\mathbf{x}_t$ but different from $\Delta_A(\mathbf{x}_t)$.

\noindent We denote the number of consecutive equalities up to position $t$ as $H_t$. Rigorously, $H_t = \max \left( \{0\} \cup \{h \in [t] : e_{t-h+1} = \cdots = e_t = 1\} \right)$, where $e_t:=\mathbf{1}_{\{t\ge l+1,\psi'(\mathbf{x}_{t})=\psi'(\mathbf{x}_{t-l})\}}$.
\\For the first SSM layer, our goal is $H_t=\mathbf{1}_{\{t\ge l+1,\psi'(\mathbf{x}_{t})=\psi'(\mathbf{x}_{t-l})\}}(H_{t-1}+1)$, whence we assign
\begin{align*}
\Delta_A(
\begin{pmatrix}
\psi'(\mathbf{x}_t) \\
\psi'(\mathbf{x}_{t-l}) \\
\phi'(t) \\
\mathbf{0} \\
\mathbf{1}
\end{pmatrix})=1-\mathbf{1}_{\{t\ge l+1,\psi'(\mathbf{x}_{t})=\psi'(\mathbf{x}_{t-l})\}}, \quad \Delta_B(
\begin{pmatrix}
\psi'(\mathbf{x}_t) \\
\psi'(\mathbf{x}_{t-l}) \\
\phi'(t) \\
\mathbf{0} \\
\mathbf{1}
\end{pmatrix})=\mathbf{1}_{\{t\ge l+1,\psi'(\mathbf{x}_{t})=\psi'(\mathbf{x}_{t-l})\}},
\end{align*}
\begin{align*}
\mathbf{W}_A=\mathbf{W}_C=I, \quad \mathbf{W}_B(
\begin{pmatrix}
\psi'(\mathbf{x}_t) \\
\psi'(\mathbf{x}_{t-l}) \\
\phi'(t) \\
\mathbf{0} \\
\mathbf{1}
\end{pmatrix})=1.
\end{align*}
Therefore, the output of the first SSM layer is as follows.
\begin{align*}
\begin{pmatrix}
\psi'(\mathbf{x}_1) & \cdots & \psi'(\mathbf{x}_{L-1}) & \psi'(\mathbf{x}_L) \\
\mathbf{0} & \cdots & \psi'(\mathbf{x}_{L-1-l}) & \psi'(\mathbf{x}_{L-l}) \\
\phi'(1) & \cdots & \phi'(L-1) & \phi'(L) \\
H_1 & \cdots & H_{L-1} & H_L \\
\mathbf{1} & \cdots & \mathbf{1} & \mathbf{1} 
\end{pmatrix}
\end{align*}
The second SSM layer uses the standard update rule, i.e. $y_t=(I-\Delta(\mathbf{x}_t)W_A)y_{t-1}+\Delta(\mathbf{x}_t)W_B \mathbf{x}_t$. As outlined in the high-level idea, the second layer is designed to compute
\begin{align*}
    y_t=(1-\mathbf{1}_{\{H_{t}=(c-1)l\}})y_{t-1}+\mathbf{1}_{\{H_{t}=(c-1)l\}}.
\end{align*}
This final expression is attained with the following weights for the second SSM:
\begin{align*}
    \Delta(
    \begin{pmatrix}
\psi'(\mathbf{x}_t) \\
\psi'(\mathbf{x}_{t-l}) \\
\phi'(t) \\
H_t \\
\mathbf{1}
\end{pmatrix}
    )=\mathbf{1}_{\{H_{t}=(c-1)l\}},
\end{align*}
\begin{align*}
\mathbf{W}_A=\mathbf{W}_C=I, \quad \mathbf{W}_B(
\begin{pmatrix}
\psi'(\mathbf{x}_t) \\
\psi'(\mathbf{x}_{t-l}) \\
\phi'(t) \\
H_t \\
\mathbf{1}
\end{pmatrix})=1.
\end{align*}

\end{proof}


\subsection*{Construction for Segmented Retrieval}

\begin{figure}
    \centering
    \includegraphics[width=0.8\linewidth]{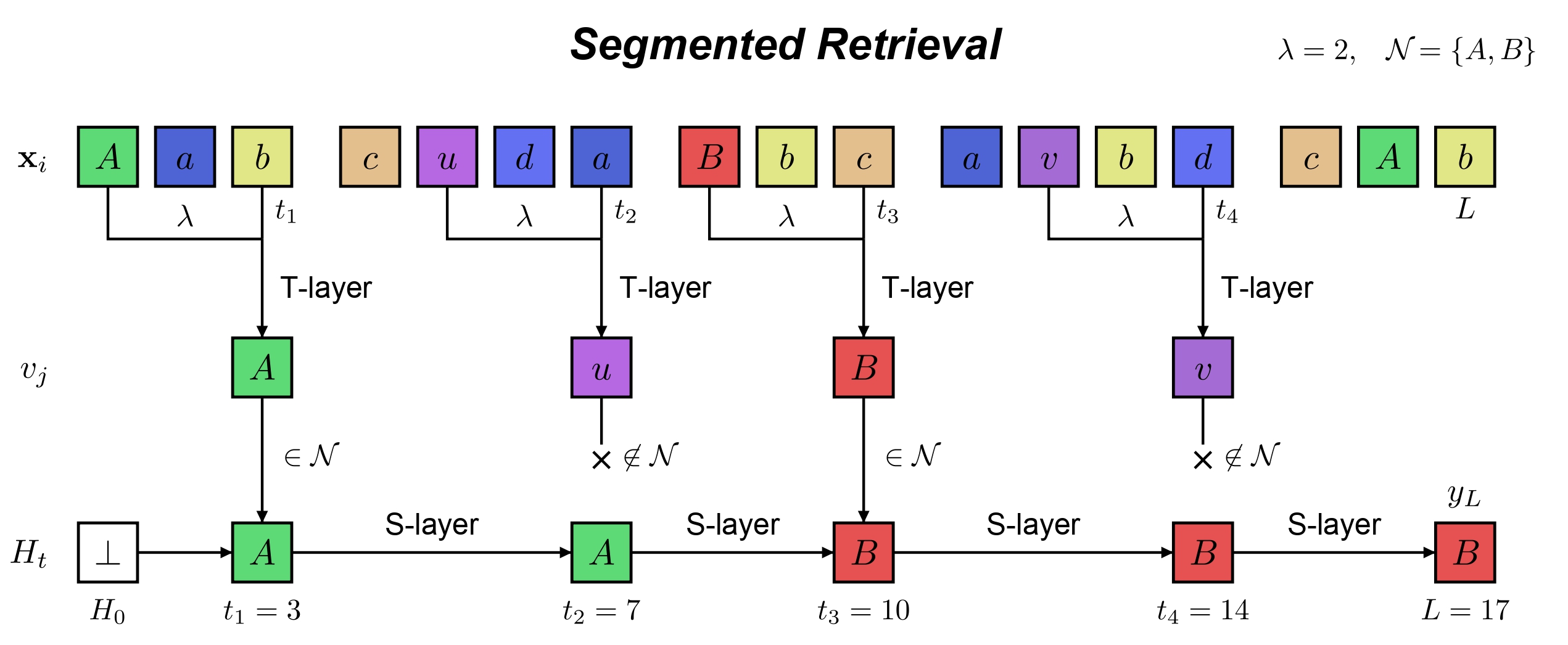}
    \caption{Structure of Segmented Retrieval. At each internal boundary $t_j$, the $\TF$ layer retrieves
    $v_j=\mathbf{x}_{t_j-\lambda}$.
    The $\SSM$ layer retains the most recent retrieved token belonging
    to $\mathcal{N}$ and outputs it at position $L$.
    Here, the retrieved tokens are $A,u,B,v$, so the output is $B$.}
    \label{fig:fig1}
\end{figure}

We first outline the high-level idea behind the construction. 
\\To solve the Segmented Retrieval task, we can first use a Transformer layer to extract the token $\mathbf{x}_{t-\lambda}$ at each position $t$ by adding a positional bias $\mathbf{B}_{t-\lambda}$. Then, at each position $t$, we need to check if $t\in \mathbf{T}:=\{ t_1,\cdots ,t_{k-1} \}$ and $\mathbf{x}_{t-\lambda} \in \mathcal{N}$ are both true. If both conditions are satisfied, then we set the SSM state as $H_t=\mathbf{x}_{t-\lambda}$; otherwise, $H_t=H_{t-1}$, maintaining the previous state. This can be achieved through a standard SSM layer, as we set 
\begin{align*}
    H_t=(I-\mathbf{1}_{\{(t\in \mathbf{T})\land (\mathbf{x}_{t-\lambda}\in \mathcal{N})\}})H_{t-1}+\mathbf{1}_{\{(t\in \mathbf{T})\land (\mathbf{x}_{t-\lambda}\in \mathcal{N})\}}\mathbf{x}_{t-\lambda}.
\end{align*}
\\

\begin{theorem}[Segmented Retrieval Construction]
There exists a model $M$ of architecture \m{TS} with window $W\leq O(\lambda)$ and state $\log |\gS| \leq O(\log |\gV|)$ that performs Segmented Retrieval (Definition \ref{dfn:seg_ret}).
\end{theorem}

\begin{proof}
We provide the various components of the \m{TS} architecture in distinct sections.
    
\textbf{Token Embeddings.} We define $\psi':\mathcal{M}\rightarrow \{\pm1 \}^{\log|\mathcal{M}|}$ to be the binary encoding of the vocabulary in $\mathcal{M}$.
\\
\\ \textbf{Positional Encoding.} We define $\phi':[L]\rightarrow \{ \pm1\}^{\log|L|}$ as the binary encoding of numbers in $[L]$. After adding the positional encoding, the input context has the following form:
\begin{align*}
\text{TF-input}(\vec{\mathbf{x}}) := \begin{pmatrix}
\psi'(\mathbf{x}_1) & \cdots & \psi'(\mathbf{x}_{L-1}) & \psi'(\mathbf{x}_L) \\
\mathbf{0} & \cdots & \mathbf{0} & \mathbf{0} \\
\phi'(1) & \cdots & \phi'(L-1) & \phi'(L) \\
\mathbf{0} & \cdots & \mathbf{0} & \mathbf{0} \\
\mathbf{1} & \cdots & \mathbf{1} & \mathbf{1} \\
\end{pmatrix}
\end{align*}
\\
\\
\textbf{Transformer Layer.} As outlined in the high-level idea above, we need to extract the token $\mathbf{x}_{t-\lambda}$ at each position $t$, from which the sliding window size is $\lambda+1$, and we add the positional bias $\mathbf{B}_{t-\lambda}$ into the attention:
\begin{align*}
    Attn_j(TF-input(\vec{\mathbf{x}})):=\sum_{i=\max(1,j-\lambda)}^{j}\frac{\exp((\mathbf{B}_{j-\lambda})_{i})}{\sum_{i=\max(1,j-\lambda)}^{j}\exp((\mathbf{B}_{j-\lambda})_i)} \psi'(\mathbf{x}_{i})=\psi'(\mathbf{x}_{j-\lambda}),
\end{align*}
where $(\mathbf{B}_{j-\lambda})_i=0,i=j-\lambda,(\mathbf{B}_{j-\lambda})=-\infty,\forall i\neq j-\lambda$. That is to say we set $\mathbf{W}_q=\mathbf{W}_k=0$ and $\mathbf{W}_v(TF-input(\vec{\mathbf{x}}))=(\psi'(\mathbf{x}_1),\cdots,\psi'(\mathbf{x}_{L}))^{\top}$.
Therefore, the output of the transformer layer is as follows.
\begin{align*}
\begin{pmatrix}
\psi'(\mathbf{x}_1) & \cdots & \psi'(\mathbf{x}_{L-1}) & \psi'(\mathbf{x}_L) \\
\mathbf{0} & \cdots & \psi'(\mathbf{x}_{L-1-\lambda}) & \psi'(\mathbf{x}_{L-\lambda}) \\
\phi'(1) & \cdots & \phi'(L-1) & \phi'(L) \\
\mathbf{0} & \cdots & \mathbf{0} & \mathbf{0} \\
\mathbf{1} & \cdots & \mathbf{1} & \mathbf{1} \\
\end{pmatrix}
\end{align*}
Here we set that $\forall j\le \lambda,  Attn_j(TF-input(\vec{\mathbf{x}}))=\mathbf{0}$.
\\
\\
\textbf{SSM Layer.} As outlined in the high-level idea of the SSM layer construction, we only need one SSM layer to solve the task. At each position $t$, we denote the largest index of the set $\{t_j-\lambda:j\in [k-1],\mathbf{x}_{t_j-\lambda}\in \mathcal{N}, 1\le t_j-\lambda\le t \}$ as $t^{\star}$, and define $H_t:=\mathbf{x}_{t^{\star}}$. Then, as stated in the high-level idea, the update rule for $H_t$ is 
\begin{align*}
    H_t=(I-\mathbf{1}_{\{(t\in \mathbf{T})\land (\mathbf{x}_{t-\lambda}\in \mathcal{N})\}})H_{t-1}+\mathbf{1}_{\{(t\in \mathbf{T})\land (\mathbf{x}_{t-\lambda}\in \mathcal{N})\}}\mathbf{x}_{t-\lambda},
\end{align*}
This is achieved by setting
\begin{align*}
\mathbf{W}_A=\mathbf{W}_C=I, \quad \mathbf{W}_B(
\begin{pmatrix}
\psi'(\mathbf{x}_t) \\
\psi'(\mathbf{x}_{t-\lambda}) \\
\phi'(t) \\
\mathbf{0} \\
\mathbf{1}
\end{pmatrix})=\mathbf{x}_{t-\lambda}, \quad \Delta(
\begin{pmatrix}
\psi'(\mathbf{x}_t) \\
\psi'(\mathbf{x}_{t-\lambda}) \\
\phi'(t) \\
\mathbf{0} \\
\mathbf{1}
\end{pmatrix})=\mathbf{1}_{\{(t\in \mathbf{T})\land (\mathbf{x}_{t-\lambda}\in \mathcal{N})\}}.
\end{align*}

\end{proof}

\subsection{SSMs with a Convolutional Front-End}
\label{app:ssm_conv}

A Mamba block applies a causal depthwise convolution before its
selective SSM. Each such layer can therefore read a few recent inputs
directly, in addition to its recurrent state. This subsection makes the
local access explicit, relates it to the $\mathrm S$-layers of
Section~2, and quantifies its effect on the lower bounds of
Section~2.2.

\begin{definition}[$\mathrm S_\kappa$-layer]
\label{dfn:s_kappa}
Let $\kappa\geq 0$. A layer
$F:(\{0,1\}^P)^L\to(\{0,1\}^P)^L$ is an \emph{$\mathrm S_\kappa$-layer}
with recurrent state space $\mathcal S$ if there exist an initial state
$S_0\in\mathcal S$ and maps
$\mu:\mathcal S\times(\{0,1\}^P)^{\kappa+1}\to\mathcal S$ and
$\psi:\mathcal S\times(\{0,1\}^P)^{\kappa+1}\to\{0,1\}^P$ such that
\[
  S_t=\mu\big(S_{t-1},\mathbf x_{t-\kappa:t}\big),
  \qquad
  F(\vec{\mathbf x})_t=\psi\big(S_t,\mathbf x_{t-\kappa:t}\big),
\]
where $\mathbf x_i$ for $i\leq 0$ is a fixed padding symbol $\diamond$
whose encoding differs from that of every token.
We call $\mathbf x_{t-\kappa:t}$ the \emph{local window} and
$\log_2|\mathcal S|$ the \emph{recurrent-state complexity};
the local window is not charged to $\mathcal S$.
\end{definition}

\paragraph{Mamba.}
With finite-precision arithmetic, a Mamba block (input projection,
causal depthwise convolution of width $\kappa+1$, selective SSM, gating,
output projection, and residual connection) is an
$\mathrm S_\kappa$-layer with $\kappa=d_{\mathrm{conv}}-1$, whose
recurrent state is the SSM state: both the state update and the output at
position $t$ depend only on the previous state and on
$\mathbf x_{t-\kappa:t}$. Lower bounds for $\mathrm S_\kappa$-layers
therefore apply to Mamba. The converse does not hold. Since $\mu$ and
$\psi$ are arbitrary, a construction in the $\mathrm S_\kappa$ class
does not by itself yield parameters of a Mamba block.

\paragraph{Relation to the $\mathrm S$-layers of Section~2.}
An $\mathrm S$-layer of Section~2 is an $\mathrm S_0$-layer with the same
state space (take $\psi(S,\mathbf x_t)=r(S)$), and every
$\mathrm S_0$-layer is an $\mathrm S_\kappa$-layer with the same state
space. The converse fails at equal budget, because $\psi$ may read the
current input. For instance, with $|\mathcal S|=1$ and
$\psi(S,\mathbf x_t)=\mathbf x_t$, an $\mathrm S_0$-layer computes the
identity with zero recurrent bits, whereas an $\mathrm S$-layer of
Section~2 needs $P$ bits of state to do so. Two facts relate the
classes.
\begin{enumerate}
  \item[(i)] \emph{Simulation.} An $\mathrm S_\kappa$-layer with
  recurrent state space $\mathcal S$ is an $\mathrm S$-layer of Section~2
  with state space $\mathcal S\times(\{0,1\}^P)^{\kappa+1}$: store
  $\widetilde S_t=(S_t,\mathbf x_{t-\kappa:t})$, update the window as a
  shift register, and set $r(\widetilde S_t)=\psi(S_t,\mathbf x_{t-\kappa:t})$.
  Its bit complexity is at most $\log_2|\mathcal S|+(\kappa+1)P$.
  \item[(ii)] \emph{Configuration at a cut.} For every $\tau$, the pair
  $(S_\tau,\mathbf x_{\tau-\kappa+1:\tau})$, of at most
  $\log_2|\mathcal S|+\kappa P$ bits, determines $F(\vec{\mathbf x})_t$ for
  all $t>\tau$ given $\mathbf x_{\tau+1:t}$. This is the case $D_{\mathrm S}=1$
  of Lemma~\ref{lem:conv_cut} below.
\end{enumerate}
By (ii), the proof of Theorem~\ref{thm:hybrid_cut} applies to
$\mathrm S_\kappa\mathrm T$ models with
$|\mathcal S|\,2^{\kappa P}$ in place of $|\mathcal S|$.
Thus, Theorems~\ref{thm:exact_copy_lb}
and~\ref{thm:seg_ret_lb} continue to hold with
$\log_2|\mathcal S|+\kappa P$ in place of
$\log_2|\mathcal S|$.
By (i) applied layerwise, a hybrid model with profile $A$ whose
recurrent layers are $\mathrm S_\kappa$-layers with at most $B$ recurrent
bits is a model in
$\mathcal M_A(B+(\kappa+1)P,P,W)$
in the sense of Section~\ref{sec theory}.
Since
\[
B+(\kappa+1)P
\le
(\kappa+2)\max\{B,P\},
\]
the conclusion of part~2 of
Theorem~\ref{th equal-depth-general-profile}
continues to hold for such models, with
$c_\tau m/(\kappa+2)$ in place of $c_\tau m$.
The constructions need no change, because every $\mathrm S$-layer is an
$\mathrm S_\kappa$-layer.
The constant-factor separation in part~3 of
Theorem~\ref{th equal-depth-general-profile}
does not follow from this conversion.

For the natural tasks of Section~2.2, charging the window to the state
is unnecessary. The following lemma and proposition show that each
$\mathrm S_\kappa$-layer reduces the lower bounds by at most $\kappa$
tokens, and that the resulting threshold on the task distance is tight within the $ S_\kappa$ abstraction.

\begin{lemma}[Cut through a convolutional stack]
\label{lem:conv_cut}
Let $M=\mathrm T\circ F_{D_{\mathrm S}}\circ\cdots\circ F_1$, where each
$F_i$ is an $\mathrm S_\kappa$-layer with recurrent state space
$\mathcal S^{(i)}$ and states $S^{(i)}_t$, and $\mathrm T$ is a
$\mathrm T$-layer with window $W$; write $M(\vec{\mathbf x})$ for the
output at position $L$. Let $D_{\mathrm S}\kappa<\tau\leq L-W$ and
$\tau_i:=\tau-(D_{\mathrm S}-i)\kappa$ for $i\in[D_{\mathrm S}]$.
Let $\mathbf u,\mathbf u'$ be prefixes of length $\tau$ that agree on
positions $\tau-D_{\mathrm S}\kappa+1,\ldots,\tau$ and satisfy
$S^{(i)}_{\tau_i}(\mathbf u)=S^{(i)}_{\tau_i}(\mathbf u')$ for all
$i\in[D_{\mathrm S}]$. Then $M(\mathbf u\mathbf q)=M(\mathbf u'\mathbf q)$
for every suffix $\mathbf q$ of length $L-\tau$.
\end{lemma}

\begin{proof}
Write $\mathbf o^{(0)}=\vec{\mathbf x}$ and
$\mathbf o^{(i)}=F_i(\mathbf o^{(i-1)})$, and compare the inputs
$\mathbf u\mathbf q$ and $\mathbf u'\mathbf q$. We show by induction on
$i$ that $S^{(i)}_t$ and $\mathbf o^{(i)}_t$ coincide on the two inputs
for all $t>\tau_i$.
For $i=1$ and $t>\tau_1$, the window $\mathbf x_{t-\kappa:t}$ starts at
position $t-\kappa\geq\tau_1-\kappa+1=\tau-D_{\mathrm S}\kappa+1$, so it
lies in the common part of the prefixes or in the common suffix.
Together with $S^{(1)}_{\tau_1}(\mathbf u)=S^{(1)}_{\tau_1}(\mathbf u')$,
induction on $t$ gives the claim.
For $i\geq2$ and $t>\tau_i$, the window $\mathbf o^{(i-1)}_{t-\kappa:t}$
starts at position $t-\kappa>\tau_i-\kappa=\tau_{i-1}$, where the two
inputs agree by the induction hypothesis. Together with
$S^{(i)}_{\tau_i}(\mathbf u)=S^{(i)}_{\tau_i}(\mathbf u')$, the claim
follows as before.
Finally, $\tau_{D_{\mathrm S}}=\tau$, and $\mathrm T$ reads only positions
$L-W+1,\ldots,L$, all larger than $\tau$. Hence the outputs coincide.
\end{proof}

\begin{theorem}[Natural tasks with convolutional recurrent layers]
\label{prop:conv_tasks}
Let $D_{\mathrm S}\geq1$, let $h$ be as in
Theorem~\ref{thm:exact_copy_lb}
(resp.\ Theorem~\ref{thm:seg_ret_lb}), and set
\[
h_{D_{\mathrm S}}
:=
(h-D_{\mathrm S}\kappa)_+.
\]
\begin{enumerate}
  \item[(a)] Under the assumptions of Theorem~\ref{thm:exact_copy_lb}
  (resp.\ Theorem~\ref{thm:seg_ret_lb}), every deterministic model $M$ as
  in Lemma~\ref{lem:conv_cut} that performs Exact Copy
  (resp.\ Segmented Retrieval) correctly on all inputs satisfies
  \[
    \sum_{i=1}^{D_{\mathrm S}}\log_2|\mathcal S^{(i)}|
    \;\geq\;h_{D_{\mathrm S}}\log_2(|\mathcal V|-2)
  \]
  (resp.\ the same with $\log_2|\mathcal N|$ in place of
  $\log_2(|\mathcal V|-2)$).
  \item[(b)] Conversely, suppose $l\leq D_{\mathrm S}\kappa$
  (resp.\ $\lambda\leq D_{\mathrm S}\kappa$) and
  $P\geq2\lceil\log_2(|\mathcal V|+1)\rceil+2$. Then there is a model $M$
  as in Lemma~\ref{lem:conv_cut}, with $\mathrm T$ the identity
  ($W=1$), in which $F_1,\ldots,F_{D_{\mathrm S}-1}$ have a single
  recurrent state and $F_{D_{\mathrm S}}$ has $O(\log(cl))$
  (resp.\ $O(\log|\mathcal N|)$) recurrent bits, that performs Exact Copy
  (resp.\ Segmented Retrieval) correctly on all inputs.
\end{enumerate}
In particular, with $W=1$ (so that $h=l$, resp.\ $h=\lambda$), the bound
in (a) is nonzero exactly when the task distance exceeds
$D_{\mathrm S}\kappa$, and otherwise (b) gives a solution with small
recurrent state.
\end{theorem}

\begin{proof}
\emph{(a).} If $h_{D_{\mathrm S}}=0$ there is nothing to prove, so assume
$h_{D_{\mathrm S}}\geq1$. Then $h=h_{D_{\mathrm S}}+D_{\mathrm S}\kappa$,
and we use the cut $\tau=h$, which satisfies
$D_{\mathrm S}\kappa<\tau\leq L-W$ by the definition of $h$.
By Lemma~\ref{lem:conv_cut}, if a family of prefixes of length $h$ shares
its last $D_{\mathrm S}\kappa$ positions and any two of its members are
mapped to different outputs by some common suffix, then the map
$\mathbf u\mapsto\big(S^{(i)}_{\tau_i}(\mathbf u)\big)_{i\in[D_{\mathrm S}]}$
is injective on the family. Its image has at most
$\prod_i|\mathcal S^{(i)}|$ elements, so it suffices to exhibit such a
family of size $(|\mathcal V|-2)^{h_{D_{\mathrm S}}}$
(resp.\ $|\mathcal N|^{h_{D_{\mathrm S}}}$).

For Exact Copy, let $\mathcal A,\alpha,g_1,g_2$ and $\mathbf g_{1:L-cl}$
be as in the proof of Theorem~\ref{thm:exact_copy_lb}, and write
$h'=h_{D_{\mathrm S}}$. For $a\in\mathcal A^{h'}$, use the prefix
$a\,\alpha^{D_{\mathrm S}\kappa}$ and the suffix
$\alpha^{\,l-h}\big(a\,\alpha^{\,l-h'}\big)^{c-1}\mathbf g_{1:L-cl}$.
For $a'\in\mathcal A^{h'}$, the resulting input is
\[
  a'\alpha^{D_{\mathrm S}\kappa}\cdot(\text{suffix of }a)
  \;=\;a'\alpha^{\,l-h'}\big(a\,\alpha^{\,l-h'}\big)^{c-1}\mathbf g_{1:L-cl},
\]
which is the input analysed in the proof of Theorem~\ref{thm:exact_copy_lb}
with $h$ replaced by $h'\in[1,l]$. It therefore contains an exact copy if
and only if $a'=a$.

For Segmented Retrieval, fix $\alpha\in\mathcal V$ and use, for
$n\in\mathcal N^{h'}$, the prefix
\[
  \big((n_1,0),\ldots,(n_{h'},0),
  \underbrace{(\alpha,0),\ldots,(\alpha,0)}_{D_{\mathrm S}\kappa}\big).
\]
For $j\in[h']$, append the suffix $q_j$ from the proof of
Theorem~\ref{thm:seg_ret_lb}. Its only internal boundary is at position
$j+\lambda>h$, the segment-length conditions of
Definition~\ref{dfn:seg_ret} hold because $L\geq3\lambda+1$ and
$j\leq\lambda$, and the required output is $\mathbf x_j=n_j$. Two
distinct prefixes differ at some $j\in[h']$, and $q_j$ separates them.

\emph{(b).} Encode each position by its token (or $\diamond$) and, for
Segmented Retrieval, its boundary bit; two such encodings fit in $P$ bits.
Let $j_0=0$ and $j_i=\min(\delta,i\kappa)$, where $\delta=l$
(resp.\ $\delta=\lambda$) is the task distance.
For $i<D_{\mathrm S}$, let $F_i$ have a single recurrent state and output
$\mathbf o^{(i)}_t=(\mathbf x_t,\mathbf x_{t-j_i})$, with $\diamond$ for
nonpositive indices. This is possible because $F_1$ reads
$\mathbf x_{t-j_1}$ directly ($j_1\leq\kappa$), and for $i\geq2$ the
window of $F_i$ contains $\mathbf o^{(i-1)}_t$, whose first entry is
$\mathbf x_t$, and $\mathbf o^{(i-1)}_{t-(j_i-j_{i-1})}$, whose second
entry is $\mathbf x_{t-j_i}$, since $0\leq j_i-j_{i-1}\leq\kappa$.
Because $\delta\leq D_{\mathrm S}\kappa$, we have
$\delta-j_{D_{\mathrm S}-1}\leq\kappa$, so the window of
$F_{D_{\mathrm S}}$ contains both $\mathbf x_t$ and $\mathbf x_{t-\delta}$
(for $D_{\mathrm S}=1$, directly from $\mathbf x_{t-\kappa:t}$).

For Exact Copy, $F_{D_{\mathrm S}}$ keeps
$H_t=\min\{(c-1)l,\ \mathbf 1\{\mathbf x_t=\mathbf x_{t-l}\}(H_{t-1}+1)\}$
and $y_t=y_{t-1}\vee\mathbf 1\{H_t=(c-1)l\}$, and outputs $y_t$. Since
$\diamond$ differs from every token, $\mathbf x_t=\mathbf x_{t-l}$ can hold
only for $t\geq l+1$. By Definition~\ref{dfn:exact_copy}, an exact copy
exists if and only if $\mathbf x_t=\mathbf x_{t-l}$ for $(c-1)l$
consecutive positions $t$. So $y_L$ is correct, and
$2((c-1)l+1)$ states suffice.
For Segmented Retrieval, $F_{D_{\mathrm S}}$ stores the last value
$\mathbf x_{t-\lambda}\in\mathcal N$ read at a boundary position $t$, or
$\bot$ before the first such value, and outputs it. This uses
$|\mathcal N|+1$ states, and the value stored at position $L$ is
$\mathbf x_{t_{\lambda^\star}-\lambda}$.
\end{proof}

\end{document}